\documentclass{article}

\usepackage{iclr2027_conference,times}

\usepackage{amsmath,amsfonts,amssymb}
\usepackage{graphicx}
\usepackage{booktabs}
\usepackage{multirow}
\usepackage{hyperref}
\usepackage{algorithm}
\usepackage{algorithmic}
\usepackage{subcaption}
\usepackage{xcolor}
\usepackage{wrapfig}
\usepackage{pifont}
\usepackage[table]{xcolor}
\usepackage{subcaption}
\usepackage{float}
\usepackage{placeins}

\title{TRACE-MAS: Diagnosing and Repairing Collaboration Failures in Multi-Agent Systems}

\author{
\normalfont
Zekai Chen$^{1}$ \quad
Yihan Sun$^{2}$ \quad
Xunkai Li$^{1}$ \quad
Xun Wu$^{1}$ \quad
Xinmo Jin$^{1}$ \\
Yinlin Zhu$^{3}$ \quad
Haodong Lu$^{1}$ \quad
Xuaner Chen$^{4}$ \quad
Rong-Hua Li$^{1}$ \\[1ex]
$^{1}$Beijing Institute of Technology \\
$^{2}$Minzu University of China \\
$^{3}$Sun Yat-sen University \\
$^{4}$Shandong University, Weihai \\
\texttt{Email of Zekai Chen: zackchen02@163.com}
}
\iclrfinalcopy
\begin{document}

\maketitle

\begin{abstract}
LLM-based multi-agent systems (MAS) rely on not only individual agent behaviors but also the interactions that connect agents, including message exchanges and task dependencies. Existing MAS safeguards mainly perform agent-level anomaly detection, which becomes insufficient when failures originate from agent interactions and detected anomalies do not necessarily reveal their true sources. To address this limitation, we propose \textbf{TRACE-MAS}, a finer-grained framework for diagnosing and repairing MAS failures. TRACE-MAS learns normal collaborative execution patterns from benign trajectories and detects deviations in agent state changes, message transitions, and task dependency evolutions. It further identifies responsible execution elements through source-aware attribution and selects minimal interventions to restore collaboration while preserving task-critical capabilities. Experiments on five MAS benchmarks demonstrate that TRACE-MAS
significantly improves root-cause localization accuracy by \textbf{60.2}\% and achieves
consistent gains in task recovery over strong MAS safeguards.
\end{abstract}

\section{Introduction}
\label{sec:introduction}

LLM-based multi-agent systems (MAS) coordinate specialized agents through role assignment, communication, tool utilization, and task planning, enabling complex applications such as software development \citep{HongEtAl2023MetaGPT}, embodied decision making \citep{SongEtAl2023LLMPlanner}, and collaborative scientific reasoning \citep{TangEtAl2023MedAgents}. However, unlike single-agent systems, MAS performance depends not only on individual agent behaviors, but also on the correct execution of interactions among agents. These dependencies introduce new attack surfaces: adversaries may inject malicious information into agent prompts or memories \citep{ChenEtAl2024AgentPoison}, manipulate messages exchanged between agents \citep{HeEtAl2025AiTM}, or interfere with task assignment and execution dependencies \citep{ZhouEtAl2025CORBA,KhanEtAl2025AgentUnderSiege}.

Recent MAS safeguards have substantially improved the detection and mitigation of compromised agents. Existing approaches mainly identify suspicious agents through interaction structures, observable behaviors, or system-level signals \citep{WangEtAl2025GSafeguard,MiaoEtAl2026BlindGuard}. However, these methods are primarily designed around agent-level anomaly detection: they treat individual agents as the basic detection units and answer the question of \emph{which agent appears abnormal}. This coarse detection granularity becomes insufficient when failures originate from the interactions among agents rather than the agents themselves. For example, attackers can modify exchanged messages or disrupt task dependencies, causing normal agents to exhibit abnormal behaviors even though the actual corruption occurs in the collaborative process. Therefore, a key challenge is to identify corrupted execution elements at a finer granularity, while determining how the corresponding collaboration can be effectively recovered.

To investigate this challenge, we conduct a diagnostic empirical study in Sec.~\ref{sec:empirical}, revealing three connected observations. \ding{182} \textbf{\underline{Agent-Level Blind Spot.}} Existing defenses are effective when attacks directly modify agent states. However, MAS execution can also be disrupted by modifying exchanged messages or altering task dependencies without changing individual agent profiles. As shown in Fig.~\ref{fig:obs1}, these execution-level attacks cause substantial task failures, indicating that agent abnormality alone cannot fully characterize MAS corruption. \ding{183} \textbf{\underline{Anomaly-Source Mismatch.}} A natural solution is to extend monitoring beyond individual agents. However, broader anomaly detection still cannot determine where the failure originates. Once corrupted information propagates through normal collaboration, downstream components may exhibit stronger abnormal signals than the original source. As shown in Fig.~\ref{fig:obs2}, the most abnormal component is not necessarily where collaboration first deviates, demonstrating that anomaly localization and source identification are different objectives. \ding{184} \textbf{\underline{Repair-Target Mismatch.}} Furthermore, the mismatch between anomaly and source directly limits recovery effectiveness, since repairing the most abnormal component may only remove propagated symptoms, while isolating suspicious components can unnecessarily eliminate useful capabilities. As shown in Fig.~\ref{fig:obs3}, source-aware repair achieves better recovery with fewer interventions than symptom-based strategies. Together, these observations show that existing MAS safeguards are limited by their agent-level detection granularity, which motivates a finer-grained modeling of collaborative execution beyond individual agents. Specifically, TRACE-MAS considers whether an agent's internal execution state changes abnormally, whether the information exchanged between agents is inconsistent with expected interactions, and whether task assignments or execution dependencies deviate from normal collaboration patterns.

Motivated by this perspective, we propose \textbf{TRACE-MAS}, a finer-grained framework for detecting, locating, and repairing MAS failures. TRACE-MAS models MAS execution as a structured event process and learns normal execution patterns from benign trajectories. It detects deviations in agent states, message transitions, and task dependency evolutions, identifies responsible execution elements through source-aware attribution, and selects minimal interventions to restore collaboration. By integrating detection, diagnosis, and recovery within a unified execution model, TRACE-MAS extends MAS safeguarding from identifying abnormal agents to fine-grained understanding and repairing corrupted collaborative execution.

\textbf{Our Contributions.} \textit{\textbf{\underline{(1) New Problem.}}} We study execution-level failure diagnosis and recovery in LLM-based MAS, where abnormal behaviors may originate from agent states, communication messages, or coordination operations. We reveal that detected anomalies and actual failure sources may not coincide, limiting existing agent-centric safeguards. \textit{\textbf{\underline{(2) New Framework.}}} We propose TRACE-MAS, which learns normal collaborative execution patterns from benign trajectories and performs unified execution-level detection, source-aware attribution, and minimal recovery. \textit{\textbf{\underline{(3) SOTA Performance.}}}
Experiments on \textbf{5 benchmarks} covering reasoning, retrieval, and
tool-use MAS under \textbf{3 attack settings} demonstrate that TRACE-MAS outperforms existing MAS
safeguards, achieving an average improvement of \textbf{15.5 points.}

\section{Problem Formulation}
\label{sec:problem}

We consider an LLM-based multi-agent system with agents
$\mathcal{A}=\{A_1,\ldots,A_N\}$ collaboratively solving a task $q$ through
role specialization, message exchange, and task dependencies. Instead of
viewing agents independently, we model MAS execution as a partially ordered
event process. Each event is represented as
\begin{equation}
e_t=
\big(
i_t,r_t,q_t,x_t,h_t,
m_t^{\mathrm{in}},m_t^{\mathrm{out}},a_t,s_t
\big),
\label{eq:event}
\end{equation}
where $i_t$, $r_t$, and $q_t$ denote the agent, role, and subtask;
$x_t$ and $h_t$ represent execution context and agent state;
$m_t^{\mathrm{in}}$ and $m_t^{\mathrm{out}}$ denote exchanged messages; and
$a_t$, $s_t$ describe availability and task progress.
The execution process is organized as a structured graph
\begin{equation}
\mathcal{G}
=
\big(
\mathcal{V},
\mathcal{E}^{\mathrm{msg}}
\cup
\mathcal{E}^{\mathrm{dep}}
\cup
\mathcal{E}^{\mathrm{temp}}
\big),
\label{eq:event_graph}
\end{equation}
where the edges capture message exchange, task dependency, and temporal order,
and $\mathrm{Pa}(t)$ denotes the direct predecessors of $e_t$ in $\mathcal{G}$.
Let $\mathbf{o}_t=(\Delta h_t,\Delta m_t,\Delta s_t)$ denote the execution
transition of event $e_t$, corresponding to changes in agent states, message
interactions, and task progress.
In implementation, textual and structured fields in Eq.~\eqref{eq:event} are embedded into the contract space as in Appendix~\ref{app:graph} (Eq.~\eqref{eq:app-z0}); the transition residual $\Delta e_t$ used in Sec.~\ref{sec:contract} denotes the same object as $\mathbf{o}_t$.
Under benign collaboration, these transitions follow:
\begin{equation}
p_0(\mathcal{G}\mid q)
=
\prod_{e_t\in\mathcal{V}}
p_0
\left(
\mathbf{o}_t
\mid
r_t,q_t,x_t,
\mathbf{o}_{\mathrm{Pa}(t)}
\right),
\label{eq:benign_execution}
\end{equation}
where $p_0$ describes normal execution dynamics conditioned on roles, tasks,
and preceding events.
Attacks can intervene on different execution elements, including agent states,
message transitions, and task dependencies. We denote the resulting corrupted
execution as $\widetilde{\mathcal{G}}$. Given benign trajectories
$\mathcal{D}_0=\{\mathcal{G}^{(1)},\ldots,\mathcal{G}^{(M)}\}$ and an observed
execution $\widetilde{\mathcal{G}}$, the objective is to:
\emph{(i)} detect deviations from normal collaboration;
\emph{(ii)} identify the responsible execution element rather than propagated
effects; and
\emph{(iii)} restore subsequent execution with minimal intervention.

\section{Related Work}
\label{sec:related}

\subsection{Attacks on LLM-based MAS.}
LLM-based MAS inherit single-model risks such as jailbreaks and prompt injection
\citep{PerezEtAl2022RedTeaming,WeiEtAl2023Jailbreak,greshake2023not}, and add new ones through agent interaction.
AgentPoison corrupts agent memory and retrieval \citep{ChenEtAl2024AgentPoison}, AiTM and Prompt
Infection show that inter-agent messages can carry and spread malicious content
\citep{HeEtAl2025AiTM,lee2025prompt}, and recent attacks target coordination structures and
execution dynamics \citep{ZhouEtAl2025CORBA}, whose damage depends on the
collaboration topology \citep{huang2025resilience}. These works analyze specific attack mechanisms,
leaving open how to diagnose failures across heterogeneous collaborative executions.

\subsection{Safeguarding LLM-based MAS.}
Existing defenses detect suspicious behaviors and mitigate compromised components. G-Safeguard and
NetSafe exploit the interaction topology \citep{WangEtAl2025GSafeguard,yu2025netsafe}, BlindGuard
handles unknown attacks \citep{MiaoEtAl2026BlindGuard}, AgentMonitor predicts risk from agent-level
signals \citep{chan2024agentmonitor}, and single-agent methods self-correct through execution
feedback \citep{ShinnEtAl2023Reflexion,YaoEtAl2023ReAct}. Failure-attribution studies catalogue MAS
failure modes \citep{cemri2025why} and locate the failing agent or step from logs
\citep{zhang2025which,zhou2025guardiansafeguardingllmmultiagent}, but report where a failure becomes visible rather
than repairing where it originates. \textsc{Trace-Mas} complements these efforts by learning benign
execution contracts, tracing the initiating violation, and repairing the source that disrupts
subsequent execution.

\section{Empirical Study}
\label{sec:empirical}

Before introducing TRACE-MAS, we investigate why existing MAS safeguards fail to achieve reliable diagnosis and recovery.
We conduct controlled attacks on agent states, message exchanges, and task dependencies while keeping tasks and system configurations unchanged (detailed attack protocols are provided in Appendix~\ref{app:attacks}).
As shown in Figs.~\ref{fig:obs1}--\ref{fig:obs3}, these studies reveal three connected observations: failures may originate beyond individual agents, observed anomalies may not indicate their sources, and effective recovery requires source-aware intervention.

\textbf{Observation 1: Agent-centric monitoring misses relational corruption.}
Existing safeguards mainly associate MAS failures with abnormal agents.
However, message exchanges and task dependencies can be disrupted without directly modifying agents.
As shown in Fig.~\ref{fig:obs1}, when attacks shift from agent states to agent interactions, representative safeguards experience a \textbf{22.7-point} performance drop under comparable task failures, indicating that agent abnormality alone cannot characterize MAS corruption.

\textbf{Observation 2: Observed anomaly does not reveal the root cause.}
Extending monitoring beyond agents does not directly solve diagnosis.
Once corruption propagates through normal collaboration, downstream events may exhibit stronger deviations than the original source.
Fig.~\ref{fig:obs2} shows that the maximum-anomaly event identifies the true source in only \textbf{32.1\%} of cases, with the strongest deviation appearing \textbf{0.7 hops} away from the source on average.
Therefore, anomaly magnitude indicates where failure becomes visible rather than where it originates.

\textbf{Observation 3: Symptom repair cannot guarantee recovery.}
The mismatch between anomaly and source directly limits recovery.
Repairing the most abnormal event may leave the actual cause unchanged, while isolating suspicious components may remove useful capabilities.
As shown in Fig.~\ref{fig:obs3}, source-aware repair improves task recovery by \textbf{30.0 points} over anomaly-based repair while requiring \textbf{50.0\%} fewer interventions.

\begin{figure}[!ht]
    \centering

    \includegraphics[width=0.48\textwidth]{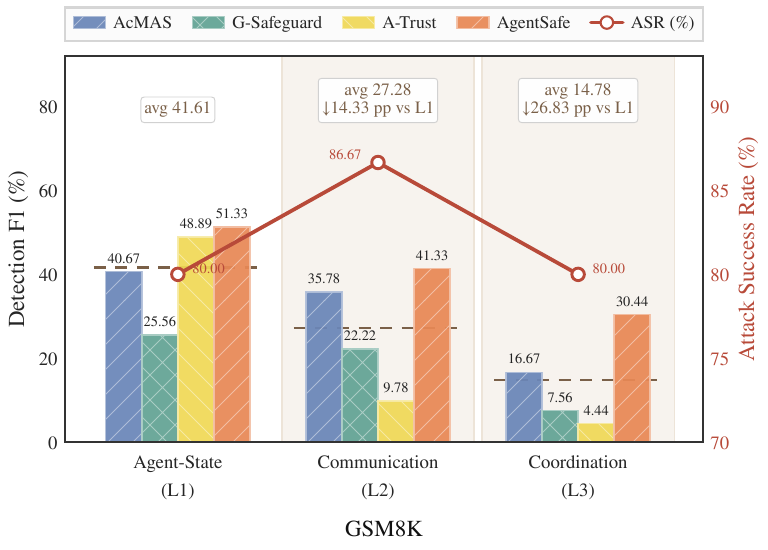}
    \hfill
    \includegraphics[width=0.48\textwidth]{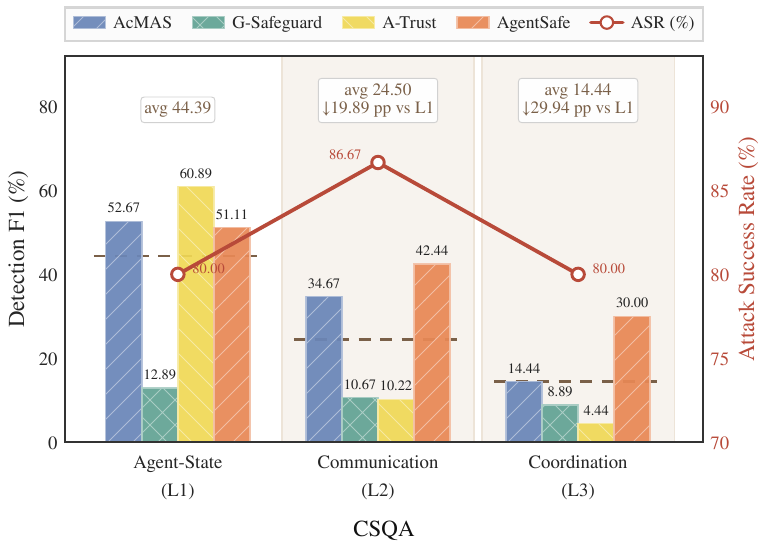}
    \captionof{figure}{Agent-centric monitoring misses relational corruption.
    Detection F1 vs.\ L1--L3 on GSM8K and CSQA (matched ASR); F1 drops 14--30 points while ASR stays flat.}
    \label{fig:obs1}

    \vspace{0em}

    \includegraphics[width=0.48\textwidth]{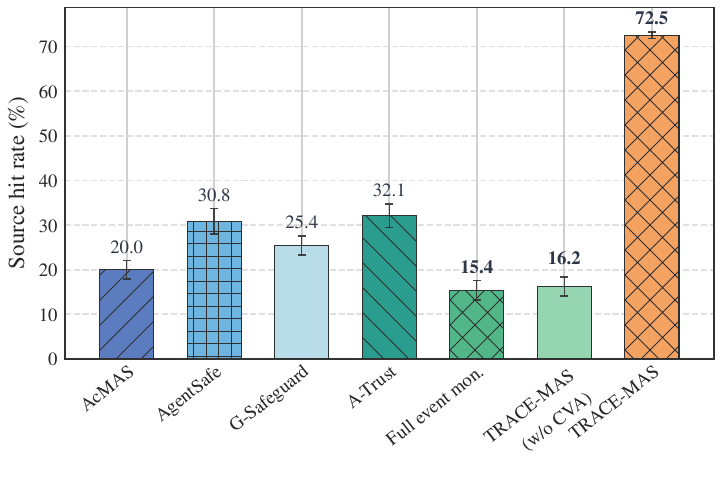}
    \hfill
    \includegraphics[width=0.48\textwidth]{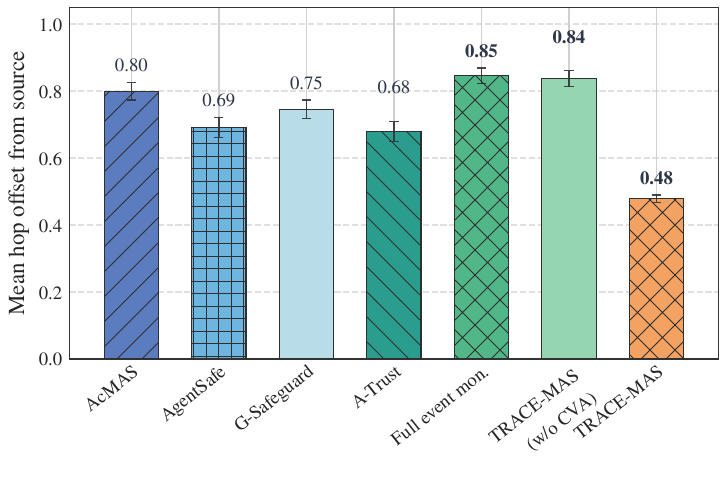}
    \captionof{figure}{Observed anomaly does not reveal the root cause.
    Source hit rate and hop offset on HotPotQA; magnitude monitors $\le$32.1\%, TRACE-MAS 72.5\%.}
    \label{fig:obs2}

    \vspace{0em}

    \includegraphics[width=0.32\textwidth]{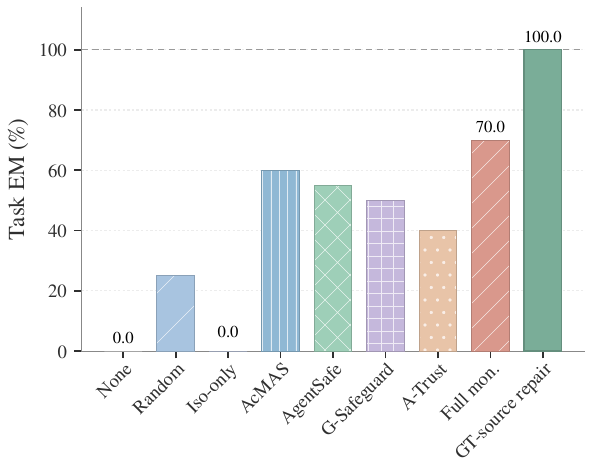}
    \hfill
    \includegraphics[width=0.32\textwidth]{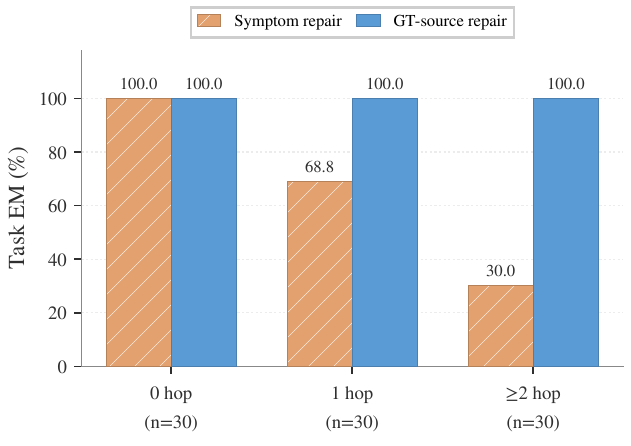}
    \hfill
    \includegraphics[width=0.32\textwidth]{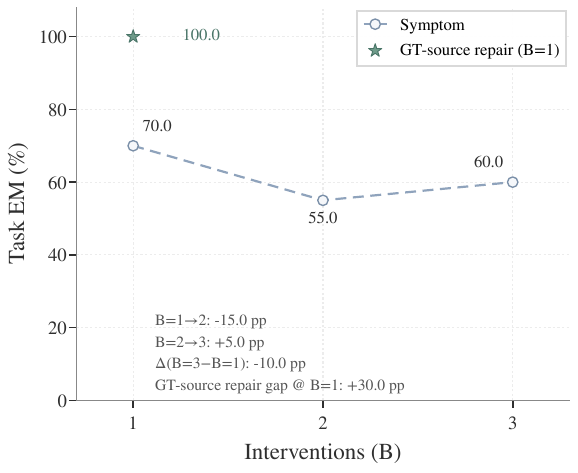}
    \captionof{figure}{Symptom repair cannot guarantee recovery.
    Task EM, hop to source, and symptom budget $B$; source repair hits 100\% EM, extra symptom steps do not help.}
    \label{fig:obs3}

\end{figure}

\begin{figure}[!t]
    \centering
    \includegraphics[width=1.0\textwidth]{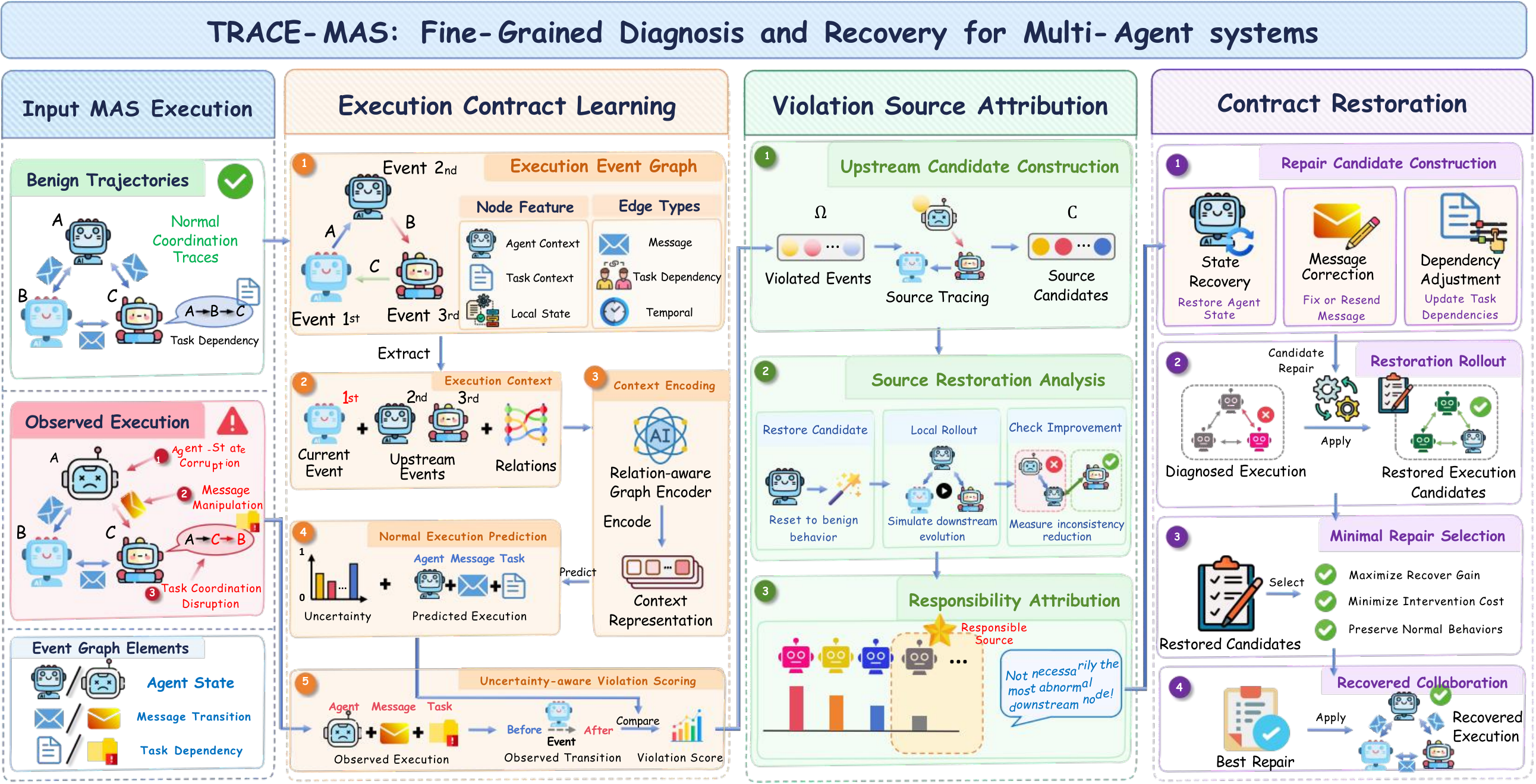}
    \caption{Overview of TRACE-MAS.
Partial-order event graph and shared contracts for detection, source attribution, and repair.}
    \label{fig:framework}
\end{figure}

\section{Methodology}
\label{sec:method}

\subsection{Overview}
\label{sec:overview}

As illustrated in Fig.~\ref{fig:framework}, TRACE-MAS builds a single execution contract that characterizes how reliable collaboration should evolve, and reuses it for three purposes: learning the benign execution
manifold conditioned on roles, tasks, and past interactions (Sec.~\ref{sec:contract}); attributing observed deviations to the execution element whose restoration removes them (Sec.~\ref{sec:attribution}); and selecting the minimal
intervention that returns future execution to that manifold (Sec.~\ref{sec:restoration}).

\subsection{Execution Contract Learning}
\label{sec:contract}

\textbf{Motivation.}
Observation~1 shows that agent-level abnormality cannot characterize MAS failures, since correctness
also depends on message interactions and task dependencies. \textsc{Trace-Mas} therefore learns a
conditional contract for how each event should evolve under its collaborative context.

\textbf{Execution Context Encoding.}
\textsc{Trace-Mas} aggregates upstream execution with a relation-aware encoder,
\begin{equation}
c_t = \mathrm{ExecFormer}\bigl(e_t, \{(e_j,\rho_{jt}) \mid e_j \in \mathrm{Pa}(t)\}\bigr),
\end{equation}
where $\mathrm{Pa}(t)$ are the direct predecessors of event $e_t$ (Sec.~2) and $\rho_{jt}$ is the
relation type (message, task dependency, or temporal order).

\textbf{Uncertainty-aware Execution Contract.}
As roles and tasks legitimately vary, the contract is a conditional distribution fit on benign
trajectories only:
\begin{equation}
p_\theta(\Delta e_t \mid c_t) = \mathcal{N}(\mu_t, \Sigma_t), \qquad
\mathcal{L}_{\mathrm{con}} = -\textstyle\sum_{G \in \mathcal{D}_0} \sum_{e_t \in G} \log p_\theta(\Delta e_t \mid c_t),
\end{equation}
where $\Delta e_t=(\Delta h_t,\Delta m_t,\Delta s_t)$ collects the agent-state, message, and
task-progress changes, with mean $\mu_t$ and adaptive uncertainty $\Sigma_t$. The violation score is
\begin{equation}
V(e_t) = (\widetilde{\Delta e}_t - \mu_t)^\top \Sigma_t^{-1} (\widetilde{\Delta e}_t - \mu_t),
\end{equation}
where $\widetilde{\Delta e}_t$ is observed in the test trajectory $\tilde{G}$; its message and progress
terms are weighted by $(\beta,\gamma)$ (Appendix~\ref{app:detection}), and $\tilde{G}$ is flagged when
$\max_{e_t} V(e_t)>\tau$.

\subsection{Violation Source Attribution}
\label{sec:attribution}

\textbf{Motivation.}
Observation~2 shows that corruption propagates through benign interactions, so a downstream event can
deviate more than the origin. \textsc{Trace-Mas} therefore tests whether restoring a candidate recovers
the subsequent execution.

\textbf{Source Restoration Analysis.}
Violated events are $\Omega=\{e_t \mid V(e_t)>\tau\}$, with
$\tau=\alpha\cdot\mathrm{Quantile}_\delta(\{V(e)\}_{e\in\mathcal{D}_{\mathrm{val}}})$ the $\alpha$-scaled
$\delta$-quantile of benign validation scores (Appendix~\ref{app:method-hparams}). Each candidate
$c\in\mathcal{C}$, an event of $\Omega$ or one of its $H$-hop ancestors, receives the contract
expectation under its context $c_c$:
\begin{equation}
\Delta e^{\mathrm{res}}_c = \mathbb{E}_{p_\theta}[\Delta e_c \mid c_c] = \mu_c, \quad
G^{\mathrm{res}}_c = \mathrm{Restore}_c(\tilde{G}, \Delta e^{\mathrm{res}}_c),
\end{equation}
where $\mathrm{Restore}_c$ propagates the restored state downstream in contract space without LLM calls
(Appendix~\ref{app:restore-rollout}).

\textbf{Responsibility Attribution.}
With $V^{\mathrm{res}}_c(e_j)$ the score recomputed on $G^{\mathrm{res}}_c$,
$\tilde{V}_c(e_j)=\omega V^{\mathrm{res}}_c(e_j)+(1-\omega)V(e_j)$, and $w_j=\psi^{\,d(c,e_j)}$ for hop
distance $d$,
\begin{equation}
R(c) = \!\!\sum_{e_j \in \{c\}\cup\mathrm{Desc}(c)}\!\! w_j \bigl[ V(e_j) - \tilde{V}_c(e_j) \bigr] - \lambda_{\mathrm{pen}} \mathcal{P}(c),
\qquad
c^\star = \arg\max_{c\in\mathcal{C}} R(c),
\end{equation}
where $\mathrm{Desc}(c)$ are the strict descendants of $c$, $\psi,\omega\in(0,1]$ are the hop decay and
evidence gate, and $\mathcal{P}(c)$ penalizes dependency or ordering violations
(Appendix~\ref{app:attribution-resp}). $R(c)$ is thus the hop-discounted violation drop on the branch
rooted at $c$, so the selected event explains the most propagated violation rather than deviating most.

\subsection{Contract Restoration}
\label{sec:restoration}

\textbf{Motivation.}
Observation~3 shows that repairing a visible anomaly may fail while the true source remains unchanged.
\textsc{Trace-Mas} therefore selects the intervention that returns future execution toward normality
with minimal change.

\textbf{Intervention-aware Future Rollout.}
Given $c^\star$, \textsc{Trace-Mas} builds a repair space $\mathcal{U}(c^\star)$ (state recovery, message
correction, dependency adjustment) and predicts the $K$-step future of each $u$ from the repaired step
$t$ as $\hat{G}^u_{t:t+K}=\mathrm{Rollout}(\tilde{G},u,p_\theta)$, with predicted events $\hat{e}^u_i$.

\textbf{Minimal Contract Restoration.}
The committed intervention minimizes residual violation, task degradation, and edit cost,
\begin{equation}
u^\star = \arg\min_{u \in \mathcal{U}(c^\star)} \left[ \sum_{i=t}^{t+K} V(\hat{e}^u_i) + \lambda_1 \mathcal{L}_{\mathrm{task}}(u) + \lambda_2 \kappa(u) \right],
\end{equation}
where $\mathcal{L}_{\mathrm{task}}$ is the fraction of workflow-template steps the rollout leaves unsatisfied, $\kappa(u)$ the operator
count, and $\lambda_1,\lambda_2$ trade-off weights (Appendix~\ref{app:ltask}); only the branches that
$u^\star$ affects are regenerated.

\section{Experiments}
\label{sec:experiments}

In this section, we conduct a comprehensive evaluation of TRACE-MAS to answer the following research questions.
\textbf{Q1: Detection Performance.} Can TRACE-MAS accurately detect execution failures across different attack types?
\textbf{Q2: Source Localization.} Can TRACE-MAS identify the true failure source rather than propagated anomalies?
\textbf{Q3: Recovery Effectiveness.} Can TRACE-MAS restore collaboration with fewer interventions while preserving task performance?
\textbf{Q4: Calibration and Robustness.} Can TRACE-MAS be calibrated from benign logs alone, remain stable under standard hyperparameters, and keep reliable detection under deployment shifts?
\textbf{Q5: Runtime Overhead.} Is the non-LLM cost of monitoring, attribution, and repair scoring practical when layered on standard MAS rollouts?

\subsection{Experimental Setup}
\label{sec:exp-setup}

We evaluate on five public LLM-based MAS benchmarks---GSM8K\citep{cobbe2021trainingverifierssolvemath} and CSQA\citep{talmor2019commonsenseqa} (multi-agent reasoning), HotPotQA\citep{yang-etal-2018-hotpotqa} and PoisonRAG\citep{zou2025poisonedrag} (retrieval-based collaboration), and InjecAgent\citep{zhan2024injecagent} (tool use)---under controlled L1--L3 corruptions on shared role graphs and splits.
We report trajectory-level detection, root-cause localization, and task-preserving recovery.
Following prior attack--defense evaluations, we compare TRACE-MAS with interaction-graph monitors (BlindGuard\citep{MiaoEtAl2026BlindGuard}, G-Safeguard\citep{WangEtAl2025GSafeguard}), trust- and agent-centric safeguards (A-Trust\citep{2025arXiv250602546H}, AgentSafe\citep{mao2025agentsafe}), activation-centric MAS monitors(AcMAS(-R)\citep{xu2026agents}), and Oracle Anomaly Ranking.
All methods share the same backbone LLM, MAS templates, attack injections, and benign-train / validation-calibration / attacked-test splits; baseline details, monitor variants, and calibration defaults are in Appendix~\ref{app:experiment} (Sec.~B.5) and Appendix~\ref{app:method}.

\subsection{Overall Performance (Q1)}
\label{sec:q1}

To answer \textbf{Q1}, we compare TRACE-MAS with the baselines in Sec.~\ref{sec:exp-setup} under the shared L1--L3 protocol.
As shown in Table~\ref{tab:detection}, TRACE-MAS achieves the highest trajectory F1 in every benchmark$\times$level cell, improving over the strongest competing method by \textbf{15.5}~points on average (per cell) and by up to \textbf{62.1}~points on PoisonRAG at Level~3.
Notably, TRACE-MAS remains above \textbf{52}~F1 on GSM8K and CSQA under Level~2--3 while graph monitors and activation-centric methods often fall below \textbf{20}~F1 when corruption targets messages or dependencies rather than agent states.
Together, the table supports learning benign state, message, and completion contracts and flagging violations $V(e_t)$ on the event graph for reliable execution-level detection across corruption types.

\begin{table*}[!t]
    \centering
    \caption{Detection on five benchmarks under L1--L3 attacks (trajectory F1 and AUROC; mean\(\pm\)std).
\textbf{Bold} best and \underline{underlined} second-best per column; shared splits and calibration.}
\scriptsize
    \setlength{\tabcolsep}{2.3pt}
    \renewcommand{\arraystretch}{1.05}
    \resizebox{0.99\textwidth}{!}{
    \begin{tabular}{lc|cc|cc|cc|cc|cc}
    \specialrule{1.5pt}{1.5pt}{1.5pt}
    \cmidrule{1-12}
    \multicolumn{2}{c|}{\textbf{Dataset}} &
    \multicolumn{2}{c|}{\textbf{GSM8K}} &
    \multicolumn{2}{c|}{\textbf{CSQA}} &
    \multicolumn{2}{c|}{\textbf{HotPotQA}} &
    \multicolumn{2}{c|}{\textbf{InjecAgent}} &
    \multicolumn{2}{c}{\textbf{PoisonRAG}} \\
    \cmidrule{1-12}
    &\textbf{Method} &
    \textbf{F1\(\uparrow\)} & \textbf{AUROC\(\uparrow\)} &
    \textbf{F1\(\uparrow\)} & \textbf{AUROC\(\uparrow\)} &
    \textbf{F1\(\uparrow\)} & \textbf{AUROC\(\uparrow\)} &
    \textbf{F1\(\uparrow\)} & \textbf{AUROC\(\uparrow\)} &
    \textbf{F1\(\uparrow\)} & \textbf{AUROC\(\uparrow\)} \\
    \midrule
    \multicolumn{1}{c|}{\multirow{7}{*}{\textbf{level 1}}}
    & \textbf{Oracle}
    & \(26.67_{\scriptstyle \pm 0.40}\) & \(49.67\)
    & \(20.00_{\scriptstyle \pm 0.41}\) & \(52.62\)
    & \(25.56_{\scriptstyle \pm 0.92}\) & \(50.69\)
    & \(14.44_{\scriptstyle \pm 0.87}\) & \(49.27\)
    & \(27.78_{\scriptstyle \pm 0.92}\) & \(51.96\) \\
    \multicolumn{1}{c|}{} & \textbf{BlindGuard}
    & \(7.12_{\scriptstyle \pm 2.61}\) & \(65.00\)
    & \(5.88_{\scriptstyle \pm 0.14}\) & \(51.47\)
    & \(10.34_{\scriptstyle \pm 2.81}\) & \(54.05\)
    & \(12.96_{\scriptstyle \pm 0.26}\) & \(54.96\)
    & \(23.87_{\scriptstyle \pm 0.58}\) & \(54.48\) \\
    \multicolumn{1}{c|}{} & \textbf{G-Safeguard}
    & \(26.00_{\scriptstyle \pm 3.46}\) & \(46.67\)
    & \(26.67_{\scriptstyle \pm 1.89}\) & \(49.33\)
    & \(25.08_{\scriptstyle \pm 2.17}\) & \(51.67\)
    & \(25.56_{\scriptstyle \pm 0.97}\) & \(47.22\)
    & \(22.22_{\scriptstyle \pm 0.31}\) & \(55.56\) \\
    \multicolumn{1}{c|}{} & \textbf{AcMAS(-R)}
    & \underline{\(77.55_{\scriptstyle \pm 3.04}\)} & \underline{\(91.84\)}
    & \underline{\(69.46_{\scriptstyle \pm 0.97}\)} & \underline{\(90.88\)}
    & \underline{\(70.68_{\scriptstyle \pm 0.78}\)} & \(52.01\)
    & \underline{\(87.68_{\scriptstyle \pm 0.79}\)} & \underline{\(91.79\)}
    & \underline{\(70.10_{\scriptstyle \pm 0.74}\)} & \underline{\(88.30\)} \\
    \multicolumn{1}{c|}{} & \textbf{A-Trust}
    & \(68.63_{\scriptstyle \pm 1.96}\) & \(89.06\)
    & \(64.59_{\scriptstyle \pm 2.09}\) & \(84.90\)
    & \(43.81_{\scriptstyle \pm 3.81}\) & \underline{\(83.86\)}
    & \(58.20_{\scriptstyle \pm 1.06}\) & \(80.86\)
    & \(46.82_{\scriptstyle \pm 0.83}\) & \(79.82\) \\
    \multicolumn{1}{c|}{} & \textbf{AgentSafe}
    & \(68.26_{\scriptstyle \pm 2.24}\) & \(91.63\)
    & \(67.77_{\scriptstyle \pm 2.65}\) & \(88.74\)
    & \(67.97_{\scriptstyle \pm 1.97}\) & \(76.67\)
    & \(69.94_{\scriptstyle \pm 0.56}\) & \(79.04\)
    & \(69.14_{\scriptstyle \pm 0.21}\) & \(78.22\) \\
    \multicolumn{1}{c|}{} & \cellcolor[HTML]{DADADA}\textbf{Ours}
    & \cellcolor[HTML]{DADADA}\(\textbf{79.04}_{\scriptstyle \pm \textbf{3.88}}\) & \cellcolor[HTML]{DADADA}\(\textbf{96.75}\)
    & \cellcolor[HTML]{DADADA}\(\textbf{87.21}_{\scriptstyle \pm \textbf{0.36}}\) & \cellcolor[HTML]{DADADA}\(\textbf{93.29}\)
    & \cellcolor[HTML]{DADADA}\(\textbf{88.76}_{\scriptstyle \pm \textbf{0.18}}\) & \cellcolor[HTML]{DADADA}\(\textbf{88.31}\)
    & \cellcolor[HTML]{DADADA}\(\textbf{94.74}_{\scriptstyle \pm \textbf{0.35}}\) & \cellcolor[HTML]{DADADA}\(\textbf{98.96}\)
    & \cellcolor[HTML]{DADADA}\(\textbf{74.75}_{\scriptstyle \pm \textbf{0.16}}\) & \cellcolor[HTML]{DADADA}\(\textbf{93.19}\) \\
    \midrule
    \multicolumn{1}{c|}{\multirow{7}{*}{\textbf{level 2}}}
    & \textbf{Oracle}
    & \(35.56_{\scriptstyle \pm 0.84}\) & \underline{\(60.60\)}
    & \(22.22_{\scriptstyle \pm 0.92}\) & \(48.66\)
    & \(26.67_{\scriptstyle \pm 0.40}\) & \(51.44\)
    & \(24.44_{\scriptstyle \pm 1.03}\) & \underline{\(64.07\)}
    & \(22.22_{\scriptstyle \pm 0.47}\) & \(55.02\) \\
    \multicolumn{1}{c|}{} & \textbf{BlindGuard}
    & \(10.29_{\scriptstyle \pm 1.08}\) & \(54.52\)
    & \(8.73_{\scriptstyle \pm 0.16}\) & \(52.59\)
    & \(16.52_{\scriptstyle \pm 0.21}\) & \underline{\(57.48\)}
    & \(11.10_{\scriptstyle \pm 0.28}\) & \(49.48\)
    & \(9.33_{\scriptstyle \pm 0.13}\) & \(50.90\) \\
    \multicolumn{1}{c|}{} & \textbf{G-Safeguard}
    & \(21.06_{\scriptstyle \pm 0.37}\) & \(50.44\)
    & \(18.64_{\scriptstyle \pm 0.34}\) & \underline{\(55.19\)}
    & \(11.13_{\scriptstyle \pm 0.24}\) & \(47.81\)
    & \(11.19_{\scriptstyle \pm 0.21}\) & \(46.29\)
    & \(22.86_{\scriptstyle \pm 0.31}\) & \underline{\(55.19\)} \\
    \multicolumn{1}{c|}{} & \textbf{AcMAS(-R)}
    & \(23.33_{\scriptstyle \pm 0.73}\) & \(48.15\)
    & \(16.29_{\scriptstyle \pm 0.94}\) & \(44.15\)
    & \(16.89_{\scriptstyle \pm 0.92}\) & \(43.38\)
    & \(15.85_{\scriptstyle \pm 0.13}\) & \(46.81\)
    & \(26.25_{\scriptstyle \pm 0.28}\) & \(36.30\) \\
    \multicolumn{1}{c|}{} & \textbf{A-Trust}
    & \(16.03_{\scriptstyle \pm 0.64}\) & \(60.42\)
    & \(6.89_{\scriptstyle \pm 0.47}\) & \(39.58\)
    & \(10.44_{\scriptstyle \pm 0.36}\) & \(33.85\)
    & \(27.70_{\scriptstyle \pm 0.56}\) & \(50.26\)
    & \(16.21_{\scriptstyle \pm 0.49}\) & \(48.18\) \\
    \multicolumn{1}{c|}{} & \textbf{AgentSafe}
    & \underline{\(49.96_{\scriptstyle \pm 0.16}\)} & \(38.67\)
    & \underline{\(51.56_{\scriptstyle \pm 0.73}\)} & \(45.11\)
    & \underline{\(64.99_{\scriptstyle \pm 0.32}\)} & \(37.70\)
    & \underline{\(64.49_{\scriptstyle \pm 0.74}\)} & \(42.63\)
    & \underline{\(64.43_{\scriptstyle \pm 0.29}\)} & \(43.19\) \\
    \multicolumn{1}{c|}{} & \cellcolor[HTML]{DADADA}\textbf{Ours}
    & \cellcolor[HTML]{DADADA}\(\textbf{52.96}_{\scriptstyle \pm \textbf{0.18}}\) & \cellcolor[HTML]{DADADA}\(\textbf{63.84}\)
    & \cellcolor[HTML]{DADADA}\(\textbf{52.84}_{\scriptstyle \pm \textbf{0.69}}\) & \cellcolor[HTML]{DADADA}\(\textbf{63.79}\)
    & \cellcolor[HTML]{DADADA}\(\textbf{67.43}_{\scriptstyle \pm \textbf{0.82}}\) & \cellcolor[HTML]{DADADA}\(\textbf{61.28}\)
    & \cellcolor[HTML]{DADADA}\(\textbf{67.68}_{\scriptstyle \pm \textbf{0.64}}\) & \cellcolor[HTML]{DADADA}\(\textbf{70.73}\)
    & \cellcolor[HTML]{DADADA}\(\textbf{66.55}_{\scriptstyle \pm \textbf{0.76}}\) & \cellcolor[HTML]{DADADA}\(\textbf{68.28}\) \\
    \midrule
    \multicolumn{1}{c|}{\multirow{7}{*}{\textbf{level 3}}}
    & \textbf{Oracle}
    & \underline{\(46.67_{\scriptstyle \pm 0.42}\)} & \underline{\(63.90\)}
    & \(3.33_{\scriptstyle \pm 0.49}\) & \(14.69\)
    & \(11.11_{\scriptstyle \pm 0.96}\) & \(15.94\)
    & \(36.67_{\scriptstyle \pm 0.96}\) & \(55.80\)
    & \(3.33_{\scriptstyle \pm 0.28}\) & \(10.37\) \\
    \multicolumn{1}{c|}{} & \textbf{BlindGuard}
    & \(35.59_{\scriptstyle \pm 0.36}\) & \(59.26\)
    & \underline{\(33.70_{\scriptstyle \pm 4.58}\)} & \(58.72\)
    & \underline{\(47.41_{\scriptstyle \pm 0.29}\)} & \underline{\(68.37\)}
    & \(5.41_{\scriptstyle \pm 0.03}\) & \(54.74\)
    & \underline{\(35.92_{\scriptstyle \pm 0.43}\)} & \underline{\(62.67\)} \\
    \multicolumn{1}{c|}{} & \textbf{G-Safeguard}
    & \(19.86_{\scriptstyle \pm 0.94}\) & \(43.41\)
    & \(15.84_{\scriptstyle \pm 0.93}\) & \(42.13\)
    & \(16.96_{\scriptstyle \pm 0.28}\) & \(42.53\)
    & \underline{\(66.67_{\scriptstyle \pm 0.07}\)} & \(66.67\)
    & \(22.37_{\scriptstyle \pm 0.89}\) & \(47.78\) \\
    \multicolumn{1}{c|}{} & \textbf{AcMAS(-R)}
    & \(16.16_{\scriptstyle \pm 0.88}\) & \(44.59\)
    & \(12.04_{\scriptstyle \pm 1.60}\) & \(37.92\)
    & \(20.00_{\scriptstyle \pm 0.27}\) & \(44.42\)
    & \(43.48_{\scriptstyle \pm 0.18}\) & \underline{\(73.33\)}
    & \(14.58_{\scriptstyle \pm 0.92}\) & \(42.07\) \\
    \multicolumn{1}{c|}{} & \textbf{A-Trust}
    & \(1.24_{\scriptstyle \pm 1.09}\) & \(45.31\)
    & \(4.44_{\scriptstyle \pm 0.03}\) & \underline{\(67.19\)}
    & \(14.42_{\scriptstyle \pm 0.86}\) & \(63.54\)
    & \(14.07_{\scriptstyle \pm 0.98}\) & \(55.34\)
    & \(3.03_{\scriptstyle \pm 0.42}\) & \(46.35\) \\
    \multicolumn{1}{c|}{} & \textbf{AgentSafe}
    & \(30.44_{\scriptstyle \pm 0.49}\) & \(55.13\)
    & \(30.08_{\scriptstyle \pm 0.07}\) & \(57.80\)
    & \(30.00_{\scriptstyle \pm 0.36}\) & \(53.93\)
    & \(26.84_{\scriptstyle \pm 0.96}\) & \(53.04\)
    & \(2.67_{\scriptstyle \pm 1.42}\) & \(22.51\) \\
    \multicolumn{1}{c|}{} & \cellcolor[HTML]{DADADA}\textbf{Ours}
    & \cellcolor[HTML]{DADADA}\(\textbf{76.54}_{\scriptstyle \pm \textbf{0.38}}\) & \cellcolor[HTML]{DADADA}\(\textbf{72.09}\)
    & \cellcolor[HTML]{DADADA}\(\textbf{73.92}_{\scriptstyle \pm \textbf{0.47}}\) & \cellcolor[HTML]{DADADA}\(\textbf{79.46}\)
    & \cellcolor[HTML]{DADADA}\(\textbf{73.08}_{\scriptstyle \pm \textbf{0.64}}\) & \cellcolor[HTML]{DADADA}\(\textbf{74.44}\)
    & \cellcolor[HTML]{DADADA}\(\textbf{80.06}_{\scriptstyle \pm \textbf{0.02}}\) & \cellcolor[HTML]{DADADA}\(\textbf{82.59}\)
    & \cellcolor[HTML]{DADADA}\(\textbf{98.04}_{\scriptstyle \pm \textbf{0.71}}\) & \cellcolor[HTML]{DADADA}\(\textbf{74.93}\) \\
    \specialrule{1.3pt}{2.0pt}{1.0pt}
    \end{tabular}}
    \label{tab:detection}
\end{table*}

\subsection{Root-Cause Localization (Q2)}
\label{sec:q2}

To answer \textbf{Q2}, we measure Hit@1 and hop distance to the labeled initiating event $c^{\mathrm{gt}}$ under the same injections as in Q1 (Table~\ref{tab:localization}).
TRACE-MAS ranks first on both metrics on every dataset, improving average Hit@1 by \textbf{28.0}~points over the baseline with the highest average (AcMAS) and reaching \textbf{86.7}\% Hit@1 with \textbf{0.18}~hops on PoisonRAG.
On HotPotQA it attains \textbf{71.1}\% Hit@1 where the strongest activation-centric baseline remains at \textbf{23.3}\%.
The gains align with attributing failures via counterfactual violation drop $R(c)$ after contract \texttt{Restore}, instead of treating the largest downstream residual as the initiating event.
Alert-pool diagnosis dominates the non-LLM cost at this stage; Q5 (Figure~\ref{fig:efficiency}) shows that a small candidate cap $C$ preserves most of this Hit@1 while substantially reducing diagnosis wall time.

\begin{table*}[htbp]
    \centering
    \caption{Root-cause localization: Hit@1 and mean hop distance to the labeled initiating event $c^{\mathrm{gt}}$.}
\scriptsize
    \setlength{\tabcolsep}{2.3pt}
    \renewcommand{\arraystretch}{1.05}
    \resizebox{0.99\textwidth}{!}{
    \begin{tabular}{c|cc|cc|cc|cc|cc}
    \specialrule{1.5pt}{1.5pt}{1.5pt}
    \cmidrule{1-11}
    \multicolumn{1}{c|}{\textbf{Dataset}} &
    \multicolumn{2}{c|}{\textbf{GSM8K}} &
    \multicolumn{2}{c|}{\textbf{CSQA}} &
    \multicolumn{2}{c|}{\textbf{HotPotQA}} &
    \multicolumn{2}{c|}{\textbf{InjecAgent}} &
    \multicolumn{2}{c}{\textbf{PoisonRAG}} \\
    \cmidrule{1-11}
    \textbf{Method} &
    \textbf{Hit@1$\uparrow$} & \textbf{Distance$\downarrow$} &
    \textbf{Hit@1$\uparrow$} & \textbf{Distance$\downarrow$} &
    \textbf{Hit@1$\uparrow$} & \textbf{Distance$\downarrow$} &
    \textbf{Hit@1$\uparrow$} & \textbf{Distance$\downarrow$} &
    \textbf{Hit@1$\uparrow$} & \textbf{Distance$\downarrow$} \\
    \midrule
    \textbf{Oracle}
    & $34.81_{\scriptstyle \pm 0.64}$ & $1.46_{\scriptstyle \pm 0.01}$
    & $15.56_{\scriptstyle \pm 1.11}$ & $2.41_{\scriptstyle \pm 0.07}$
    & $20.37_{\scriptstyle \pm 0.68}$ & $1.47_{\scriptstyle \pm 0.03}$
    & $27.04_{\scriptstyle \pm 0.64}$ & $1.52_{\scriptstyle \pm 0.02}$
    & $23.70_{\scriptstyle \pm 1.28}$ & $1.51_{\scriptstyle \pm 0.04}$ \\
    \textbf{BlindGuard}
    & $2.31_{\scriptstyle \pm 1.89}$ & $2.33_{\scriptstyle \pm 0.02}$
    & $1.33_{\scriptstyle \pm 1.73}$ & $2.48_{\scriptstyle \pm 0.04}$
    & $1.60_{\scriptstyle \pm 1.96}$ & $2.60_{\scriptstyle \pm 0.12}$
    & $2.22_{\scriptstyle \pm 0.84}$ & $3.78_{\scriptstyle \pm 0.08}$
    & $4.45_{\scriptstyle \pm 3.14}$ & $1.17_{\scriptstyle \pm 0.62}$ \\
    \textbf{G-Safeguard}
    & $31.00_{\scriptstyle \pm 1.73}$ & \underline{$0.38_{\scriptstyle \pm 0.01}$}
    & $46.67_{\scriptstyle \pm 3.77}$ & \underline{$0.41_{\scriptstyle \pm 0.07}$}
    & \underline{$45.33_{\scriptstyle \pm 3.77}$} & $0.76_{\scriptstyle \pm 0.02}$
    & \underline{$75.55_{\scriptstyle \pm 0.38}$} & \underline{$0.40_{\scriptstyle \pm 0.17}$}
    & $31.11_{\scriptstyle \pm 0.83}$ & \underline{$0.43_{\scriptstyle \pm 0.05}$} \\
    \textbf{AcMAS}
    & \underline{$48.00_{\scriptstyle \pm 3.27}$} & $0.59_{\scriptstyle \pm 0.08}$
    & \underline{$54.67_{\scriptstyle \pm 3.77}$} & $0.44_{\scriptstyle \pm 0.03}$
    & $23.33_{\scriptstyle \pm 5.77}$ & $1.01_{\scriptstyle \pm 0.15}$
    & $55.55_{\scriptstyle \pm 3.14}$ & $1.04_{\scriptstyle \pm 0.11}$
    & \underline{$51.11_{\scriptstyle \pm 0.62}$} & $0.65_{\scriptstyle \pm 0.03}$ \\
    \textbf{A-Trust}
    & $21.67_{\scriptstyle \pm 2.89}$ & $1.05_{\scriptstyle \pm 0.21}$
    & $15.55_{\scriptstyle \pm 3.13}$ & $1.32_{\scriptstyle \pm 0.20}$
    & $15.00_{\scriptstyle \pm 2.88}$ & \underline{$0.72_{\scriptstyle \pm 0.08}$}
    & $4.38_{\scriptstyle \pm 0.30}$ & $2.75_{\scriptstyle \pm 0.27}$
    & $4.37_{\scriptstyle \pm 0.32}$ & $1.95_{\scriptstyle \pm 0.27}$ \\
    \textbf{AgentSafe}
    & $23.34_{\scriptstyle \pm 0.57}$ & $1.56_{\scriptstyle \pm 0.11}$
    & $22.22_{\scriptstyle \pm 0.62}$ & $1.62_{\scriptstyle \pm 0.03}$
    & $24.44_{\scriptstyle \pm 0.63}$ & $1.75_{\scriptstyle \pm 0.03}$
    & $22.22_{\scriptstyle \pm 3.14}$ & $1.93_{\scriptstyle \pm 0.05}$
    & $26.67_{\scriptstyle \pm 0.54}$ & $1.66_{\scriptstyle \pm 0.09}$ \\
    \cellcolor[HTML]{DADADA}\textbf{Ours}
    & \cellcolor[HTML]{DADADA}$\textbf{67.33}_{\scriptstyle \pm \textbf{0.94}}$ & \cellcolor[HTML]{DADADA}$\textbf{0.33}_{\scriptstyle \pm \textbf{0.01}}$
    & \cellcolor[HTML]{DADADA}$\textbf{68.67}_{\scriptstyle \pm \textbf{0.94}}$ & \cellcolor[HTML]{DADADA}$\textbf{0.37}_{\scriptstyle \pm \textbf{0.02}}$
    & \cellcolor[HTML]{DADADA}$\textbf{71.11}_{\scriptstyle \pm \textbf{4.16}}$ & \cellcolor[HTML]{DADADA}$\textbf{0.64}_{\scriptstyle \pm \textbf{0.04}}$
    & \cellcolor[HTML]{DADADA}$\textbf{78.89}_{\scriptstyle \pm \textbf{0.57}}$ & \cellcolor[HTML]{DADADA}$\textbf{0.29}_{\scriptstyle \pm \textbf{0.03}}$
    & \cellcolor[HTML]{DADADA}$\textbf{86.67}_{\scriptstyle \pm \textbf{0.27}}$ & \cellcolor[HTML]{DADADA}$\textbf{0.18}_{\scriptstyle \pm \textbf{0.01}}$\\
    \specialrule{1.3pt}{2.0pt}{1.0pt}
    \end{tabular}}
    \label{tab:localization}
\end{table*}

\subsection{Task-Preserving Recovery (Q3)}
\label{sec:q3}

\textbf{Repair results.}
To answer \textbf{Q3}, we evaluate post-alert recovery on attacked test trajectories under a \textbf{balanced} L1--L3 attack mix (pooled across levels, unlike the level-wise rows in Table~\ref{tab:detection}).
Table~\ref{tab:recovery} reports ASR and task ACC after TRACE-MAS commits one repair at the attributed source; TCR and mean operator cost are in Appendix Table~\ref{tab:recovery-cost} (Appendix~\ref{app:recovery-cost-table}).
TRACE-MAS achieves the best ASR and ACC on every benchmark (e.g., PoisonRAG ASR \textbf{3.0}\%, ACC \textbf{93.8}\%), and Appendix Table~\ref{tab:recovery-cost} shows the strongest TCR--cost trade-off (PoisonRAG TCR \textbf{82.9}\%, cost \textbf{0.54}), whereas retry-based recovery (AcMAS-R) and agent-level sanitization or trust-based exclusion (AgentSafe, A-Trust) lag on utility or intervention cost under the same balanced-attack protocol.
Predictive contract repair at $c^\star$ therefore restores collaboration with minimal intervention, whereas quarantining agents or patching downstream symptoms alone does not reach the same utility--cost balance.

\begin{table*}[htbp]
    \centering
    \caption{Post-defense ASR and task ACC on attacked trajectories.}
\scriptsize
    \setlength{\tabcolsep}{2.3pt}
    \renewcommand{\arraystretch}{1.05}
    \resizebox{0.99\textwidth}{!}{
    \begin{tabular}{c|cc|cc|cc|cc|cc}
    \specialrule{1.5pt}{1.5pt}{1.5pt}
    \cmidrule{1-11}
    \multicolumn{1}{c|}{\textbf{Dataset}} &
    \multicolumn{2}{c|}{\textbf{GSM8K}} &
    \multicolumn{2}{c|}{\textbf{CSQA}} &
    \multicolumn{2}{c|}{\textbf{HotPotQA}} &
    \multicolumn{2}{c|}{\textbf{InjecAgent}} &
    \multicolumn{2}{c}{\textbf{PoisonRAG}} \\
    \cmidrule{1-11}
    \textbf{Method} &
    \textbf{ASR$\downarrow$} & \textbf{ACC$\uparrow$} & 
    \textbf{ASR$\downarrow$} & \textbf{ACC$\uparrow$} & 
    \textbf{ASR$\downarrow$} & \textbf{ACC$\uparrow$} & 
    \textbf{ASR$\downarrow$} & \textbf{ACC$\uparrow$} & 
    \textbf{ASR$\downarrow$} & \textbf{ACC$\uparrow$} \\
    \midrule
    \textbf{Oracle}
    & $65.19_{\scriptstyle \pm 0.62}$ & $47.78_{\scriptstyle \pm 0.40}$
    & $84.44_{\scriptstyle \pm 1.11}$ & $31.85_{\scriptstyle \pm 0.64}$
    & $79.63_{\scriptstyle \pm 0.56}$ & $44.81_{\scriptstyle \pm 0.53}$
    & $71.48_{\scriptstyle \pm 0.64}$ & $62.22_{\scriptstyle \pm 0.50}$
    & $76.30_{\scriptstyle \pm 0.29}$ & $23.70_{\scriptstyle \pm 0.29}$ \\
    \textbf{BlindGuard}
    & $77.19_{\scriptstyle \pm 0.16}$ & $36.67_{\scriptstyle \pm 0.47}$
    & $85.39_{\scriptstyle \pm 0.67}$ & $35.33_{\scriptstyle \pm 0.47}$
    & $83.63_{\scriptstyle \pm 1.12}$ & $46.13_{\scriptstyle \pm 0.75}$
    & $90.69_{\scriptstyle \pm 0.36}$ & $44.89_{\scriptstyle \pm 1.66}$
    & $94.77_{\scriptstyle \pm 1.12}$ & $30.56_{\scriptstyle \pm 0.79}$ \\
    \textbf{G-Safeguard}
    & $85.85_{\scriptstyle \pm 0.38}$ & $32.67_{\scriptstyle \pm 0.43}$
    & $85.84_{\scriptstyle \pm 1.30}$ & $28.33_{\scriptstyle \pm 1.70}$
    & $85.82_{\scriptstyle \pm 0.99}$ & $37.78_{\scriptstyle \pm 1.26}$
    & $75.77_{\scriptstyle \pm 0.30}$ & $51.11_{\scriptstyle \pm 1.26}$
    & $89.46_{\scriptstyle \pm 2.83}$ & $28.33_{\scriptstyle \pm 2.72}$ \\
    \textbf{AcMAS-R}
    & $61.73_{\scriptstyle \pm 0.18}$ & $50.00_{\scriptstyle \pm 0.12}$
    & $51.14_{\scriptstyle \pm 0.33}$ & $52.67_{\scriptstyle \pm 1.25}$
    & $82.92_{\scriptstyle \pm 1.06}$ & $53.07_{\scriptstyle \pm 1.36}$
    & $56.12_{\scriptstyle \pm 1.25}$ & $63.11_{\scriptstyle \pm 1.66}$
    & $54.21_{\scriptstyle \pm 1.13}$ & $50.42_{\scriptstyle \pm 0.72}$ \\
    \textbf{A-Trust}
    & $67.15_{\scriptstyle \pm 1.81}$ & $51.04_{\scriptstyle \pm 1.47}$
    & $74.27_{\scriptstyle \pm 0.51}$ & $44.79_{\scriptstyle \pm 1.48}$
    & $95.84_{\scriptstyle \pm0.42 }$ & $45.00_{\scriptstyle \pm 0.41}$
    & $76.97_{\scriptstyle \pm 0.46}$ & $49.17_{\scriptstyle \pm 0.31}$
    & $68.54_{\scriptstyle \pm 0.24}$ & $51.67_{\scriptstyle \pm 1.18}$ \\
    \textbf{AgentSafe}
    & \underline{$6.05_{\scriptstyle \pm 0.99}$} & \underline{$83.89_{\scriptstyle \pm 0.31}$}
    & \underline{$38.63_{\scriptstyle \pm 1.16}$} & \underline{$83.34_{\scriptstyle \pm 0.24}$}
    & \underline{$23.81_{\scriptstyle \pm 1.13}$} & \underline{$73.33_{\scriptstyle \pm 1.09}$}
    & \underline{$4.55_{\scriptstyle \pm 0.15}$} & $\underline{70.22_{\scriptstyle \pm 0.27}}$
    & \underline{$4.67_{\scriptstyle \pm 0.17}$} & \underline{$75.11_{\scriptstyle \pm 0.74}$} \\
    \cellcolor[HTML]{DADADA}\textbf{Ours}
    & \cellcolor[HTML]{DADADA}$\textbf{3.33}_{\scriptstyle \pm \textbf{0.25}}$ & \cellcolor[HTML]{DADADA}$\textbf{96.00}_{\scriptstyle \pm \textbf{0.18}}$
    & \cellcolor[HTML]{DADADA}$\textbf{29.33}_{\scriptstyle \pm \textbf{1.16}}$ & \cellcolor[HTML]{DADADA}$\textbf{84.62}_{\scriptstyle \pm \textbf{0.42}}$
    & \cellcolor[HTML]{DADADA}$\textbf{18.42}_{\scriptstyle \pm \textbf{1.17}}$ & \cellcolor[HTML]{DADADA}$\textbf{96.67}_{\scriptstyle \pm \textbf{0.27}}$
    & \cellcolor[HTML]{DADADA}$\textbf{2.98}_{\scriptstyle \pm \textbf{0.24}}$ & \cellcolor[HTML]{DADADA}$\textbf{82.89}_{\scriptstyle \pm \textbf{0.41}}$
    & \cellcolor[HTML]{DADADA}$\textbf{3.02}_{\scriptstyle \pm \textbf{0.82}}$ & \cellcolor[HTML]{DADADA}$\textbf{93.76}_{\scriptstyle \pm \textbf{0.49}}$ \\
    \specialrule{1.3pt}{2.0pt}{1.0pt}
    \end{tabular}}
    \label{tab:recovery}
\end{table*}

\textbf{Ablation study.}
\label{sec:ablation-study}
Building on the recovery results, we ablate the key designs of TRACE-MAS in Table~\ref{tab:ablation}, retraining each variant end to end on benign data with the same schedule as the full model.
We replace the shared relation-aware contract with independent per-agent encoders, substitute absolute-embedding scoring for transition residuals, remove contract-based violation alerting, and replace rollout-based repair ranking with fixed operators.
The resulting rows isolate how unified execution modeling, residual transition contracts, the alert--attribute loop, and predictive repair selection contribute to the detection, localization, and recovery trends reported in Q1--Q3.

\begin{table}[htbp]
    \centering
    \caption{Ablation---one core module removed per row, each retrained end to end as Full and evaluated on balanced L1--L3 pooled splits.}
    \small
    \setlength{\tabcolsep}{4.1pt}
    \renewcommand{\arraystretch}{1.08}
    \resizebox{\columnwidth}{!}{%
    \begin{tabular}{c|ccccc|ccccc|ccccc}
    \toprule
    \multirow{2}{*}{\textbf{Variant}} &
    \multicolumn{5}{c|}{\textbf{Detect(F1)}} &
    \multicolumn{5}{c|}{\textbf{Locate(Hit@1)}} &
    \multicolumn{5}{c}{\textbf{Repair(TCR)}} \\
    \cmidrule(lr){2-6} \cmidrule(lr){7-11} \cmidrule(lr){12-16}
    &
    \textbf{GSM8K} & \textbf{CSQA} & \textbf{HotPotQA} & \textbf{InjecAgent} & \textbf{PoisonRAG} & \textbf{GSM8K} & \textbf{CSQA} & \textbf{HotPotQA} & \textbf{InjecAgent} & \textbf{PoisonRAG} & \textbf{GSM8K} & \textbf{CSQA} & \textbf{HotPotQA} & \textbf{InjecAgent} & \textbf{PoisonRAG} \\
    \midrule
    \cellcolor[HTML]{DADADA}\textbf{Full TRACE-MAS} 
    & \cellcolor[HTML]{DADADA}$\textbf{70.58}_{\scriptstyle \pm \textbf{0.37}}$ & \cellcolor[HTML]{DADADA}$\textbf{73.04}_{\scriptstyle \pm \textbf{0.44}}$ & \cellcolor[HTML]{DADADA}$\textbf{70.34}_{\scriptstyle \pm \textbf{0.96}}$ & \cellcolor[HTML]{DADADA}$\textbf{67.29}_{\scriptstyle \pm \textbf{1.05}}$ & \cellcolor[HTML]{DADADA}$\textbf{73.97}_{\scriptstyle \pm \textbf{0.29}}$ & \cellcolor[HTML]{DADADA}$\textbf{67.38}_{\scriptstyle \pm \textbf{0.54}}$ & \cellcolor[HTML]{DADADA}$\textbf{68.29}_{\scriptstyle \pm \textbf{0.22}}$ & \cellcolor[HTML]{DADADA}$\textbf{71.47}_{\scriptstyle \pm \textbf{0.90}}$ & \cellcolor[HTML]{DADADA}$\textbf{78.65}_{\scriptstyle \pm \textbf{0.58}}$ & \cellcolor[HTML]{DADADA}$\textbf{88.50}_{\scriptstyle \pm \textbf{0.66}}$ & \cellcolor[HTML]{DADADA}$\textbf{95.93}_{\scriptstyle \pm \textbf{0.67}}$ & \cellcolor[HTML]{DADADA}$\textbf{76.60}_{\scriptstyle \pm \textbf{0.37}}$ & \cellcolor[HTML]{DADADA}$\textbf{81.67}_{\scriptstyle \pm \textbf{0.79}}$ & \cellcolor[HTML]{DADADA}$\textbf{73.25}_{\scriptstyle \pm \textbf{0.27}}$ & \cellcolor[HTML]{DADADA}$\textbf{82.00}_{\scriptstyle \pm \textbf{0.36}}$\\
    \textbf{Independent Contract}
    & $60.77_{\scriptstyle \pm 0.84}$ & $62.09_{\scriptstyle \pm 0.75}$ & $60.16_{\scriptstyle \pm 0.74}$ & $59.32_{\scriptstyle \pm 0.57}$ & $62.71_{\scriptstyle \pm 0.89}$ & $51.17_{\scriptstyle \pm 0.53}$ & $53.37_{\scriptstyle \pm 0.25}$ & $56.93_{\scriptstyle \pm 0.93}$ & $44.68_{\scriptstyle \pm 0.69}$ & $53.88_{\scriptstyle \pm 0.27}$ & $55.14_{\scriptstyle \pm 0.77}$ & $55.79_{\scriptstyle \pm 2.05}$ & $53.23_{\scriptstyle \pm 2.20}$ & $39.16_{\scriptstyle \pm 0.72}$ & $55.47_{\scriptstyle \pm 1.02}$\\
    \textbf{w/o Transition Contract}
    & $63.69_{\scriptstyle \pm 0.24}$ & $65.95_{\scriptstyle \pm 0.20}$ & $63.50_{\scriptstyle \pm 0.88}$ & $60.82_{\scriptstyle \pm 1.09}$ & $66.72_{\scriptstyle \pm 0.23}$ & $54.93_{\scriptstyle \pm 0.34}$ & $58.10_{\scriptstyle \pm 0.62}$ & $61.49_{\scriptstyle \pm 0.53}$ & $48.20_{\scriptstyle \pm 0.41}$ & $58.50_{\scriptstyle \pm 0.37}$ & $60.18_{\scriptstyle \pm 0.77}$ & $61.24_{\scriptstyle \pm 0.30}$ & $58.26_{\scriptstyle \pm 0.46}$ & $42.37_{\scriptstyle \pm 0.87}$ & $60.94_{\scriptstyle \pm 0.42}$\\
    \textbf{w/o Violation}
    & $67.86_{\scriptstyle \pm 0.31}$ & $70.23_{\scriptstyle \pm 0.34}$ & $67.63_{\scriptstyle \pm 0.93}$ & $64.70_{\scriptstyle \pm 1.01}$ & $71.12_{\scriptstyle \pm 0.22}$ & $60.87_{\scriptstyle \pm 0.75}$ & $63.98_{\scriptstyle \pm 0.73}$ & $67.82_{\scriptstyle \pm 0.45}$ & $52.72_{\scriptstyle \pm 0.96}$ & $64.58_{\scriptstyle \pm 0.55}$ & \underline{$68.35_{\scriptstyle \pm 1.01}$} & $68.66_{\scriptstyle \pm 0.39}$ & \underline{$70.03_{\scriptstyle \pm 0.40}$} & $46.79_{\scriptstyle \pm 0.96}$ & \underline{$72.98_{\scriptstyle \pm 0.54}$}\\
    \textbf{w/o Predictive Repair}
    & \underline{$69.22_{\scriptstyle \pm 0.34}$} & \underline{$71.63_{\scriptstyle \pm 0.39}$} & \underline{$68.99_{\scriptstyle \pm 0.94}$} & \underline{$65.99_{\scriptstyle \pm 1.03}$} & \underline{$72.55_{\scriptstyle \pm 0.26}$} & \underline{$62.62_{\scriptstyle \pm 0.50}$} & \underline{$66.40_{\scriptstyle \pm 0.14}$} & \underline{$69.52_{\scriptstyle \pm 0.82}$} & \underline{$75.12_{\scriptstyle \pm 0.52}$} & \underline{$86.61_{\scriptstyle \pm 0.61}$} & $67.81_{\scriptstyle \pm 0.76}$ & \underline{$69.03_{\scriptstyle \pm 0.61}$} & $65.63_{\scriptstyle \pm 0.77}$ & \underline{$47.66_{\scriptstyle \pm 0.11}$} & $68.75_{\scriptstyle \pm 0.86}$\\
    \bottomrule
    \end{tabular}}
    \label{tab:ablation}
\end{table}

\subsection{Calibration Convergence and Sensitivity (Q4)}
\label{sec:q4}

To answer \textbf{Q4}, we study benign-only calibration, standard hyperparameter choices, and deployment stress factors.

\begin{figure}[htbp]
    \centering
    \begin{subfigure}[b]{0.24\textwidth}
        \centering
        \includegraphics[width=\linewidth]{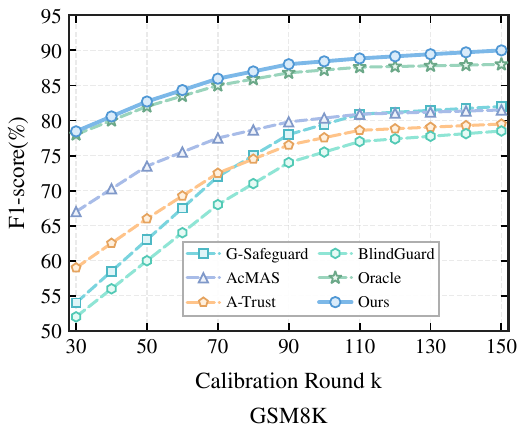}
        \caption{GSM8K, F1}
        \label{fig:q4a}
    \end{subfigure}
    \hfill
    \begin{subfigure}[b]{0.24\textwidth}
        \centering
        \includegraphics[width=\linewidth]{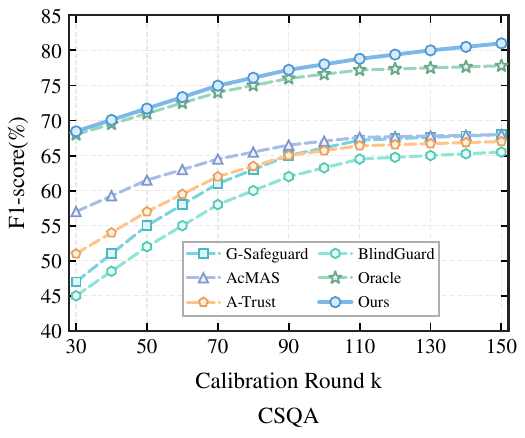}
        \caption{CSQA, F1}
        \label{fig:q4b}
    \end{subfigure}
    \hfill
    \begin{subfigure}[b]{0.24\textwidth}
        \centering
        \includegraphics[width=\linewidth]{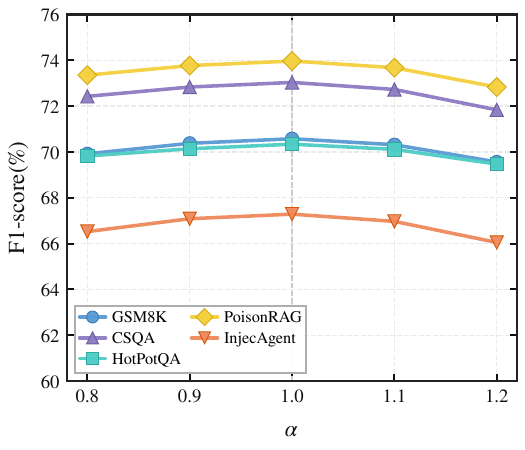}
        \caption{F1 vs.\ $\alpha$}
        \label{fig:q4c}
    \end{subfigure}
    \hfill
    \begin{subfigure}[b]{0.24\textwidth}
        \centering
        \includegraphics[width=\linewidth]{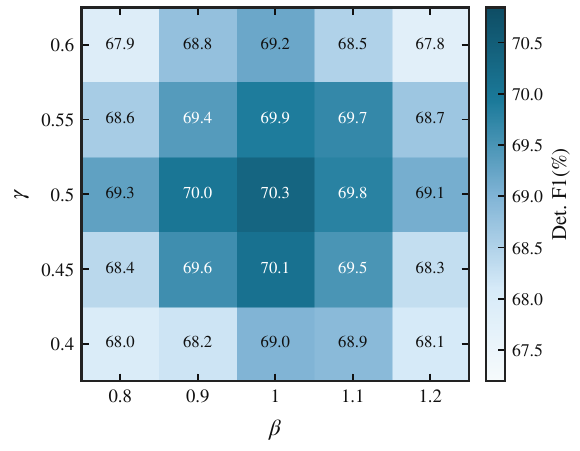}
        \caption{F1 vs.\ $\beta$ and $\gamma$}
        \label{fig:q4d}
    \end{subfigure}

    \caption{Benign calibration and hyperparameter sensitivity.}
    \label{fig:q4}
\end{figure}

\textbf{Benign-only calibration.}
We fit contracts and refresh the alert threshold $\tau$ from held-out benign trajectories only ($k\in[30,150]$ calibration rounds; no attack labels for TRACE-MAS).
Fig.~\ref{fig:q4}(a--b) reports detection F1 on GSM8K and CSQA: TRACE-MAS gains F1 faster and saturates higher than interaction-graph, activation-centric, and trust-centric monitors.

\textbf{Threshold scale $\alpha$.}
We perform a hyperparameter sensitivity sweep on the alert scale $\alpha$ in $\tau=\alpha\cdot\mathrm{Quantile}_\delta(\cdot)$ over $\alpha\in\{0.8,0.9,1.0,1.1,1.2\}$, with $(\beta,\gamma)$ fixed at their defaults and contracts calibrated as in the main protocol.
Fig.~\ref{fig:q4}(c) plots trajectory-level detection F1 against $\alpha$, with \textbf{one line per benchmark} ; all curves peak at the default $\alpha{=}1$ and change by only about \textbf{1--2}~F1 points across the sweep, indicating mild sensitivity to threshold scaling on every workflow.

\textbf{Contract weights $(\beta,\gamma)$.}
We perturb the message- and completion-violation weights around $(\beta,\gamma)=(1,0.5)$ on HotPotQA benign validation.
Fig.~\ref{fig:q4}(d) exhibits a wide high-F1 region near the default, indicating modest sensitivity to these weights in standard settings.

\textbf{Deployment stress tests.}
We further stress-test detection under alert delay, agent scale, dynamic edges, and execution imbalance; full curves and baseline comparisons are in Figs.~\ref{fig:app-delay}--\ref{fig:topo-loc}.

Together, the calibration and sensitivity panels in Fig.~\ref{fig:q4} support fitting TRACE-MAS from benign logs with a quantile threshold on $\max_t V(e_t)$, without attack labels during threshold selection.

\subsection{Runtime Overhead (Q5)}
\label{sec:q5}

\begin{figure}[htbp]
    \centering
    \begin{subfigure}[b]{0.32\textwidth}
        \centering
        \includegraphics[width=\linewidth]{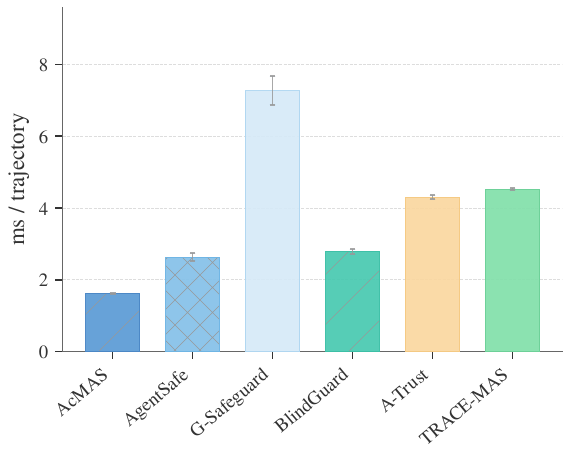}
        \caption{Detection}
        \label{fig:efficiency_detection}
    \end{subfigure}
    \hfill
    \begin{subfigure}[b]{0.32\textwidth}
        \centering
    \includegraphics[width=\linewidth]{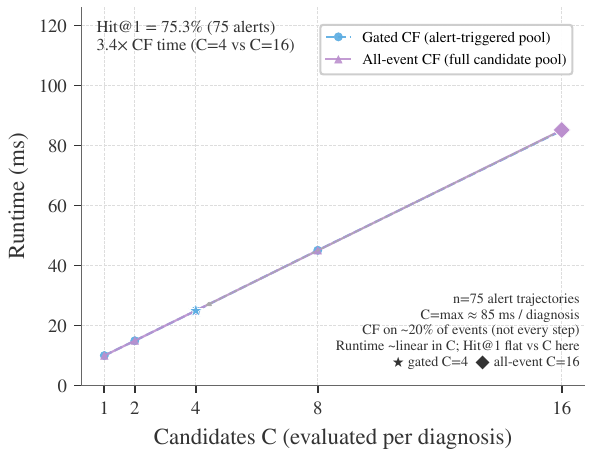}
        \caption{Diagnosis, C}
        \label{fig:efficiency_diagnosis}
    \end{subfigure}
    \hfill
    \begin{subfigure}[b]{0.32\textwidth}
        \centering
        \includegraphics[width=\linewidth]{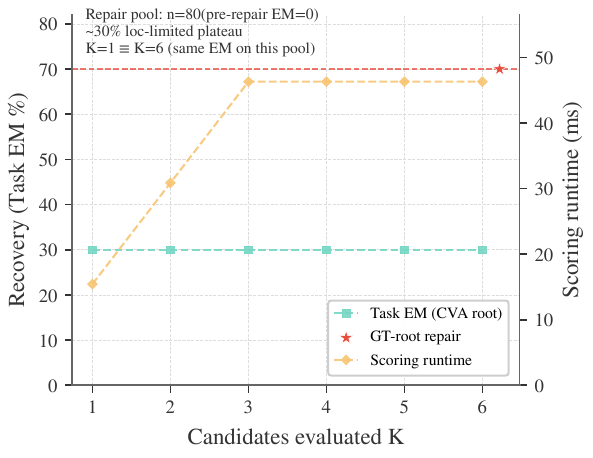}
        \caption{Repair, K}
        \label{fig:efficiency_repair}
    \end{subfigure}

    \caption{Runtime overhead of TRACE-MAS.}
    \label{fig:efficiency}
\end{figure}

To answer \textbf{Q5}, we profile the non-LLM wall-clock cost of contract alerting, gated counterfactual diagnosis on the alert pool, and rollout-based repair scoring in Fig.~\ref{fig:efficiency} (bulk LLM generation is excluded because it dominates standard MAS rollouts and is invoked only for committed repairs).

\textbf{Detection-stage overhead.}
Fig.~\ref{fig:efficiency}(a) breaks down per-trajectory contract scoring and graph maintenance; this stage is lightweight relative to downstream diagnosis because it reuses cached event features from logging.

\textbf{Alert-pool diagnosis.}
We vary the counterfactual candidate cap $C$ on the alert-triggered pool.
Fig.~\ref{fig:efficiency}(b) shows that $C{=}4$ retains \textbf{75.3}\% Hit@1 while cutting diagnosis time by about \textbf{3.4$\times$} versus $C{=}16$, matching the gated pool design used after contract alerts in Q2.

\textbf{Repair-pool scoring.}
We vary rollout depth $K$ when ranking repair operators on the repair pool.
Fig.~\ref{fig:efficiency}(c) shows that TCR on the attack-successful repair pool of Q3 is unchanged from $K{=}1$ to $K{=}6$ while scoring cost grows with $K$, so we use $K{=}1$ (Appendix~\ref{app:method-hparams}).

\section{Conclusion}
\label{sec:conclusion}

In this work, we study collaboration failures in LLM-based multi-agent
systems from an execution-centric perspective. We reveal that the attacked
component, observed anomaly, and optimal repair target may not coincide during
collaborative execution. To address this gap, we propose TRACE-MAS, which
learns benign execution contracts and unifies violation detection,
counterfactual root-cause attribution, and task-preserving recovery within a
shared execution model. Experiments across diverse MAS benchmarks demonstrate
that TRACE-MAS achieves more accurate failure diagnosis and effective recovery with minimal intervention. Future work will investigate adaptive execution contracts under open-ended environments where collaboration structures and agent behaviors continuously evolve.


\section*{AI Usage Disclosure}

Generative AI tools were used during the preparation of this manuscript for limited purposes, including language refinement, literature retrieval and discovery, and research ideation support. Specifically, AI tools assisted with improving writing clarity, organizing related work, exploring potential research directions, and discussing possible methodological designs and experimental considerations.

\bibliography{iclr2027_conference}
\bibliographystyle{iclr2027_conference}

\appendix

\section{Methodology in Detail}
\label{app:method}

This appendix expands the method of the main text. It keeps the same pipeline and the same
symbols, and adds the parts that do not fit in the main paper: how execution logs become a typed
event graph, how textual fields become transition features, how the execution contract is trained
and calibrated, how candidate sources are constructed and scored, how repairs are ranked, and what
each stage costs. Sec.~\ref{app:pipeline} summarizes the end-to-end procedure, and
Sec.~\ref{app:method-hparams} lists every default used in the main tables. Analytical support for
the design choices is in Appendix~\ref{app:theory}; shared experimental settings are in
Appendix~\ref{app:experiment}.

Table~\ref{tab:notation} collects the symbols used throughout.

\begin{table}[H]
\centering
\small
\caption{Notation used in Appendix~A and referenced by the main text.}
\begin{tabular}{ll}
\toprule
Symbol & Meaning \\
\midrule
$e_t=(i_t,r_t,q_t,x_t,h_t,m^{\mathrm{in}}_t,m^{\mathrm{out}}_t,a_t,s_t)$ & event of Sec.~2: agent, role, subtask, context, state, messages, availability, progress \\
$G=(\mathcal{V},E)$, $\tilde{G}$ & event node set and edges; $\tilde{G}$ is the observed (attacked) graph \\
$\rho_{ij}\in\{\rho_{\mathrm{msg}},\rho_{\mathrm{dep}},\rho_{\mathrm{temp}}\}$ & message / task-dependency / temporal edge type \\
$\mathcal{H}_t$, $c_t$ & ancestor neighborhood of $e_t$, encoded execution context \\
$z^0_t$, $z_t$, $\Delta e_t$ & initial node features, contract-space embedding, transition residual \\
$p_\theta(\Delta e_t\mid c_t)$, $\mathcal{L}_{\mathrm{con}}$ & execution contract $\mathcal{N}(\mu_t,\Sigma_t)$ and its benign training loss \\
$\widetilde{\Delta e}_t$ & transition observed in $\tilde{G}$ (scored against $\mu_t$) \\
$V(e_t)$, $\tau$, $\Omega$ & violation score, alert threshold, violated event set \\
$\varphi_t$ & contract hidden state; the transition itself is $\Delta e_t$ ($=o_t$ of Sec.~2) \\
$\mathcal{C}$, $\mathrm{Anc}_H$, $\mathrm{Desc}(c)$ & candidate sources, $H$-hop ancestors, strict descendants of $c$ (excluding $c$) \\
$V^{\mathrm{res}}_c(\cdot)$, $R(c)$, $c^\star$ & score after $\mathrm{Restore}_c$, responsibility, attributed source \\
$c^{\mathrm{gt}}$ & labeled initiating event (evaluation only; never used by the method) \\
$\mathcal{U}(c^\star)$, $u^\star$, $\kappa(u)$ & repair operators, selected repair, intervention cost \\
$\delta,\alpha$ & benign quantile level and threshold scale in $\tau$ \\
$\omega,\psi,\lambda_{\mathrm{pen}}$ & restored-evidence gate, hop decay, validity penalty weight \\
$\beta,\gamma$ & message- and completion-violation weights inside $V$ (Fig.~\ref{fig:q4}(d)) \\
$H,C,K$ & ancestor hop limit, scored candidates, rollout depth \\
$\hat{e}^u_i$, $\hat{\mathcal{J}}(u)$, $\mathcal{J}(u)$ & rolled-out event, rollout estimate and realized value of the repair objective \\
\bottomrule
\end{tabular}
\label{tab:notation}
\end{table}

\subsection{Execution Graph Construction}
\label{app:graph}

A multi-agent execution is a sequence of interleaved agent steps, messages, and task updates.
\textsc{Trace-Mas} represents one trajectory as a typed event graph so that a single
representation covers both local agent behavior and cross-agent coordination.

Each execution event is recorded as
\begin{equation}
e_t = \bigl(i_t, r_t, q_t, x_t, h_t, m^{\mathrm{in}}_t, m^{\mathrm{out}}_t, a_t, s_t\bigr),
\label{eq:app-event}
\end{equation}
which is the event of Sec.~2 field by field: $i_t$ is the executing agent, $r_t$ its role in the
workflow template, $q_t$ the subtask it works on, $x_t$ the execution context visible to the step
(task input and received content), $h_t$ the agent state (prompt, memory, and scratchpad content),
$m^{\mathrm{in}}_t$ and $m^{\mathrm{out}}_t$ the consumed and emitted message or tool payloads, $a_t$ the
agent's availability flag, and $s_t$ the task-progress summary after the step. Every field is read
from the log of the unmodified benchmark workflow; we do not add instrumentation to the agents
themselves.

Events become nodes of a partial-order graph
\begin{equation}
G = (\mathcal{V}, E),
\end{equation}
and each edge carries a relation type
\begin{equation}
\rho_{ij} \in \{\rho_{\mathrm{msg}}, \rho_{\mathrm{dep}}, \rho_{\mathrm{temp}}\}.
\end{equation}
The three types are built by fixed rules rather than learned: a $\rho_{\mathrm{msg}}$ edge connects
the event that emits a message (or tool result) to the event that consumes it; a
$\rho_{\mathrm{dep}}$ edge connects events whose tasks are linked by the workflow template (for
example a verifier step that depends on a solver output); and a $\rho_{\mathrm{temp}}$ edge connects
consecutive events of the same agent. Edges always point from an earlier producer to a later
consumer, so $G$ is a directed acyclic graph and defines a partial order over events.

Two properties follow from this construction and are used later. First, the representation never
assumes lock-step rounds: two events are ordered only when a message, a task dependency, or the same
agent's own history links them, which is why delayed or asynchronous execution does not change the
graph semantics (Appendix~\ref{app:robustness}). Second, the three corruption levels of our threat
model touch different parts of the same object: L1 changes $h_t$ on a node, L2 changes
$m^{\mathrm{out}}_t$ on a $\rho_{\mathrm{msg}}$ edge payload (and hence $m^{\mathrm{in}}$ of its
consumer), and L3 changes the $\rho_{\mathrm{dep}}$ structure itself. A
monitor defined on this graph can therefore express all three, whereas agent-level state monitors
observe only the first.

The initial node representation concatenates the embedded execution factors,
\begin{equation}
z^0_t = [f_i(i_t); f_r(r_t); f_q(q_t); f_a(a_t); f_x(x_t); f_h(h_t); f_m(m^{\mathrm{in}}_t, m^{\mathrm{out}}_t); f_s(s_t)],
\label{eq:app-z0}
\end{equation}
where each $f(\cdot)$ embeds the corresponding component. Structured fields ($i_t$, $r_t$, $q_t$,
$a_t$) use learnable embedding tables, and textual fields ($x_t$, $h_t$, $m^{\mathrm{in}}_t$,
$m^{\mathrm{out}}_t$, $s_t$) are mapped into
the contract space as described in Sec.~\ref{app:features}. When several events share a timestamp,
parsing uses deterministic tie-breaking (agent id, then log line number) so that the graph, and
hence every score below, is reproducible across runs.

\subsection{Relation-Aware Execution Context Encoder}
\label{app:encoder}

The contract must predict what an event \emph{should} look like given the collaboration that
preceded it. \textsc{Trace-Mas} therefore encodes, for each event, the execution context formed by
its ancestors and the relation types through which they act.

The neighborhood of $e_t$ is
\begin{equation}
\mathcal{H}_t = \{ e_j \mid (e_j, e_t) \in E \},
\end{equation}
that is, the direct parents of $e_t$ in the partial order. Because $E$ only contains
producer-to-consumer edges, $\mathcal{H}_t$ never includes events that occur after $e_t$; the
encoder is causal by construction and cannot use post-hoc information when scoring a step.

Each neighbor is transformed with parameters tied to its relation type,
\begin{equation}
k_{jt} = W^{\rho_{jt}}_k z^0_j, \qquad
v_{jt} = W^{\rho_{jt}}_v z^0_j,
\end{equation}
so that a corrupted message and a corrupted dependency enter the context through different
projections. With the query $\tilde{q}_t = W_q z^0_t$, attention weights over the neighborhood are
\begin{equation}
\pi_{jt} = \frac{\exp\bigl( \tilde{q}_t^\top k_{jt} / \sqrt{d} \bigr)}{\sum_{l \in \mathcal{H}_t} \exp\bigl( \tilde{q}_t^\top k_{lt} / \sqrt{d} \bigr)},
\end{equation}
and the context representation is
\begin{equation}
c_t = \sigma\!\left( W_o \sum_{e_j \in \mathcal{H}_t} \pi_{jt} v_{jt} + W_s z^0_t \right).
\end{equation}
The residual term $W_s z^0_t$ keeps the event's own identity (agent, role, task) in the context, so
the contract is conditioned on \emph{who} is acting as well as on what it received.

Relation-specific projections matter for our setting rather than being a generic design choice: the
expected transition of a verifier that just received a retrieved document differs from that of the
same agent following its own previous step, even when the surface text is similar. A single
untyped aggregation would average these regimes and inflate the benign variance, which directly
weakens the score in Eq.~(\ref{eq:app-violation}).

In implementation, one parent-event multi-head attention layer (4 heads, hidden dimension $d{=}256$)
realizes this aggregation; role and relation types use 64-dimensional learnable embeddings over nine
role ids and $\{\rho_{\mathrm{msg}},\rho_{\mathrm{dep}},\rho_{\mathrm{temp}}\}$
(Table~\ref{tab:execformer}). Encoding one trajectory costs $O(|E|\,d + |\mathcal{V}|\,d^2)$ time and
$O(|\mathcal{V}|\,d)$ memory, and events with disjoint ancestor sets are batched together.

\subsection{Text-to-Transition Featureization}
\label{app:features}

Logs are tokenized per event. Let $\phi_{\text{text}}$ be a frozen \texttt{all-MiniLM-L6-v2} Sentence-BERT \citep{reimers2019sentence}
encoder (384-dimensional pooled vectors) applied to the concatenated agent text of the event
(thought, inbound and outbound message snippets, and task-state summary), followed by a learned
linear map into the 256-dimensional contract space. Writing $z_t \in \mathbb{R}^{256}$ for the
projected embedding of the active fields at $e_t$, the quantity modeled by the contract is the
transition residual
\begin{equation}
\Delta e_t = z_t - z_{t-1},
\end{equation}
where $z_{t-1}$ is the embedding of the immediately preceding state of the same execution thread.

Modeling the residual rather than the absolute embedding is what makes benign-only training usable
across benchmarks: absolute embeddings differ strongly between, say, a math scratchpad and a
retrieved passage, while the \emph{change} induced by one legitimate step is far more regular under
a fixed role and relation type. The frozen encoder plays the same stabilizing role. Because it is
never updated, the geometry of the feature space does not shift between the benign fitting split and
the attacked test split, and no attack supervision can leak into the representation.

Three two-layer MLP heads (state, message, completion) predict the channel-wise mean $\mu_t$, and
the Gaussian head operates on the 256-dimensional residual block with a diagonal $\Sigma_t$.
Following the main text, $\Delta e_t$ denotes the transition variable of the contract,
$\widetilde{\Delta e}_t$ its value observed in $\tilde{G}$, and $\mu_t$ the contract mean. At
inference the implementation shrinks the observed residual toward the contract mean,
\begin{equation}
\bar{\Delta e}_t = \gamma_{\mathrm{res}}\, \widetilde{\Delta e}_t + \eta_{\mathrm{res}}\, \mu_t,
\qquad (\gamma_{\mathrm{res}}, \eta_{\mathrm{res}}) = (0.5, 0.5),
\label{eq:app-shrink}
\end{equation}
before Mahalanobis scoring. Because $\gamma_{\mathrm{res}}+\eta_{\mathrm{res}}=1$, this gives
$\bar{\Delta e}_t-\mu_t=\gamma_{\mathrm{res}}(\widetilde{\Delta e}_t-\mu_t)$, so the scored value is
exactly $\gamma_{\mathrm{res}}^2\,V(e_t)$ with $V$ as in Eq.~(\ref{eq:app-violation}). Since $\tau$ is
a quantile of the same scaled score, every alert, the set $\Omega$, and every ranking of events are
identical to those obtained from $V(e_t)$ itself; the shrinkage only fixes the numerical scale of the
violation terms relative to $\lambda_{\mathrm{pen}}$, $\lambda_1$, and $\lambda_2$, whose defaults
(Sec.~\ref{app:method-hparams}) were set on this scale. All statements below and in
Appendix~\ref{app:theory} are therefore written for $V(e_t)$. The coefficient $\gamma_{\mathrm{res}}$
is distinct both from the completion-violation weight $\gamma$ of Sec.~\ref{app:detection} and from
the attribution hop decay $\psi$ of Sec.~\ref{app:attribution-resp}.

\subsection{Probabilistic Execution Contract Learning}
\label{app:contract}

Given the context $c_t$, the contract states how a normal event evolves under that collaboration
condition. The transition is decomposed by channel,
\begin{equation}
\Delta e_t = [\Delta h_t; \Delta m_t; \Delta s_t],
\end{equation}
where $\Delta h_t$, $\Delta m_t$, and $\Delta s_t$ are the changes in agent state, message
interaction, and task progress. The decomposition is deliberate: it gives each corruption level a
channel in which it is expressed, so a single scalar score can react to state edits, payload edits,
and coordination edits without a separate detector per attack type.

A transition network predicts the expected evolution from a shared hidden state $\varphi_t$ (kept
distinct from the execution transition $o_t$ of Sec.~2, which is the same object as $\Delta e_t$),
\begin{equation}
\varphi_t = \mathrm{MLP}_\theta(c_t), \qquad
\mu_t = [g_h(\varphi_t); g_m(\varphi_t); g_s(\varphi_t)],
\end{equation}
and, because legitimate executions are themselves variable, the contract also predicts a spread
\begin{equation}
\Sigma_t = \mathrm{Softplus}(W_\Sigma \varphi_t),
\end{equation}
kept diagonal and floored away from zero for numerical stability. The contract is therefore
\begin{equation}
p_\theta(\Delta e_t \mid c_t) = \mathcal{N}(\mu_t, \Sigma_t).
\label{eq:app-contract}
\end{equation}

Predicting $\Sigma_t$ is what allows one threshold to serve heterogeneous roles. An agent that
speaks rarely and an agent that speaks at every round have different benign transition spreads; with
a shared fixed variance, the busier agent would accumulate large raw deviations and dominate the
alerts. Conditioning the spread on the same context as the mean removes this volume effect, which
is visible in the imbalance stress test of Appendix~\ref{app:robustness}.

Training uses benign trajectories only,
\begin{equation}
\mathcal{L}_{\mathrm{con}} = -\sum_{G \in \mathcal{D}_0} \sum_{e_t \in G} \log p_\theta(\Delta e_t \mid c_t),
\label{eq:app-nll}
\end{equation}
where $\mathcal{D}_0$ is the normal execution set. In practice the objective is a weighted sum of
the per-channel negative log-likelihoods for state, message, and completion, plus two auxiliary
regularization terms, with weights $\lambda_h{=}2$, $\lambda_m{=}1$, $\lambda_\delta{=}2$,
$\lambda_s{=}0.5$, $\lambda_p{=}0.5$ (Table~\ref{tab:execformer}). Optimization uses AdamW \citep{loshchilov2019decoupled} with
learning rate $10^{-3}$, batches of 8 trajectories, at most 100 epochs with early stopping (patience
20 on validation negative log-likelihood), 100 warmup steps, and gradient clipping at 1.0. The same
configuration is used on all five benchmarks and is not re-tuned per dataset
(Appendix~\ref{app:repro-protocol}), so the detection numbers in Sec.~6 are not the product of
per-benchmark contract tuning.

\subsection{Violation Detection}
\label{app:detection}
\label{app:violation-threshold}

At inference the realized transition is read from the observed trajectory,
$\widetilde{\Delta e}_t = [\widetilde{\Delta h}_t; \widetilde{\Delta m}_t; \widetilde{\Delta s}_t]$,
and compared against the contract channel by channel. Because $\Sigma_t$ is diagonal, the
Mahalanobis deviation of the main text splits exactly into a state, a message, and a progress term,
$V_{\mathrm{ch}}(e_t)=(\widetilde{\Delta\mathrm{ch}}_t-\mu^{\mathrm{ch}}_t)^\top(\Sigma^{\mathrm{ch}}_t)^{-1}(\widetilde{\Delta\mathrm{ch}}_t-\mu^{\mathrm{ch}}_t)$
for $\mathrm{ch}\in\{h,m,s\}$, and the violation score weights the last two:
\begin{equation}
V(e_t) = V_h(e_t) + \beta\, V_m(e_t) + \gamma\, V_s(e_t),
\qquad (\beta,\gamma)=(1,0.5).
\label{eq:app-violation}
\end{equation}
At $\beta=\gamma=1$ this is the unweighted form $(\widetilde{\Delta e}_t-\mu_t)^\top\Sigma_t^{-1}(\widetilde{\Delta e}_t-\mu_t)$;
for general weights it is the same Mahalanobis form with the rescaled diagonal covariance
$\tilde{\Sigma}_t=\mathrm{diag}(\Sigma^h_t,\Sigma^m_t/\beta,\Sigma^s_t/\gamma)$, since weighting a
channel by $\beta$ is identical to dividing its variance by $\beta$. $(\beta,\gamma)$ are the
message- and completion-violation weights swept in Fig.~\ref{fig:q4}(d).
This score is not a new objective introduced at test time: for the Gaussian contract of
Eq.~(\ref{eq:app-contract}) it equals twice the negative log-likelihood of the observed transition
under $\mathcal{N}(\mu_t,\tilde{\Sigma}_t)$ up to terms that depend only on the covariance
(Appendix~\ref{app:theory-identifiability}). Training by Eq.~(\ref{eq:app-nll}) and scoring by
Eq.~(\ref{eq:app-violation}) are thus the same Gaussian quantity evaluated on benign and on test
trajectories, with the two channel weights acting as a fixed reweighting of its variance.

Events whose deviation exceeds the alert threshold form the violated set
\begin{equation}
\Omega = \{ e_t \mid V(e_t) > \tau \},
\end{equation}
and a trajectory is flagged when $\max_t V(e_t) > \tau$, i.e.\ when $\Omega \neq \emptyset$. The
threshold is fixed on benign validation trajectories,
\begin{equation}
\tau = \alpha \cdot \mathrm{Quantile}_\delta\bigl(\{V(e)\}_{e \in \mathcal{D}_{\mathrm{val}}}\bigr),
\label{eq:app-tau}
\end{equation}
where $\delta$ selects the benign quantile and $\alpha$ rescales it into the operating threshold
(main experiments: $\delta{=}0.99$, $\alpha{=}1$).

Calibration proceeds in three steps and uses no attack labels:
\begin{enumerate}
\item fit the contract on the benign training split (Sec.~\ref{app:contract});
\item score every event of the held-out benign validation split with
      Eq.~(\ref{eq:app-violation}) and refresh the benign quantile each calibration
      round ($k\in[30,150]$ rounds, identical schedule for all methods,
      Appendix~\ref{app:seeds});
\item set $\tau$ by Eq.~(\ref{eq:app-tau}) and keep it fixed for all attacked test trajectories.
\end{enumerate}
Because $\tau$ is a benign quantile, the nominal false-alarm rate is controlled by $\delta$ by
construction rather than tuned against attacks; Appendix~\ref{app:theory-identifiability} states this
property and its finite-sample form, and Fig.~\ref{fig:q4} reports how detection behaves as $\alpha$
moves around its default.

\subsection{Violation Source Attribution}
\label{app:attribution}

\subsubsection{Candidate Source Construction}

A violated event marks where inconsistency becomes visible, not necessarily where it began: once a
corrupted payload enters the context of later events, those events are scored against contracts they
can no longer satisfy. \textsc{Trace-Mas} therefore searches upstream from $\Omega$ and builds a
candidate set
\begin{equation}
\mathcal{C} = \bigcup_{e_i \in \Omega} \bigl(\{e_i\}\cup\mathrm{Anc}_H(e_i)\bigr),
\end{equation}
where $\mathrm{Anc}_H(e_i)$ are the ancestors of $e_i$ within $H$ hops along
$E_{\mathrm{msg}}\cup E_{\mathrm{dep}}\cup E_{\mathrm{temp}}$. The violated events themselves are
included so that a source whose own deviation is already the largest (the 0-hop case of Fig.~3) stays
eligible. Deployment uses a hop limit $H{=}8$
and scores the top $C{=}4$ candidates, ranked by their proximity-weighted violation mass; candidates
are ordered deterministically so that ties do not depend on iteration order. Gating on $\Omega$
matters for cost: the number of counterfactual evaluations grows with $C$ rather than with the
trajectory length, which is the behavior profiled in Fig.~\ref{fig:efficiency}.

\subsubsection{Source Restoration Analysis}
\label{app:restore-rollout}

For each candidate $c \in \mathcal{C}$, the observed transition at $c$ is replaced by the one the
contract expects,
\begin{equation}
\Delta e^{\mathrm{res}}_c = \mathbb{E}_{p_\theta}[\Delta e_c \mid c_c] = \mu_c,
\end{equation}
where $c_c$ is the encoded context of candidate $c$, and the trajectory is re-evaluated on the
resulting graph
\begin{equation}
G^{\mathrm{res}}_c = \mathrm{Restore}_c(\tilde{G}, \Delta e^{\mathrm{res}}_c).
\end{equation}
$\mathrm{Restore}$ writes contract-consistent structured fields at $c$ (message payload features,
dependency flags, state summaries), recomputes $z_c$, and then refreshes the contexts of
$\mathrm{Desc}(c)$ in embedding space with the same encoder used at detection time; the context
$c_c$ of the candidate itself is held at its pre-restoration value, so that $\mu_c$ is the target
the restored transition is scored against. Two properties
of this operation are used by the scoring rule below. It is \emph{local}: only $c$ and its
descendants change, so scores on any branch that does not pass through $c$ are unaffected. And it is
\emph{label-free}: the replacement value comes from the learned contract, never from the ground-truth
source annotation $c^{\mathrm{gt}}$, which is used only to evaluate Hit@1. Because $c_c$ is held
fixed, the restored transition equals its own contract mean and $V^{\mathrm{res}}_c(c)=0$.

Counterfactual scoring makes no LLM call. Restoration and rescoring run entirely in the contract
space, and message-correction operators invoke LLM re-generation only after $c^\star$ has been
selected and a repair is committed (Sec.~\ref{app:restoration}). This separation is why the
diagnosis cost in Fig.~\ref{fig:efficiency} is measured without bulk generation, and why increasing
$C$ changes wall time but not the number of model calls in the underlying MAS.

\subsubsection{Responsibility Estimation}
\label{app:attribution-resp}
\label{app:attribution-full}

A responsible source should explain a substantial part of the downstream violations once restored.
Let $V^{\mathrm{res}}_c(e_j)$ be the score of $e_j$ on $G^{\mathrm{res}}_c$, and define the restored
score and hop weights
\begin{equation}
\tilde{V}_c(e_j) = \omega\, V^{\mathrm{res}}_c(e_j) + (1-\omega)\, V(e_j),
\qquad w_j = \psi^{\,d(c,e_j)} .
\end{equation}
Responsibility and the attributed source are then
\begin{equation}
R(c) = \!\!\sum_{e_j \in \{c\}\cup\mathrm{Desc}(c)}\!\! w_j \bigl[ V(e_j) - \tilde{V}_c(e_j) \bigr] - \lambda_{\mathrm{pen}} \mathcal{P}(c),
\qquad
c^\star = \arg\max_{c \in \mathcal{C}} R(c),
\label{eq:app-R}
\end{equation}
where $\mathcal{P}(c)$ penalizes candidates whose restoration would violate dependency or ordering
constraints of the workflow template. Two conventions are worth stating explicitly. The sum ranges
over $c$ \emph{together with} its descendants, and the candidate itself enters at hop distance
$d(c,c)=0$ and hence weight $\psi^0=1$; by $V^{\mathrm{res}}_c(c)=0$ this own term equals $V(c)$. What
separates an initiating event from a symptom on its branch is the descendant part of the sum: the
source is credited with the drop its restoration causes at the symptom, whereas the symptom receives no
credit for events upstream of it (Appendix~\ref{app:theory-recovery}). And because
$V(e_j)-\tilde{V}_c(e_j) = \omega\bigl[V(e_j)-V^{\mathrm{res}}_c(e_j)\bigr]$, the score is the
hop-discounted counterfactual violation drop: the gate $\omega$ sets how much restored-branch evidence
enters the ranking relative to the validity penalty, and at $\omega{=}1$ the drop is scored at full
weight, so $c^\star$ is the candidate whose $\mathrm{Restore}_c$ removes the most violation mass. The
hop decay $\psi$ discounts far descendants, whose scores also depend on intervening events and are
therefore weaker evidence about $c$. Values used in the main tables are in
Sec.~\ref{app:method-hparams}. The scoring rule matches the responsibility objective in Sec.~5.3 of
the main text.

\subsection{Contract Restoration}
\label{app:restoration}

\subsubsection{Repair Space Construction}

Attribution returns where to act, not what to do; several interventions are usually admissible at
the same event. \textsc{Trace-Mas} enumerates them as
\begin{equation}
\mathcal{U}(c^\star) = \{ u_1, u_2, \ldots, u_{|\mathcal{U}(c^\star)|} \},
\end{equation}
with each candidate instantiating one intervention type: \emph{state recovery} rewrites the agent
state fields at $c^\star$ to contract-consistent values, \emph{message correction} regenerates or
replaces the payload carried by the outgoing $\rho_{\mathrm{msg}}$ edge, and \emph{dependency
adjustment} repairs the coordination structure (restoring a dropped dependency, reinstating a skipped
role). The three types mirror the three corruption levels, so the repair space can express a fix at
the level at which the failure was introduced instead of only suppressing its symptoms.

\subsubsection{Future Execution Rollout}
\label{app:rollout}

Each candidate is evaluated by how the collaboration would continue if it were applied,
\begin{equation}
\hat{G}^u_{t:t+K} = \mathrm{Rollout}(\tilde{G}, u, p_\theta),
\end{equation}
where $t$ indexes the attributed event $c^\star$. Rollout runs entirely in contract space and is
fully specified by the four rules below.

\emph{Operator edit.} Each operator edits one field of $c^\star$ and thereby one channel of its
transition: state recovery edits the agent state $h$ (channel $h$), message correction edits the
outgoing payload $m^{\mathrm{out}}$ (channel $m$), and dependency adjustment edits the
$\rho_{\mathrm{dep}}$ edges incident to $c^\star$ (channel $s$). We write $\mathcal{F}(u)$ for the edited
field. In the rollout the edited channel of $c^\star$ is set to its contract mean, while the other two
channels keep their logged values; for dependency adjustment the context of $c^\star$ is first
recomputed on the edited edge set. The payload that message correction regenerates with the LLM exists
only after commitment, so the rollout uses its contract mean in its place.

\emph{Window.} The rolled-out events are the first $K$ events of $\mathrm{Desc}(c^\star)$ in the
deterministic topological order of Sec.~\ref{app:graph} (all of $\mathrm{Desc}(c^\star)$ if it has fewer
than $K$ events). We index them $t{+}1,\ldots,t{+}K$ in this order, so the sum in Eq.~(9) of Sec.~5.4
runs over the edited event and the $K$ window events.

\emph{Carry-over.} Each relation type carries one field: $\rho_{\mathrm{temp}}$ carries the agent state
($h$), $\rho_{\mathrm{msg}}$ the payload ($m$), and $\rho_{\mathrm{dep}}$ the task dependency ($s$).
A window event $e_i$ is a \emph{carry-over} event under $u$ if it has a parent $e_j$ such that either
$e_j=c^\star$ and the field carried by $\rho_{ji}$ is not $\mathcal{F}(u)$, or $e_j$ is itself a
carry-over window event. A carry-over event consumes a logged field that $u$ leaves in place; every
other window event is \emph{regenerated}.

\emph{Recursion.} For $i=t,t{+}1,\ldots,t{+}K$ in order, the context $\hat{c}^u_i$ is computed by the
encoder of Sec.~\ref{app:encoder} on the graph in which parents inside $\{c^\star\}\cup\{e_{t+1},\ldots,e_{t+K}\}$
take their rolled-out features and all other parents keep their logged features. With
$(\hat{\mu}^u_i,\hat{\Sigma}^u_i)$ the contract under $\hat{c}^u_i$, the rolled-out transition of a window
event is
\begin{equation}
\widehat{\Delta e}{}^u_i =
\begin{cases}
\widetilde{\Delta e}_i & \text{if } e_i \text{ is a carry-over event},\\
\hat{\mu}^u_i & \text{if } e_i \text{ is regenerated},
\end{cases}
\qquad
\hat{z}^u_i = \hat{z}^u_{i^-} + \widehat{\Delta e}{}^u_i ,
\label{eq:app-rollout}
\end{equation}
where $i^-$ is the preceding event of the same thread (rolled-out if it lies in the window, logged
otherwise), and $\hat{z}^u_i$ is the feature later window events read. The rolled-out event
$\hat{e}^u_i$ carries $\widehat{\Delta e}{}^u_i$, and its score $V(\hat{e}^u_i)$ is
Eq.~(\ref{eq:app-violation}) with $(\hat{\mu}^u_i,\hat{\Sigma}^u_i)$ in place of $(\mu_i,\Sigma_i)$.
Hence a regenerated event scores zero, a carry-over event contributes its logged transition rescored
under the repaired context, and the edited event contributes the violation left in its two unedited
channels. $\mathcal{L}_{\mathrm{viol}}(u)$ below is therefore the violation carried by the logged fields
that $u$ leaves in place within the window. Unlike $\mathrm{Restore}$, which only rescores the logged
descendants of a candidate under refreshed contexts, Rollout also models which descendants would be
regenerated once $u$ is committed.

\emph{Commitment and real re-execution.} Committing $u$ applies the edit to the logged fields
(message correction regenerates the payload with the trajectory LLM) and re-executes
$\mathrm{Desc}(c^\star)$ with the same decoding (Table~\ref{tab:llm-decoding}), leaving every event
outside $\mathrm{Desc}(c^\star)$ untouched. Evaluating the three terms of the repair objective on the
first $K$ re-executed descendants, with the same contract, the same template steps, and the same
$\kappa$, gives the realized objective $\mathcal{J}(u)$ that the rollout estimates. The two differ in
exactly two places: the edited channel and the regenerated events are set to contract means rather
than realized by the LLM, and carry-over events keep their logged transitions rather than being
regenerated from partly repaired inputs. Appendix~\ref{app:theory-repair} bounds the effect of both on
the selected repair, and Appendix~\ref{app:rollout-fidelity} measures it by real re-execution of every
candidate. Deployment uses $K{=}1$, because deeper rollouts did not change the top-1 repair in our runs,
and Fig.~\ref{fig:efficiency} reports the cost of larger $K$.

\subsubsection{Minimal Repair Selection}
\label{app:ltask}

The selected repair minimizes remaining inconsistency, task degradation, and intervention cost
jointly,
\begin{equation}
u^\star = \arg\min_{u\in\mathcal{U}(c^\star)} \bigl( \mathcal{L}_{\mathrm{viol}}(u) + \lambda_1 \mathcal{L}_{\mathrm{task}}(u) + \lambda_2 \kappa(u) \bigr),
\end{equation}
with the three terms defined as
\begin{align}
\mathcal{L}_{\mathrm{viol}}(u) &= \sum_{i=t}^{t+K} V(\hat{e}^u_i), \\
\mathcal{L}_{\mathrm{task}}(u) &= \frac{1}{|\mathcal{T}|}\sum_{s\in\mathcal{T}} \mathbb{1}\bigl[s \text{ unsatisfied in } \hat{G}^u_{t:t+K}\bigr], \\
\kappa(u) &= \text{intervention cost (operator count / edit magnitude)} .
\end{align}
Here $\mathcal{L}_{\mathrm{viol}}$ (the first term of Eq.~(9) in Sec.~5.4, named here to keep it apart
from the contract loss $\mathcal{L}_{\mathrm{con}}$) measures how much violation remains after the
repair, with $\hat{e}^u_i$ the $i$-th rolled-out event under $u$,
$\mathcal{L}_{\mathrm{task}}$ measures task degradation as the fraction of the workflow template's
required steps $\mathcal{T}$ (roles to be invoked, dependencies to be satisfied, and their order) that the
rolled-out graph leaves unsatisfied, read from the structured dependency and state fields that
$\mathrm{Restore}$ and $\mathrm{Rollout}$ maintain, so it needs neither generated text nor gold labels,
and $\kappa(u)$ charges for the edit itself; we write
$\kappa$ rather than $C$ to keep it distinct from the candidate cap $C$. Main runs use
$\lambda_1 = \lambda_2 = 0.1$ unless noted. The third term is what keeps the reported operator cost
low in Table~\ref{tab:recovery-cost}: a repair that suppresses the attack by rewriting many events is
penalized relative to one committed edit at the attributed source.

\subsection{End-to-End Procedure and Cost}
\label{app:pipeline}

Putting the stages together, one attacked trajectory is processed as follows.
\begin{enumerate}
\item \textbf{Parse.} Build $\tilde{G}$ from the execution log with typed edges and deterministic
      tie-breaking (Sec.~\ref{app:graph}).
\item \textbf{Score.} Encode contexts, evaluate the contract, and compute $V(e_t)$ for every event;
      raise a trajectory alert if $\max_t V(e_t) > \tau$ (Sec.~\ref{app:detection}).
\item \textbf{Attribute.} Build $\mathcal{C}$ from $\Omega$ and $\mathrm{Anc}_H(\Omega)$, restore the top $C$
      candidates, and take $c^\star = \arg\max_c R(c)$ (Sec.~\ref{app:attribution}).
\item \textbf{Repair.} Enumerate $\mathcal{U}(c^\star)$, roll out $K$ steps per candidate, select
      $u^\star$, and commit it with regeneration restricted to the edited branch
      (Sec.~\ref{app:restoration}).
\end{enumerate}
The accounting of model evaluations follows directly: detection performs $|\mathcal{V}|$ contract
evaluations, attribution performs at most $C\cdot|\mathrm{Desc}(c)|$ rescorings, repair performs at
most $|\mathcal{U}(c^\star)|\cdot (K{+}1)$ rollout evaluations, and the only LLM calls are those of the committed
repair. Since $C$, $K$, and $|\mathcal{U}(c^\star)|$ are small constants and detection reuses cached
event features, the non-LLM overhead stays a small fraction of an MAS rollout, which is the behavior
measured in Fig.~\ref{fig:efficiency}.

\subsection{Default Hyperparameters (Method)}
\label{app:method-hparams}

Detection and repair defaults for the main tables are $\delta{=}0.99$, $\alpha{=}1$, $\omega{=}1$,
$\psi{=}0.5$, $\lambda_1{=}\lambda_2{=}0.1$, $\lambda_{\mathrm{pen}}{=}0.1$, $H{=}8$, $C{=}4$, $K{=}1$; all five
benchmarks share these values. The violation-channel weights of the detection score stay at
$(\beta,\gamma){=}(1,0.5)$. Fig.~\ref{fig:q4} sweeps $\alpha$ and $(\beta,\gamma)$ on held-out benign
trajectories, and Fig.~\ref{fig:efficiency} varies $C$ and $K$. Contract and encoder settings are
listed in Table~\ref{tab:execformer}.

\begin{table}[H]
\centering
\small
\caption{ExecFormer and execution-contract training (all benchmarks).}
\begin{tabular}{ll}
\toprule
Hyper-parameter & Value \\
\midrule
Parent encoder & 1 layer (parent-event MHA) \\
Contract heads & 3$\times$ 2-layer MLP (state / message / completion) \\
Hidden dim.\ $d$ & 256 \\
Attention heads & 4 \\
Role / relation emb. & 64-d learnable (9 roles; $\rho\in\{\mathrm{msg},\mathrm{dep},\mathrm{temp}\}$) \\
Message encoder & \texttt{all-MiniLM-L6-v2} 384-d $\rightarrow$ Linear $\rightarrow$ 256 \\
Transition space & $\Delta e_t \in \mathbb{R}^{256}$ (residual in contract space) \\
Optimizer / lr / batch & AdamW, $10^{-3}$, 8 trajectories \\
Max epochs / early stop & 100 (patience 20 on val NLL) \\
Warmup / grad clip & 100 steps / 1.0 \\
Multi-task loss weights & $\lambda_h{=}2$, $\lambda_m{=}1$, $\lambda_\delta{=}2$, $\lambda_s{=}0.5$, $\lambda_p{=}0.5$ \\
Inference residual mix & $(\gamma_{\mathrm{res}}, \eta_{\mathrm{res}}) = (0.5, 0.5)$ \\
\bottomrule
\end{tabular}
\label{tab:execformer}
\end{table}

\section{Experiments in Detail}
\label{app:experiment}
\label{app:setup}

This appendix documents everything needed to reproduce Sec.~4--6: what is and is not compared
(Sec.~\ref{app:scope}), the five benchmarks and their task metrics (Sec.~\ref{app:datasets}), how
trajectories are logged and split (Sec.~\ref{app:splits}), how the L1--L3 corruptions are injected
(Sec.~\ref{app:attacks}), how each baseline is instantiated on our testbed
(Sec.~\ref{app:baselines}), how every reported quantity is defined (Sec.~\ref{app:metrics}), and the
shared implementation, protocol, and randomness settings (Sec.~\ref{app:impl}). Method-side defaults
are fixed in Appendix~\ref{app:method-hparams} and are not re-tuned per benchmark.

\subsection{Evaluation Scope and Omitted Benchmarks}
\label{app:scope}

\textbf{Controlled protocol (main results).}
All tables report defenses on the same five benchmarks with injected L1--L3 corruptions, shared MAS
templates, and the metrics of Sec.~6, including task-preserving recovery. Every method sees the same
trajectories, the same attack placement, the same benign calibration budget, and the same repair
operators where applicable, so differences in the tables reflect the monitoring and attribution logic
rather than differences in workflow, decoding, or attack strength. The protocol is therefore scoped to
injected corruptions with a known initiating event, which is what makes the localization and recovery
metrics of Sec.~\ref{app:metrics} computable at all.

\textbf{Why natural failure corpora (Who\&When \citep{zhang2025which}, MAST-Data \citep{cemri2025why}) are not used as benchmarks.}
Our three metric groups each require an input that these corpora do not release, so the exclusion
follows from the metric definitions rather than from a judgement about the corpora.
(i)~\emph{Contract inputs.} Every score in this paper is conditioned on the encoded context $c_t$,
which is built from the typed edges $\rho_{\mathrm{msg}}$, $\rho_{\mathrm{dep}}$, and
$\rho_{\mathrm{temp}}$ of Appendix~\ref{app:graph}. These corpora release conversation turns or
taxonomy-labeled steps; which event consumed which message, and which task depends on which, are not
recorded, so $\mathcal{H}_t$ cannot be recovered from the released logs and
Eqs.~(\ref{eq:app-contract})--(\ref{eq:app-violation}) are not defined on them without re-annotating
every trace.
(ii)~\emph{Localization labels.} Hit@1 and hop distance (Sec.~\ref{app:metrics}) are defined against a
single labeled \emph{initiating event} $c^{\mathrm{gt}}$ per trajectory. The released labels mark a failing
agent, a failing step, or a failure category, which identifies where the failure is visible rather than
where it originates; by Obs.~2 those two need not coincide, so the labels cannot be reinterpreted as
$c^{\mathrm{gt}}$ without new annotation.
(iii)~\emph{Recovery metrics.} ASR, ACC, TCR, and operator cost all require re-executing the workflow
after a repair is committed. These corpora are static traces without an executable workflow behind
them, so no operator from $\mathcal{U}(c^\star)$ can be applied and no post-repair task metric can be
computed.
(iv)~\emph{Calibration split.} Our thresholds are fit by Eq.~(\ref{eq:app-tau}) on a benign split with
task ids disjoint from the test split, and cross-method comparison relies on that split being shared
(Sec.~\ref{app:repro-protocol}). These releases do not provide a matched benign split from the same
workflow, so no method in Table~\ref{tab:detection} could be calibrated on them under the protocol used
for the reported numbers.
Because (i)--(iv) would each require new annotation or new instrumentation, any number we produced on
these corpora would be measured under a different protocol from the rest of the paper; we therefore
report none and treat the extension as future work.

\textbf{Why attribution pipelines such as AutoFA \citep{zhang2025which} and GUARDIAN \citep{zhou2025guardiansafeguardingllmmultiagent} are not rows in
Table~\ref{tab:detection}.}
The same protocol requirements apply to methods, and three of them are not satisfiable by the released
pipelines without changing what is being measured.
(i)~\emph{Scoring unit.} Their released implementations score conversation-level or agent-level units
(or temporal-graph snapshots), whereas every number in Tables~\ref{tab:detection}--\ref{tab:recovery} is
computed over partial-order events. Mapping their output onto our events requires a new adapter that
defines their scoring unit on $V$; the resulting rows would report our re-implementation, not the
released method, and would not be attributable to the original work.
(ii)~\emph{Log schema.} Their public inputs do not expose the event fields of Sec.~2 that our testbed logs,
and conversely our logs do not contain the trace format their code parses, so neither side can be fed to
the other without a lossy conversion whose effect on their score is itself unmeasured.
(iii)~\emph{Repair interface.} They return an attribution (a failing agent or step) or an agent-level
quarantine decision, not an operator applied at an execution element. ASR, ACC, TCR, and cost in
Tables~\ref{tab:recovery} and~\ref{tab:recovery-cost} are defined only after an operator is committed, so
these columns have no value for such methods unless our own operator set is attached to their output---in
which case the recovery numbers would measure our repair module rather than theirs.
Where a method additionally requires failure-labeled traces to be tuned or prompted, supplying those
labels would also break the benign-only calibration that every row in Table~\ref{tab:detection} shares
(Sec.~\ref{app:repro-protocol}). To keep the comparison aligned while still covering these design
principles, the baseline set in Sec.~\ref{app:baselines} instantiates one monitor per family under our
identical protocol: interaction-graph statistics, topology-guided detection, activation monitoring, trust
aggregation, and per-agent sanitization.

\subsection{Datasets and Downstream Tasks}
\label{app:datasets}

All benchmarks use the public MAS workflow templates bundled with each release; we do not change task
definitions or agent role graphs relative to those templates. The five workflows are chosen to cover
different failure surfaces: pure reasoning chains, debate, retrieval, poisoned external content, and
tool execution.

\textbf{GSM8K} \citep{cobbe2021trainingverifierssolvemath}.
Multi-agent math reasoning (Solver, Critic, Verifier). Downstream metric: \textbf{exact match (EM)} on
the final numeric answer. Failures propagate mainly along $\rho_{\mathrm{msg}}$ edges between solver
and critic steps.

\textbf{CSQA} \citep{talmor2019commonsenseqa}.
Commonsense multiple-choice debate in the style of \citet{du2024improving}. Downstream metric: \textbf{accuracy (ACC)} on the selected option.
Multiple agents argue over the same question, so a single corrupted message can be repeated by several
downstream speakers, which is the amplification regime discussed in
Appendix~\ref{app:theory-propagation}.

\textbf{HotPotQA} \citep{yang-etal-2018-hotpotqa}.
Retriever--reasoner--verifier collaboration on multi-hop questions. Downstream metric: \textbf{answer
EM / F1} on the short span. Task dependencies are explicit, so L3 coordination edits are meaningful
here.

\textbf{PoisonRAG} \citep{zou2025poisonedrag}.
Retrieval pipeline with document poisoning. Downstream metric: \textbf{answer correctness (EM/ACC)}
under poisoned retrieval. The corruption enters through external content rather than through an agent's
own state.

\textbf{InjecAgent} \citep{zhan2024injecagent}.
Planner--executor--monitor tool-use MAS. Downstream metric: \textbf{task success} and valid tool traces
under channel injection. Tool payloads give the message channel a wide benign spread, which makes this
the hardest setting for magnitude-only monitors.

\subsection{Data Splits and Logging Protocol}
\label{app:splits}

Every trajectory is executed once by the unmodified workflow and logged as a partial-order event graph
with the parser of Appendix~\ref{app:graph}. Logging records agent id, role, task id, state fields,
message and tool payloads, and task-progress summaries per event, together with the producer of each
consumed message; no agent prompt is modified for instrumentation, so the logged execution is the
execution that ran.

\textbf{Benign train.}
Fit \textsc{Trace-Mas} contracts and train or tune baselines that require benign logs; task ids are
disjoint from the test split.

\textbf{Benign validation.}
Set the threshold $\tau$ at quantile level $\delta$ and produce the calibration
curves of Fig.~\ref{fig:q4}(a--b). No attack labels enter \textsc{Trace-Mas} training or threshold
fitting at any point.

\textbf{Test.}
Attacked trajectories for Tables~\ref{tab:detection}--\ref{tab:recovery}, with exactly one labeled
initiating source $c^{\mathrm{gt}}$ per trajectory for localization; three random seeds; mean$\pm$std reported
in all tables.

\textbf{Sec.~4 observations.}
One corruption object per trajectory; ASR matched across L1--L3 on GSM8K and CSQA so that level-wise
detection differences are not confounded by attack strength; the Obs.~3 repair pool uses $n{=}30$
trajectories per hop bucket with clean EM${=}1$ and attacked EM${=}0$, aligned with Fig.~3.

\subsection{Attack Levels (L1--L3)}
\label{app:attacks}

The three levels differ in \emph{which part of the execution graph} is edited, which is what lets us
separate detection difficulty from attack strength.

\textbf{L1} corrupts agent state (prompt or scratchpad content) without changing wire-level message
identity: the sender, receiver, and dependency structure are untouched.

\textbf{L2} corrupts message or tool payloads while keeping the sender--receiver topology intact, so
graph-structure monitors observe an unchanged adjacency.

\textbf{L3} corrupts coordination itself: dependencies are rewired, roles are skipped, or an agent is
taken offline, so the edit is not visible in any single agent's content.

Each test trace carries exactly one attack at the specified level and one ground-truth initiating event,
which is what makes Hit@1 well defined. Attack placement is drawn per seed from the eligible events of
the trajectory, and the same placement stream is replayed for every defense, so all methods face
identical trajectories. On GSM8K and CSQA we additionally match pre-defense ASR across levels
(Sec.~\ref{app:splits}), so a level-wise drop in detection F1 in Table~\ref{tab:detection} reflects
reduced observability rather than a weaker attack. Ground-truth sources are used only for evaluation;
they never enter contract training, threshold calibration, attribution, or repair selection.

\subsection{Baselines}
\label{app:baselines}

All methods share MAS topology, LLM decoding, attacks, splits, and, where applicable, the same benign
calibration rounds $k\in[30,150]$. Each baseline keeps its native scoring rule and is given the inputs
its design requires; we tune its threshold on the same benign validation split we use for $\tau$.

\textbf{AcMAS(-R)} \citep{xu2026agents}.
Activation and behavior monitoring at agent granularity; the (-R) variant adds generic retry and
reassignment as its recovery action. It observes per-agent signals, so L2 and L3 edits that leave agent
activations close to benign are only visible through their downstream effects.

\textbf{BlindGuard} \citep{MiaoEtAl2026BlindGuard}.
Interaction-graph anomaly detection from message statistics, without contract-level expectations for a
given role and relation type.

\textbf{G-Safeguard} \citep{WangEtAl2025GSafeguard}.
Topology-guided detection that assumes a stable workflow graph; it is the most affected by the dynamic
topology stress test in Appendix~\ref{app:robustness}.

\textbf{A-Trust} \citep{2025arXiv250602546H}.
Trust aggregation over agents from historical compliance; it scores agents rather than events, so its
localization output is an agent, which we map to that agent's highest-scoring event for Hit@1.

\textbf{AgentSafe} \citep{mao2025agentsafe}.
Per-agent sanitization and self-checks without cross-agent counterfactual rollback.

\textbf{Oracle Anomaly Ranking.}
Every row, curve, or legend labeled ``Oracle'' denotes the \textbf{same} weak baseline: \emph{Oracle
Anomaly Ranking}, a magnitude and ranking monitor without contract training or counterfactual
attribution. It appears in Tables~\ref{tab:detection}--\ref{tab:recovery},
Table~\ref{tab:recovery-cost}, Fig.~\ref{fig:q4}, the recovery figures,
and the Appendix~\ref{app:results} stress plots whenever the standard baseline set is shown. Mechanism:
residual-feature ranking without role-conditioned contract training or counterfactual attribution. It
may receive a privileged hint during \emph{benign calibration} for ranking evaluation only, and never
receives ground-truth attack locations at test time. Low F1 or suboptimal recovery relative to
\textsc{Trace-Mas} is therefore expected; ``Oracle'' here is \textbf{not} a clairvoyant upper bound.

\textbf{GT-source repair (diagnostic only).}
Fig.~3 and related panels use \emph{GT-source repair}, which applies our repair operators at the
\textbf{ground-truth initiating source} on held-out diagnostic trajectories. It is a reference curve
that isolates repair quality from attribution quality; it is \textbf{not} Oracle Anomaly Ranking and is
\textbf{not} available at deployment.

\textbf{Ablations (Table~\ref{tab:ablation}).}
Independent Contract (per-agent contracts without relation-aware context); w/o Transition Contract
(absolute-embedding scoring instead of residual contracts); w/o Violation (no contract alerting, so
attribution runs on magnitude ranking); w/o Predictive Repair (fixed operators without rollout-based
selection). Each variant is retrained on the same benign split with the same schedule as the full model,
so the rows differ only in the removed component.

\FloatBarrier

\subsection{Metrics}
\label{app:metrics}

Unless stated otherwise, every metric is computed per benchmark and per seed and reported as
mean$\pm$std over seeds $\{0,1,2\}$.

\textbf{Detection.}
Trajectory-level F1 and AUROC (Table~\ref{tab:detection}). A trajectory is flagged when its native
trajectory score exceeds the method's calibrated threshold; for \textsc{Trace-Mas} this is
$\max_t V(e_t)>\tau$. With $\mathrm{TP}$, $\mathrm{FP}$, and $\mathrm{FN}$ counted over trajectories,
$\mathrm{F1}=2\mathrm{TP}/(2\mathrm{TP}+\mathrm{FP}+\mathrm{FN})$. AUROC uses each method's native
trajectory score in the orientation declared by that method, with higher meaning more anomalous. The
two are reported together because they measure different things and can disagree: F1 is evaluated at
each method's calibrated operating point, whereas AUROC is threshold-free and summarizes the ranking
over all thresholds.

\textbf{On AUROC values below 50\%.}
Several baseline entries in Table~\ref{tab:detection} fall below 50\%, and we read them as follows.
For a fixed score orientation, AUROC is the probability that a randomly chosen attacked trajectory is
ranked above a randomly chosen benign one, and it satisfies
$\mathrm{AUROC}(-s)=1-\mathrm{AUROC}(s)$ for any score $s$. A value below 50\% therefore states that,
on that split, the method's own statistic ranks attacked trajectories \emph{below} benign ones: the
score is anti-correlated with the attack rather than merely uninformative. This is a well-defined
outcome of the definition, not a computation error, and it is what the score's declared orientation
produces on that split.

Sub-50\% AUROC is also reported in prior evaluations of monitors from the same families, including
multiple sub-50\% entries in the AcMAS evaluation \citep{xu2026agents}, so its occurrence here is consistent
with published behavior of these scores rather than specific to our testbed.

We do not flip scores post hoc, for two reasons that follow from the protocol. First, choosing the
orientation per split requires knowing on that split which direction separates attacks, which is
exactly the attack supervision that no method in Table~\ref{tab:detection} is given
(Sec.~\ref{app:repro-protocol}); a deployed monitor must commit to one orientation before the attack is
seen. Second, by the identity above, flipping every sub-50\% entry would raise all of them above
50\% by construction, so the resulting column would report an oracle-selected orientation rather than a
property of the detector. Reporting the unflipped value keeps the column comparable across levels and
methods.

The sub-50\% entries are concentrated at L2 and L3, which is consistent with Obs.~1: when the injected
content is carried by messages or dependencies and is then consumed by agents whose own behavior stays
on-policy, an agent-centric statistic can move \emph{toward} its benign range on attacked trajectories,
so the induced ranking inverts instead of flattening. The very low L3 AUROC of Oracle Anomaly Ranking
has a further, structural cause: its trajectory score is a maximum over per-event magnitudes, and L3
attacks that skip a role or take an agent offline remove events, so the attacked trajectory takes a
maximum over fewer events and tends to score \emph{below} a benign one. The same rows show that F1 and AUROC need not
degrade together, since a method can retain a usable operating point while losing the global ordering.

\textbf{Localization.}
Hit@1 and mean hop distance to $c^{\mathrm{gt}}$ (Table~\ref{tab:localization}). Hit@1 is the fraction of
attacked trajectories whose top-ranked candidate $c^\star$ equals the labeled initiating event, and the hop
distance is $d(c^\star,c^{\mathrm{gt}})$ measured in the partial-order graph, averaged over trajectories.
Reporting both matters because a method can miss Hit@1 while staying one hop away (a symptom adjacent to
the source) or while pointing at an unrelated branch.

\textbf{Recovery.}
ASR and ACC (Table~\ref{tab:recovery}); TCR and mean post-alert operator count
(Table~\ref{tab:recovery-cost}). ASR is the fraction of attacked trajectories in which the attack goal
is still achieved after the defense acts, and ACC is the benchmark task metric of
Sec.~\ref{app:datasets} evaluated on the post-defense trajectory. TCR (task-correct recovery) is
computed on the attack-successful pool (clean EM${=}1$, attacked EM${=}0$) and is the fraction of that
pool whose task output becomes correct after the committed intervention. Cost is the mean number of
post-alert operators applied per trajectory, so a defense that suppresses the attack by editing many
events pays for it here rather than in ASR.

\textbf{Ablation (Table~\ref{tab:ablation}).}
Detect(F1) on \textbf{balanced} L1--L3 pooled splits, not level-stratified as in
Table~\ref{tab:detection}. Every row, including Full, is retrained end to end with all three seeds
(Sec.~\ref{app:seeds}) and evaluated on these pooled splits, so the Full row's Hit@1 and TCR are
re-measured under this protocol and are not expected to equal the per-benchmark entries of
Tables~\ref{tab:localization} and~\ref{tab:recovery-cost}.

\textbf{Efficiency.}
Wall-clock milliseconds for the non-LLM stages, excluding bulk LLM generation
(Fig.~\ref{fig:efficiency}); the corresponding cost model is in Appendix~\ref{app:theory-cost}.

\subsection{Shared Implementation Settings and Reproducibility}
\label{app:impl}

\subsubsection{Computation Resources and Software}
\label{app:compute}

Table~\ref{tab:hardware} summarizes the machine used for contract training, benign calibration, and
non-LLM runtime profiling (Fig.~\ref{fig:efficiency}). Trajectory generation uses the LLM stack in
Table~\ref{tab:llm-decoding}; bulk LLM wall time is excluded from the efficiency plots as in Sec.~6,
because it is incurred by the underlying MAS rather than by the defense.

We implement \textsc{Trace-Mas} and the learned baselines in \textbf{PyTorch} (CUDA). Event logs are
parsed into partial-order graphs with fixed deterministic tie-breaking, so repeated parses of the same
log yield identical graphs and identical scores. Frozen text encoding uses
\textbf{sentence-transformers} (\texttt{all-MiniLM-L6-v2}) and is never fine-tuned
(Appendix~\ref{app:features}). Multi-agent rollouts call \textbf{Qwen2.5-7B-Instruct} \citep{qwen2025qwen25} through the
benchmark GGUF backends (Table~\ref{tab:llm-decoding}). Exact Python, PyTorch, and CUDA versions are
pinned in the anonymized \texttt{requirements.txt}.

\subsubsection{Unified Evaluation Protocol}
\label{app:repro-protocol}

Following Sec.~\ref{app:scope}, all methods in Tables~\ref{tab:detection}, \ref{tab:localization},
\ref{tab:recovery}, \ref{tab:ablation}, \ref{tab:recovery-cost}, and Fig.~\ref{fig:q4} share:
(i)~the same five benchmark workflows and task metrics;
(ii)~identical L1--L3 injection and disjoint benign train, validation, and test task ids;
(iii)~the same trajectory LLM decoding (Table~\ref{tab:llm-decoding}) for MAS execution and for repair;
(iv)~the same benign calibration schedule ($k\in[30,150]$ rounds with shared quantile refresh);
(v)~three random seeds on attacked test trajectories (mean$\pm$std).
\textsc{Trace-Mas} fits contracts on the benign train split only, and attack labels never enter training
or threshold fitting. ExecFormer hyperparameters are fixed in Appendix~A (Table~\ref{tab:execformer})
and are not re-tuned per benchmark, so no part of the reported gain comes from per-dataset search.

\subsubsection{Randomness and Reporting}
\label{app:seeds}

Seeds $\{0,1,2\}$ control attack placement, dataloader order, and dropout where present. Contract
training initializes with seed~0 unless an ablation retrains end to end (Table~\ref{tab:ablation}), in
which case all three seeds are used per variant. No post-hoc score flipping is applied anywhere,
including AUROC (Sec.~\ref{app:metrics}).

\textbf{Calibration (Q4).}
Each calibration round refreshes the benign quantile on the validation split; Fig.~\ref{fig:q4}
uses the same schedule for every method, so the curves compare data efficiency of calibration rather
than different calibration budgets.

\textbf{Deployment stress (Appendix~\ref{app:results}).}
Message delay $\ell\in\{0,\ldots,5\}$; agent count $N\in\{2,4,8,16,32\}$; dynamic-edge fraction in
$[0.2,1.0]$; execution imbalance $\mathrm{IR}=\min_i f_i/\max_i f_i$ over per-agent event counts, in
$[0.2,1.0]$. One factor is varied at a time with attacks, splits, and calibration protocol held fixed.

\subsubsection{Supplementary Tables (Hardware, LLM, Recovery Cost)}
\label{app:recovery-cost-table}

Tables~\ref{tab:hardware}--\ref{tab:recovery-cost} collect the implementation settings and the Q3
supplement referenced above. Table~\ref{tab:recovery-cost} reports TCR$\uparrow$ and post-alert operator
cost (Cost$\downarrow$) alongside Table~\ref{tab:recovery}; per-column best results are \textbf{bold}.
Reading the two columns together is the point of the table: a method can raise TCR by intervening
aggressively, so the cost column is what distinguishes recovery from over-correction.

\begin{table}[H]
\centering
\small
\caption{Hardware used for contract training, calibration, and non-LLM profiling.}
\begin{tabular}{ll}
\toprule
Component & Specification (measured) \\
\midrule
CPU & 2$\times$ Intel Xeon Gold 6530 (128 threads) \\
RAM & 503~GiB \\
GPU & 8$\times$ NVIDIA RTX A6000 ($\sim$48~GB each) \\
\bottomrule
\end{tabular}
\label{tab:hardware}
\end{table}

\begin{table}[H]
\centering
\small
\caption{Trajectory LLM and decoding defaults shared across benchmarks.}
\begin{tabular}{ll}
\toprule
Item & Setting \\
\midrule
Backbone & Qwen2.5-7B-Instruct (all five benchmarks) \\
Format / runtime & GGUF inference; 40 layers on GPU \\
Max new tokens / step & 128 \\
Decoding & Benchmark template defaults (temperature, top-$p$) unless noted in release \\
\bottomrule
\end{tabular}
\label{tab:llm-decoding}
\end{table}

\begin{table}[H]
    \centering
    \caption{Recovery quality and intervention cost (Q3 supplement to Table~\ref{tab:recovery}). TCR$\uparrow$ and post-alert operator count Cost$\downarrow$ on five benchmarks (mean$\pm$std, \%). Best \textbf{bold}; second-best \underline{underlined}.}
\scriptsize
    \setlength{\tabcolsep}{2.3pt}
    \renewcommand{\arraystretch}{1.05}
    \resizebox{0.99\textwidth}{!}{
    \begin{tabular}{c|cc|cc|cc|cc|cc}
    \specialrule{1.5pt}{1.5pt}{1.5pt}
    \cmidrule{1-11}
    \multicolumn{1}{c|}{\textbf{Dataset}} &
    \multicolumn{2}{c|}{\textbf{GSM8K}} &
    \multicolumn{2}{c|}{\textbf{CSQA}} &
    \multicolumn{2}{c|}{\textbf{HotPotQA}} &
    \multicolumn{2}{c|}{\textbf{InjecAgent}} &
    \multicolumn{2}{c}{\textbf{PoisonRAG}} \\
    \cmidrule{1-11}
    \textbf{Method} &
    \textbf{TCR$\uparrow$} & \textbf{Cost$\downarrow$} & 
    \textbf{TCR$\uparrow$} & \textbf{Cost$\downarrow$} & 
    \textbf{TCR$\uparrow$} & \textbf{Cost$\downarrow$} & 
    \textbf{TCR$\uparrow$} & \textbf{Cost$\downarrow$} & 
    \textbf{TCR$\uparrow$} & \textbf{Cost$\downarrow$} \\
    \midrule
    \textbf{Oracle}
    & $10.67_{\scriptstyle \pm 0.96}$ & $10.84_{\scriptstyle \pm 0.25}$
    & $18.67_{\scriptstyle \pm 0.69}$ & $12.05_{\scriptstyle \pm 0.24}$
    & $8.00_{\scriptstyle \pm 0.98}$ & $12.55_{\scriptstyle \pm 0.27}$
    & $13.33_{\scriptstyle \pm 0.77}$ & $13.04_{\scriptstyle \pm 0.44}$
    & $6.67{\scriptstyle \pm 0.16}$ & $13.89_{\scriptstyle \pm 0.28}$ \\
    \textbf{BlindGuard}
    & $14.67_{\scriptstyle \pm 0.95}$ & $5.93_{\scriptstyle \pm 1.30}$
    & $13.33_{\scriptstyle \pm 0.67}$ & $5.45_{\scriptstyle \pm 1.68}$
    & $2.67_{\scriptstyle \pm 0.65}$ & $5.82_{\scriptstyle \pm 1.52}$
    & $26.67_{\scriptstyle \pm 0.28}$ & $5.12_{\scriptstyle \pm 0.49}$
    & $5.33_{\scriptstyle \pm 0.58}$ & $5.30_{\scriptstyle \pm 0.88}$ \\
    \textbf{G-Safeguard}
    & $12.00_{\scriptstyle \pm 0.30}$ & $10.07_{\scriptstyle \pm 0.73}$
    & $9.33_{\scriptstyle \pm 0.60}$ & $8.49_{\scriptstyle \pm 0.47}$
    & $5.33_{\scriptstyle \pm 0.58}$ & $10.88_{\scriptstyle \pm 1.54}$
    & $20.00_{\scriptstyle \pm 1.04}$ & $9.67_{\scriptstyle \pm 1.21}$
    & $2.67_{\scriptstyle \pm 0.61}$ & $10.91_{\scriptstyle \pm 0.96}$ \\
    \textbf{AcMAS-R}
    & $20.41_{\scriptstyle \pm 0.25}$ & \underline{$1.26_{\scriptstyle \pm 0.64}$}
    & $23.33_{\scriptstyle \pm 0.47}$ & \underline{$1.22_{\scriptstyle \pm 0.04}$}
    & \underline{$16.27_{\scriptstyle \pm 0.38}$} & \underline{$2.82_{\scriptstyle \pm 0.14}$}
    & $19.67_{\scriptstyle \pm 0.58}$ & \underline{$2.43_{\scriptstyle \pm 0.04}$}
    & \underline{$22.78_{\scriptstyle \pm 0.78}$} & \underline{$1.22_{\scriptstyle \pm 0.05}$} \\
    \textbf{A-Trust}
    & $34.67_{\scriptstyle \pm 0.95}$ & $12.78_{\scriptstyle \pm 0.96}$
    & $30.67_{\scriptstyle \pm 0.96}$ & $12.86_{\scriptstyle \pm 0.38}$
    & $9.33_{\scriptstyle \pm 0.16}$ & $10.77_{\scriptstyle \pm 0.96}$
    & \underline{$45.33_{\scriptstyle \pm 1.07}$} & $12.38_{\scriptstyle \pm 0.72}$
    & $16.00_{\scriptstyle \pm 0.55}$ & $12.89_{\scriptstyle \pm 0.92}$ \\
    \textbf{AgentSafe}
    & \underline{$38.67_{\scriptstyle \pm 0.20}$} & $9.56_{\scriptstyle \pm 0.63}$
    & \underline{$40.00_{\scriptstyle \pm 0.43}$} & $11.57_{\scriptstyle \pm 0.41}$
    & $13.33_{\scriptstyle \pm 0.07}$ & $8.79_{\scriptstyle \pm 0.65}$
    & $33.33_{\scriptstyle \pm 0.16}$ & $9.59_{\scriptstyle \pm 0.54}$
    & $13.33_{\scriptstyle \pm 0.67}$ & $9.92_{\scriptstyle \pm 0.47}$ \\
    \cellcolor[HTML]{DADADA}\textbf{Ours}
    & \cellcolor[HTML]{DADADA}$\textbf{96.00}_{\scriptstyle \pm \textbf{0.18}}$ & \cellcolor[HTML]{DADADA}$\textbf{0.96}_{\scriptstyle \pm \textbf{0.52}}$
    & \cellcolor[HTML]{DADADA}$\textbf{76.67}_{\scriptstyle \pm \textbf{0.94}}$ & \cellcolor[HTML]{DADADA}$\textbf{1.06}_{\scriptstyle \pm \textbf{0.13}}$
    & \cellcolor[HTML]{DADADA}$\textbf{81.58}_{\scriptstyle \pm \textbf{0.12}}$ & \cellcolor[HTML]{DADADA}$\textbf{0.89}_{\scriptstyle \pm \textbf{0.08}}$
    & \cellcolor[HTML]{DADADA}$\textbf{73.77}_{\scriptstyle \pm \textbf{0.49}}$ & \cellcolor[HTML]{DADADA}$\textbf{0.53}_{\scriptstyle \pm \textbf{0.04}}$
    & \cellcolor[HTML]{DADADA}$\textbf{82.90}_{\scriptstyle \pm \textbf{0.39}}$ & \cellcolor[HTML]{DADADA}$\textbf{0.54}_{\scriptstyle \pm \textbf{0.06}}$ \\
    \specialrule{1.3pt}{2.0pt}{1.0pt}
    \end{tabular}}
    \label{tab:recovery-cost}
\end{table}

\section{Additional Results}
\label{app:results}

Sec.~6.5 reports benign calibration and hyperparameter sensitivity in the main text
(Fig.~\ref{fig:q4}). This appendix reports the two experiment groups deferred from there:
(i)~deployment robustness, i.e.\ whether contracts calibrated on benign logs remain useful when the
runtime environment differs from the one assumed by the workflow template
(Sec.~\ref{app:robustness}); and (ii)~localization under different MAS communication structures
(Sec.~\ref{app:topo-loc}). It then reports two checks of the repair stage: (iii)~how faithfully the
contract-space rollout of Appendix~\ref{app:rollout} predicts real re-execution
(Sec.~\ref{app:rollout-fidelity}); and (iv)~a comparison of localization and repair under the same
input, the same operators, and the same budget (Sec.~\ref{app:controlled-repair}).

\textbf{Protocol (Secs.~\ref{app:robustness}--\ref{app:topo-loc}).}
All stress tests vary \textbf{one} factor at a time. Attacks, task splits, seeds, and the metric
definitions are those of Appendix~\ref{app:experiment}, and the benign calibration procedure of
Appendix~\ref{app:repro-protocol} is applied unchanged in every configuration, so no method is
evaluated with a threshold fitted under a different regime. Method-side hyperparameters stay at the
defaults of Appendix~\ref{app:method-hparams}; nothing is re-tuned per configuration. Each panel
therefore isolates the effect of the environment on the monitors themselves rather than on their
tuning.

\subsection{Deployment Robustness}
\label{app:robustness}

The four factors below correspond to assumptions that monitoring methods commonly make about an MAS:
synchronous visibility, fixed scale, static topology, and comparable per-agent activity. Each factor
breaks one of them.

\textbf{Message delay (Fig.~\ref{fig:app-delay}).}
End-to-end message visibility is delayed by $\ell\in\{0,\ldots,5\}$ units, so an event may be observed
well after the step that produced it. \textsc{Trace-Mas} keeps detection F1 stable because the
partial-order event graph is built from producer--consumer links rather than from wall-clock rounds
(Appendix~\ref{app:graph}): delaying delivery changes when an edge is observed but not which events it
connects, and the context $c_t$ the contract conditions on is unchanged. Timing-sensitive baselines
(AcMAS(-R), A-Trust, AgentSafe) degrade as delay grows, since their per-round aggregates mix events
that no longer belong to the same logical step.

\textbf{Agent scale (Fig.~\ref{fig:app-scale}).}
We replicate role templates for $N\in\{2,4,8,16,32\}$, keeping the task and attack protocol fixed.
\textsc{Trace-Mas} remains comparatively flat because scoring is per event and conditioned on role and
relation type, so adding agents adds events without diluting the signal at the corrupted one.
Trust- and activation-heavy monitors lose signal as graphs grow, since they aggregate over more agents
to reach one decision and the corrupted contribution shrinks in the pooled statistic.

\textbf{Dynamic topology (Fig.~\ref{fig:app-topology-dynamic}).}
We vary the fraction of time-varying communication edges in $[0.2,1.0]$. Contract scores do not
reference a fixed adjacency matrix: the encoder aggregates whichever parents an event actually has, so
a re-routed message is still scored against the contract for its relation type. Topology-centric
baselines drop when edges churn, because edge churn is itself the anomaly signal they rely on.

\textbf{Execution imbalance (Fig.~\ref{fig:app-imbalance}).}
The per-agent event-rate ratio $\mathrm{IR}=\min_i f_i/\max_i f_i\in[0.2,1.0]$ (1 is balanced; smaller
is more skewed) sets how unevenly the agents speak. Because the
contract predicts a context-dependent spread $\Sigma_t$ as well as a mean
(Appendix~\ref{app:contract}), a talkative agent's larger raw deviations are normalized by its own
benign variability, which is exactly the effect described by Property~3 in
Appendix~\ref{app:theory-identifiability}. Role-conditioned residuals therefore avoid the volume-only
false alarms that hurt uniform-rhythm heuristics.

Taken together, the four panels test the deployment assumption behind Q4: thresholds fit on benign logs
should stay usable when coordination timing, scale, and structure move away from the calibration
setting, and the factors that degrade the baselines are the ones their representations depend on.

\newcommand{\appStressFig}[1]{%
  \IfFileExists{#1}{\includegraphics[width=\linewidth]{#1}}{%
    \fbox{\parbox[c][2.8cm][c]{\linewidth}{\centering\small Place \texttt{#1}}}}}

\begin{figure}[!ht]
  \centering
  \appStressFig{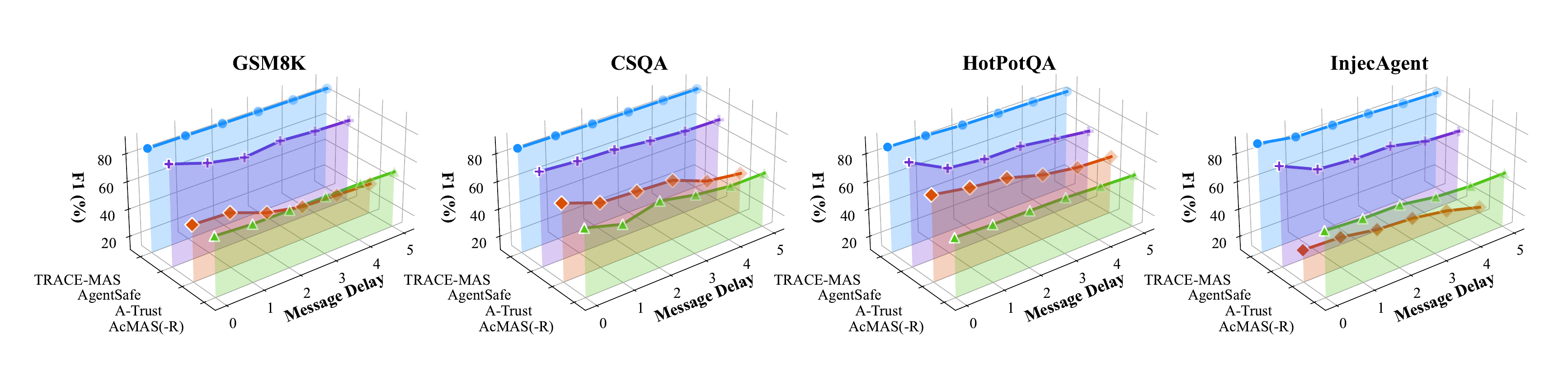}
  \caption{Effect of alert delay. Detection F1 under increasing end-to-end latency on GSM8K, CSQA, HotPotQA, and InjecAgent.
  \textsc{Trace-Mas} does not assume synchronous communication and stays comparatively flat.}
  \label{fig:app-delay}
\end{figure}

\begin{figure}[!ht]
  \centering
  \appStressFig{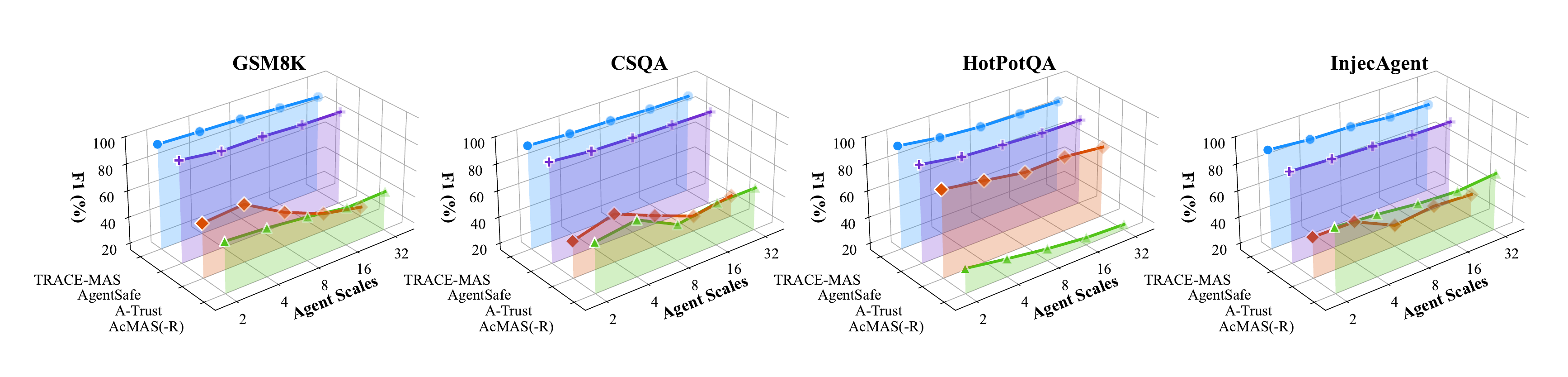}
  \caption{Robustness to agent scale. Detection F1 vs.\ $N\in\{2,4,8,16,32\}$ on four benchmarks.}
  \label{fig:app-scale}
\end{figure}

\begin{figure}[!ht]
  \centering
  \appStressFig{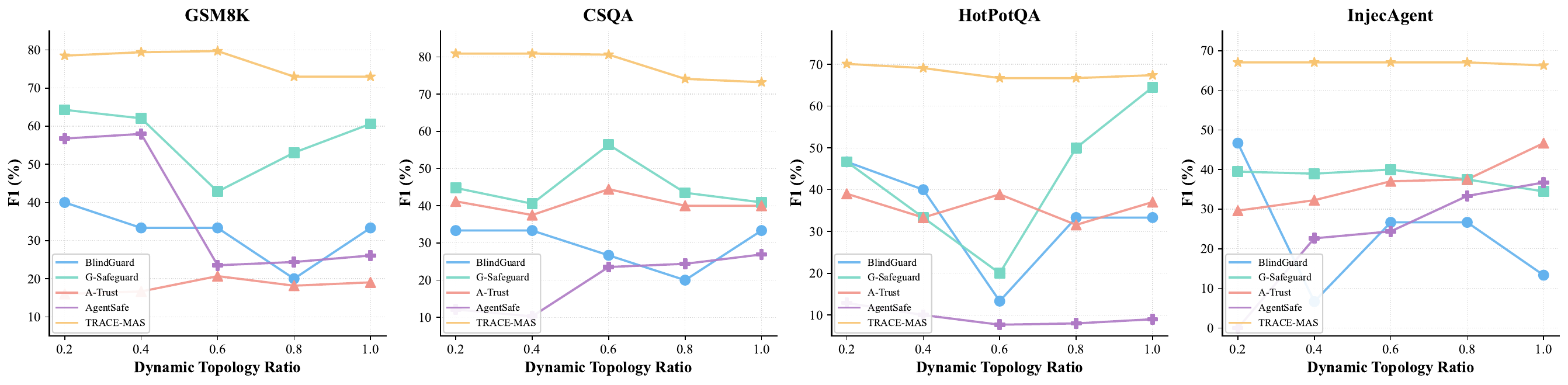}
  \caption{Robustness to dynamic topology. Detection F1 vs.\ dynamic-edge fraction on four benchmarks.}
  \label{fig:app-topology-dynamic}
\end{figure}

\begin{figure}[!ht]
  \centering
  \appStressFig{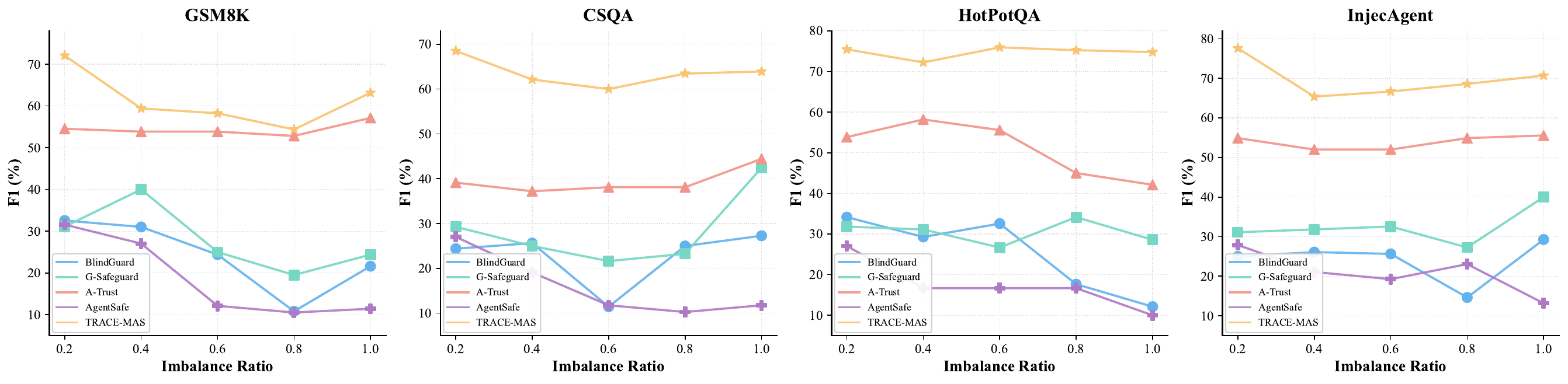}
  \caption{Robustness to execution imbalance. Detection F1 vs.\ imbalance ratio IR on four benchmarks.}
  \label{fig:app-imbalance}
\end{figure}

\FloatBarrier

\subsection{Localization under Varying Topologies}
\label{app:topo-loc}

\begin{figure}[H]
  \centering
  \IfFileExists{fig_topology_3d_hit_at_1_4datasets.pdf}{%
    \includegraphics[width=\linewidth]{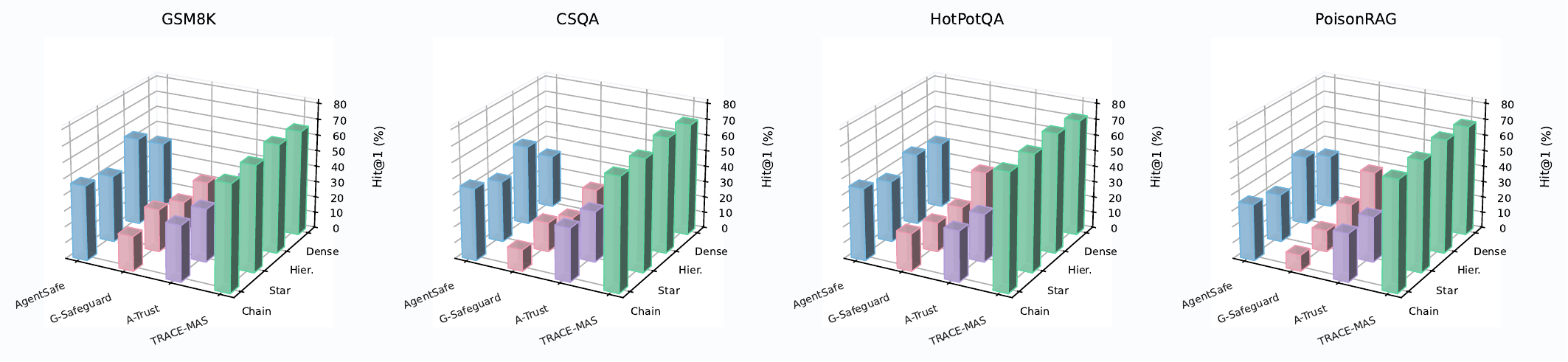}}{%
    \fbox{\parbox[c][3cm][c]{0.95\linewidth}{\centering\small Place topology Hit@1 figure as \texttt{fig\_topology\_3d\_hit\_at\_1\_4datasets.pdf}}}}%
  \caption{Localization under varying MAS topologies. Hit@1 on four benchmarks over Chain, Star, Hierarchical, and Dense structures.}
  \label{fig:topo-loc}
\end{figure}

Detection robustness does not imply attribution robustness: changing the communication structure changes
the ancestor sets that candidate sources are drawn from, and hence the number of plausible explanations
per alert. We therefore evaluate Hit@1 on GSM8K, CSQA, HotPotQA, and PoisonRAG under Chain, Star,
Hierarchical, and Dense communication templates (Fig.~\ref{fig:topo-loc}), with attacks, splits, and
defaults unchanged.

The four templates span the range that matters for attribution. A Chain gives each alert a single
upstream path, so candidates are few and mostly on one branch; a Star concentrates traffic through a
hub whose events are ancestors of almost everything; a Hierarchy adds depth, which increases the hop
distance between an injected source and the event where it becomes visible; and a Dense template
maximizes the number of ancestors within the hop limit $H$, so magnitude-based ranking has the most
opportunity to pick a symptom.

\textsc{Trace-Mas} consistently outperforms AgentSafe, G-Safeguard, and A-Trust across these structures
because attribution scores the violation drop obtained by restoring a candidate rather than the
violation level of an event (Appendix~\ref{app:attribution-resp}). By the locality of $\mathrm{Restore}$
(Property~4 in Appendix~\ref{app:theory-recovery}), candidates that are not upstream of the corrupted
branch contribute no drop and are ranked out automatically, which is what keeps the candidate set
manageable in the Star and Dense templates where the raw ancestor count is largest.

\FloatBarrier

\subsection{Rollout Fidelity under Real Re-execution}
\label{app:rollout-fidelity}

Repair selection ranks the operators in $\mathcal{U}(c^\star)$ by the rollout estimate
$\hat{\mathcal{J}}(u)$ instead of executing each of them. This subsection tests that substitution
directly by executing every candidate.

\textbf{Paired pool.}
Secs.~\ref{app:rollout-fidelity} and~\ref{app:controlled-repair} share one \emph{paired ablation pool}
$\mathcal{P}$: attack-successful trajectories from the Q3 repair pool (Sec.~\ref{app:metrics}), with
$|\mathcal{P}|{=}6$ trajectories, the \textbf{E2-lite} commit path, i.e.\ the repair stage of
Sec.~\ref{app:restoration} run at a budget of $B{=}1$ committed operator per trajectory with LLM
regeneration restricted to the edited branch, the deployment defaults $K{=}1$ and
Appendix~\ref{app:method-hparams}, and seed~0. Because both subsections need every candidate operator
re-executed by the trajectory LLM, $\mathcal{P}$ is sized for exhaustive re-execution rather than for
the per-benchmark resolution of the main tables, and all numbers below are single-seed. This pool is smaller than the full
attacked-test splits behind Table~\ref{tab:localization} and is used only to pair localization, rollout,
and re-execution under identical inputs; it is \textbf{not} a re-report of Table~\ref{tab:localization}.

\textbf{Procedure.}
For each trajectory in $\mathcal{P}$ the attributed event $c^\star$ is fixed. For every
$u\in\mathcal{U}(c^\star)$ we compute (i)~$\hat{\mathcal{J}}(u)$ from Appendix~\ref{app:rollout} and
(ii)~$\mathcal{J}(u)$ after committing $u$ and re-executing $\mathrm{Desc}(c^\star)$ with the same
decoding and template steps. Every candidate is executed once offline; deployment still commits only
$u^\star=\arg\min_u\hat{\mathcal{J}}(u)$.

\textbf{Reported quantities.}
\emph{Match} is the fraction of trajectories on which re-execution under every $u$ yields the same
$K$ window events as the rollout (Property~7 precondition). \emph{Agr} is the fraction with
$u^\star=u^{\mathrm{opt}}$ for $u^{\mathrm{opt}}=\arg\min_u\mathcal{J}(u)$. Spearman $\rho$ is the rank
correlation between $\hat{\mathcal{J}}$ and $\mathcal{J}$ over $\mathcal{U}(c^\star)$, averaged over
trajectories; it measures agreement of the induced \emph{ordering}, which is all that selection uses, and
not the size of $\lvert\hat{\mathcal{J}}-\mathcal{J}\rvert$. Once $u^\star=u^{\mathrm{opt}}$ on every
trajectory, the regret of Property~6 and $\Delta$TCR are zero by definition.

\begin{table}[H]
\centering
\small
\caption{Rollout fidelity under real LLM re-execution on the paired ablation pool (offline exhaustive
re-execution per candidate; deployed cost remains one commit).}
\begin{tabular}{lc}
\toprule
Metric (paired pool $\mathcal{P}$, $K{=}1$, E2-lite, $B{=}1$) & Value \\
\midrule
Match: rollout window $=$ re-execution window (\%) $\uparrow$ & 100 \\
Agr: $u^\star=u^{\mathrm{opt}}$ (\%) $\uparrow$ & 100 \\
Spearman$\bigl(\hat{\mathcal{J}},\mathcal{J}\bigr)$ $\uparrow$ & 1.00 \\
$\Delta$TCR: TCR($u^\star$)$-$TCR($u^{\mathrm{opt}}$) (points) & 0 \\
\bottomrule
\end{tabular}
\label{tab:rollout-fidelity}
\end{table}

Three readings follow, and they are deliberately limited to what the table measures. First, the
precondition of Property~7 holds throughout: at $K{=}1$ no operator in $\mathcal{U}(c^\star)$ added or
removed a window event under re-execution, so the matched-window form of
Eq.~(\ref{eq:app-rollout-decomp}) applies on every trajectory. Second, the rollout orders the operators
exactly as re-execution does ($\rho{=}1.00$) and selects the same operator on every trajectory, so the
conclusion of Property~6 and Corollary~1 holds here with zero regret, and the task cost of ranking by
$\hat{\mathcal{J}}$ instead of executing all candidates is zero. Third, this is evidence about the
\emph{ordering} rather than about the magnitude of $\nu(u)$ and $\theta(u)$: a rank correlation of $1.00$
is compatible with a non-zero value gap between $\hat{\mathcal{J}}$ and $\mathcal{J}$, and what the result
shows is that whatever gap remains is smaller than the separation between operators, which is the
condition Corollary~1 requires. Since the deployed rule uses only the ordering, this is the property the
repair stage depends on. Both tables are measured on $\mathcal{P}$; the Q3 numbers in
Tables~\ref{tab:recovery} and~\ref{tab:recovery-cost} remain those of the full test splits.

\subsection{Localization and Repair under the Same Input, Operators, and Budget}
\label{app:controlled-repair}

Tables~\ref{tab:recovery} and~\ref{tab:recovery-cost} evaluate each defense with its native response
(retry, quarantine, sanitization, or our repair), so they compare complete defenses. This subsection
separates the two stages on the paired pool $\mathcal{P}$ of Sec.~\ref{app:rollout-fidelity}.

\textbf{Protocol.}
\emph{Same input:} every row localizes on the same six trajectories in $\mathcal{P}$, whether or not its
own detector would have fired. \emph{Same operators:} each row's top-ranked event is passed to the
identical E2-lite repair module ($\mathcal{U}(\cdot)$, rollout with $K{=}1$, $\lambda_1,\lambda_2$ as in
Sec.~\ref{app:restoration}). \emph{Same budget:} $B{=}1$ committed operator per trajectory and the same
regeneration scope. Hit@1 is the fraction with $c^\star=c^{\mathrm{gt}}$ on $\mathcal{P}$ only; because
$|\mathcal{P}|{=}6$, percentages appear in steps of $100/6{\approx}16.7$ and \textbf{must not be compared}
to Table~\ref{tab:localization}, which aggregates the full attacked-test protocol.

\textbf{Localizers (Table~\ref{tab:controlled-loc}).}
Each baseline supplies its own ranking; \emph{Full monitoring} is contract violation alerting with
$\arg\max_t V(e_t)$ localization, i.e.\ the ``w/o CVA'' diagnostic of Fig.~2, in which counterfactual
violation attribution (CVA) is replaced by max-violation ranking, so neither $\mathrm{Restore}$ nor $R(c)$
is used;
\emph{\textsc{Trace-Mas} (CVA)} is the full rule $c^\star=\arg\max_c R(c)$ of Sec.~5.3. TCR uses the definition of Sec.~\ref{app:metrics}, evaluated on $\mathcal{P}$,
whose trajectories are attack-successful by construction.

\textbf{Selection rules (Table~\ref{tab:controlled-sel}).}
With $c^\star$ fixed, we vary only how $u$ is chosen over $\mathcal{U}(c^\star)$: the fixed operator used
in the w/o Predictive Repair ablation, a uniform random operator, rollout ranking
($\arg\min_u\hat{\mathcal{J}}(u)$), and exhaustive re-execution ranking ($\arg\min_u\mathcal{J}(u)$, the
reference from Sec.~\ref{app:rollout-fidelity}).

\begin{table}[H]
\centering
\small
\caption{Paired ablation on pool $\mathcal{P}$ ($|\mathcal{P}|{=}6$): localizer varied, repair module and
budget fixed (Hit@1 and TCR in \%, single seed). One trajectory is $16.7$ points, so single-trajectory gaps
are within the resolution of the pool and no entry is bolded; the paper's method comparisons are
Tables~\ref{tab:localization} and~\ref{tab:recovery-cost}, to which these values are not comparable.}
\begin{tabular}{lcc}
\toprule
Localization source (same E2-lite repair, $B{=}1$) & Hit@1$\uparrow$ & TCR$\uparrow$ \\
\midrule
\textsc{Trace-Mas} (CVA) & 16.7 & 66.7 \\
Full monitoring (Max-$V$) & 33.3 & 33.3 \\
AcMAS & 16.7 & 16.7 \\
AgentSafe & 33.3 & 50.0 \\
G-Safeguard & 0.0 & 33.3 \\
A-Trust & 33.3 & 33.3 \\
\bottomrule
\end{tabular}
\label{tab:controlled-loc}
\end{table}

\begin{table}[H]
\centering
\small
\caption{Paired ablation on $\mathcal{P}$: selection rule varied (TCR in \%, single seed). Rollout selection
commits the same operator as exhaustive re-execution, consistent with Table~\ref{tab:rollout-fidelity}.}
\begin{tabular}{lc}
\toprule
Selection over $\mathcal{U}(c^\star)$ ($c^\star$ fixed, E2-lite, $B{=}1$) & TCR$\uparrow$ \\
\midrule
Rollout $\hat{\mathcal{J}}$, $K{=}1$ (ours) & 66.7 \\
Fixed operator & 50.0 \\
Random operator & 33.3 \\
\midrule
Re-execution $\mathcal{J}$ (ref.) & 66.7 \\
\bottomrule
\end{tabular}
\label{tab:controlled-sel}
\end{table}

On $\mathcal{P}$, sharing the repair module isolates \emph{where} to act from \emph{how} to act. The pool is
small enough that a one-trajectory difference ($16.7$ points) is within its resolution, so we read only the
larger gaps. Read with that caveat, responsibility attribution recovers the most tasks of any localizer in the
table: two trajectories more than max-violation ranking ($66.7$ vs.\ $33.3$) and than the trust- and
topology-centric localizers G-Safeguard and A-Trust, three more than the activation-centric AcMAS
($16.7$), and one more than AgentSafe ($50.0$). The comparison with Max-$V$ is the informative one, because
that row shares our violation score and our repair module and differs only in replacing $R(c)$ with
$\arg\max_t V(e_t)$; the gap is the direction Obs.~2 and Property~5 predict, namely that the event with the
largest violation is not the event whose repair restores the task. Two entries should not be over-read.
Max-$V$ is one trajectory above CVA on Hit@1 here, which this pool cannot resolve and which the full
attacked-test protocol of Table~\ref{tab:localization} settles in the other direction; and the one-trajectory
gap to AgentSafe is at the resolution of the pool. What the pool adds over Table~\ref{tab:recovery-cost} is
that Hit@1 and TCR are measured on identical inputs through an identical repair module at an identical
budget, so a difference between rows is attributable to the localization stage alone; it is not itself a
method ranking, and the paper's recovery claims remain those of Tables~\ref{tab:recovery}
and~\ref{tab:recovery-cost}.

Two features of the table follow from the pool rather than from the methods and are worth stating so
they are not read as inconsistencies. First, every Hit@1 here is far below the corresponding column of
Table~\ref{tab:localization}: $\mathcal{P}$ contains only attack-successful trajectories at a single seed, i.e.\
the cases in which the corruption defeated the undefended workflow, and six such trajectories are not a
localization measurement. This is why the localization claims of the paper are made on the full protocol
and why we read only TCR differences here. Second, TCR exceeds Hit@1 for every row, which is expected
rather than contradictory: recovery does not require the exact initiating event, since committing an
operator on the corrupted branch near the source already restores many tasks, the effect quantified by
the hop buckets of Fig.~3 and by the tolerance to bounded amplification in Property~5. Hit@1 and TCR
therefore answer different questions on the same inputs, which is what makes the paired design useful.

Table~\ref{tab:controlled-sel} isolates the selection rule at a fixed $c^\star$, so its first row is the
same configuration as the \textsc{Trace-Mas} (CVA) row of Table~\ref{tab:controlled-loc} and carries the
same value by construction. Rollout selection equals exhaustive re-execution, which is forced by
Table~\ref{tab:rollout-fidelity}: at Agr $=100\%$ the two rules commit the same operator on every
trajectory, so they cannot differ in TCR. It is one trajectory above the fixed operator and two above
random selection, in the direction of the w/o Predictive Repair ablation, whose effect is quantified on the
full splits in Table~\ref{tab:ablation}. The point of the table is the pair of equalities and inequalities
together: choosing the operator by $\hat{\mathcal{J}}$ costs no task recovery relative to executing all
candidates, while ranking by anything cheaper than the rollout does.

\clearpage

\section{Analytical Support}
\label{app:theory}
This appendix explains, at the level of the quantities \textsc{Trace-Mas} computes, why the three
stages of Sec.~5 are built the way they are. Sec.~\ref{app:theory-identifiability} shows that the
violation score of Sec.~5.2 is the training objective of that section evaluated at test time, that the
benign quantile threshold of Eq.~(\ref{eq:app-tau}) controls false alarms without attack labels, and
that the detectable margin of a corruption is exactly the Mahalanobis norm of the deviation it
induces. Sec.~\ref{app:theory-propagation} derives why ranking events by that margin does not
localize the source. Sec.~\ref{app:theory-recovery} shows how the locality of $\mathrm{Restore}$ makes
the responsibility score of Sec.~5.3 separate an initiating event from its symptoms, and states the one
behavioral condition under which the separation holds. Sec.~\ref{app:theory-repair} gives the
stability of the repair-selection rule of Sec.~5.4 under rollout error and bounds that error for the
rollout of Appendix~\ref{app:rollout} in terms of the calibrated threshold $\tau$ and the depth $K$, and
Sec.~\ref{app:theory-cost} accounts for the cost of the pipeline. Sec.~\ref{app:theory-scope} states
what is not claimed.

All statements below are properties of the model we fit and the scores we report; none of them is used
to substitute for the empirical results in Sec.~6.

\subsection{Setup and Notation Bridge}
\label{app:theory-bridge}

We reuse the notation of Appendix~\ref{app:method}. For event $e_t$ with encoded context $c_t$, the
contract is the Gaussian $p_\theta(\Delta e_t\mid c_t)=\mathcal{N}(\mu_t,\Sigma_t)$ with
diagonal $\Sigma_t \succ 0$ (Eq.~(\ref{eq:app-contract})), and the violation score is the Mahalanobis
deviation $V(e_t)$ of Eq.~(\ref{eq:app-violation}) between the observed transition
$\widetilde{\Delta e}_t$ and $\mu_t$. Since the channel weights $(\beta,\gamma)$ of
Eq.~(\ref{eq:app-violation}) are equivalent to the rescaled diagonal covariance
$\tilde{\Sigma}_t=\mathrm{diag}(\Sigma^h_t,\Sigma^m_t/\beta,\Sigma^s_t/\gamma)$, we keep $\Sigma_t$ for the
trained covariance and write $\tilde{\Sigma}_t$ for the scoring covariance, so that
$V(e_t)=\lVert\widetilde{\Delta e}_t-\mu_t\rVert^2_{\tilde{\Sigma}_t^{-1}}$; the two coincide at
$\beta=\gamma=1$. With $d_h,d_m,d_s$ the channel dimensions ($d_z=d_h+d_m+d_s$), we write
$\bar{d}=\mathrm{tr}(\tilde{\Sigma}_t^{-1}\Sigma_t)=d_h+\beta d_m+\gamma d_s$, which equals $d_z$ at
$\beta=\gamma=1$. The residual shrinkage of Appendix~\ref{app:features} multiplies every violation term
by the same constant $\gamma_{\mathrm{res}}^2$. This changes no alert and not $\Omega$, and rescales the
margins of Property~3; in $R(c)$ and in the repair objective it rescales the violation terms relative to
$\lambda_{\mathrm{pen}}$, $\lambda_1$, and $\lambda_2$, so the statements below, written for the unshrunk
score, hold with these weights read as $\lambda/\gamma_{\mathrm{res}}^2$. We write
$\lVert b\rVert^2_{A} = b^\top A\, b$, $d_z$ for the dimension of $\Delta e_t$,
$d(c,e)$ for the hop distance in the partial-order graph, $\mathrm{Desc}(c)$ for the strict descendants
of $c$ (excluding $c$), and
\begin{equation}
\Delta_c(e) \;=\; V(e) - V^{\mathrm{res}}_c(e)
\end{equation}
for the violation drop at $e$ after restoring candidate $c$, where the sum in Eq.~(\ref{eq:app-R})
runs over $c$ together with its descendants. Because $\mathrm{Restore}_c$ writes the contract mean
$\mu_c$ while holding the context $c_c$ at its pre-restoration value (Appendix~\ref{app:restore-rollout}),
$V^{\mathrm{res}}_c(c)=0$ and hence $\Delta_c(c)=V(c)$. Since
$V(e)-\tilde{V}_c(e)=\omega\,\Delta_c(e)$, the responsibility score separates into a counterfactual
term and the validity penalty,
\begin{equation}
R(c) \;=\; \omega\,R_{\mathrm{cf}}(c) \;-\; \lambda_{\mathrm{pen}}\,\mathcal{P}(c),
\qquad
R_{\mathrm{cf}}(c) \;=\; \sum_{e\in\{c\}\cup\mathrm{Desc}(c)}\psi^{\,d(c,e)}\,\Delta_c(e),
\label{eq:app-Rcf}
\end{equation}
so the gate $\omega$ only rescales $R_{\mathrm{cf}}$ against $\mathcal{P}$. All comparisons below are
statements about $R_{\mathrm{cf}}$ and about the penalty difference between two candidates, so they
hold for any $\omega>0$.

Throughout, a corruption at event $e$ is described by the deviation it induces in contract space: the
realized residual becomes
\begin{equation}
\widetilde{\Delta e}_t \;=\; \mu_t + b_t + \varepsilon_t,
\qquad \varepsilon_t \sim \mathcal{N}(0,\Sigma_t),
\label{eq:app-shift}
\end{equation}
where $b_t$ is the induced shift ($b_t = 0$ on benign events) and $\varepsilon_t$ is the benign
variability the contract already accounts for. This description is not an assumption about the
attacker: any edit to a logged field produces some $b_t$, and $b_t$ is exactly the object our score
sees. It does assume that the contract is well specified on benign data, i.e.\ that benign residuals
follow $\mathcal{N}(0,\Sigma_t)$ with the trained $\Sigma_t$; Property~3 and Sec.~\ref{app:theory-propagation}
rely on this, whereas Properties~1, 2, 4, and~6 do not, and Property~7(b) uses instead the
exchangeability assumption stated there.

\subsection{From Benign Training to the Violation Score}
\label{app:theory-identifiability}

\medskip\noindent\textbf{Property 1 (the score is the training objective).}
For the scoring Gaussian $\mathcal{N}(\mu_t,\tilde{\Sigma}_t)$,
\begin{equation}
-\log \mathcal{N}(\widetilde{\Delta e}_t;\mu_t,\tilde{\Sigma}_t)
\;=\; \tfrac{1}{2}\Bigl[V(e_t) + \log\det\tilde{\Sigma}_t + d_z\log 2\pi\Bigr],
\end{equation}
so $V(e_t) = -2\log \mathcal{N}(\widetilde{\Delta e}_t;\mu_t,\tilde{\Sigma}_t) - \log\det\tilde{\Sigma}_t - d_z\log 2\pi$.

\noindent\emph{Support.} Substitute $\mathcal{N}(\mu_t,\tilde{\Sigma}_t)$ into the Gaussian density; since
$\tilde{\Sigma}_t^{-1}=\mathrm{diag}\bigl((\Sigma^h_t)^{-1},\beta(\Sigma^m_t)^{-1},\gamma(\Sigma^s_t)^{-1}\bigr)$,
the quadratic form is $V_h+\beta V_m+\gamma V_s$, i.e.\ Eq.~(\ref{eq:app-violation}). \hfill$\square$

Property~1 is an identity, not an approximation. At $\beta=\gamma=1$ it states that detection does not
introduce a new criterion at test time: the score is the quadratic term of the negative log-likelihood
in Eq.~(\ref{eq:app-nll}), evaluated on the test transition. With $(\beta,\gamma)=(1,0.5)$ the scored
quantity is the same likelihood with the completion channel's variance doubled, i.e.\ a fixed
reweighting of the trained contract rather than a different model. The implemented training loss
additionally weights the per-channel likelihoods by $(\lambda_h,\lambda_m,\lambda_s)$ and adds two
regularizers (Appendix~\ref{app:contract}); the identity therefore ties the score to the Gaussian
model that training fits (the same $\mu_t$ and $\Sigma_t$), not to the exact value of the implemented
loss. It also explains why the spread must be predicted rather than fixed: normalizing by
$\tilde{\Sigma}_t^{-1}$ is what makes the residual comparable across roles with different benign
variability, and the $\log\det$ term of the training likelihood is what prevents the learned spread
from growing to absorb large residuals.

\medskip\noindent\textbf{Property 2 (benign calibration controls false alarms).}
Let $V_1,\dots,V_n$ be the violation scores of the benign validation events and let $V^\ast$ be the
score of a fresh benign event, and suppose the calibration and test benign events are exchangeable and
the score function is fixed before calibration (the contract is fit on the disjoint training split).
If $\tau$ is the empirical $\delta$-quantile of $\{V_i\}$ as in Eq.~(\ref{eq:app-tau}) with
$\alpha{=}1$, then
\begin{equation}
\Pr\bigl[V^\ast > \tau\bigr] \;\le\; 1-\delta + \tfrac{1}{n+1},
\end{equation}
and for a benign trajectory with $|\mathcal{V}|$ events the alarm probability obeys
\begin{equation}
\Pr\bigl[\exists\,t:\ V(e_t)>\tau\bigr] \;\le\; |\mathcal{V}|\,\bigl(1-\delta+\tfrac{1}{n+1}\bigr).
\end{equation}

\noindent\emph{Support.} The first inequality is the standard exchangeability (conformal) bound \citep{vovk2005algorithmic,angelopoulos2021gentle} for an
empirical quantile threshold; the second follows by a union bound over the events of the trajectory,
which requires no independence between events. \hfill$\square$

Property~2 is the reason the calibration protocol of Sec.~\ref{app:detection} needs no attack labels:
the nominal false-alarm level is a property of $\delta$ and the benign calibration sample, so $\tau$ can
be fixed before any attack is observed. It also quantifies the choice $\delta{=}0.99$, since the
trajectory-level bound degrades linearly with trajectory length and is informative only when
$|\mathcal{V}|(1-\delta)$ is small; for long trajectories it is a worst-case statement rather than a
prediction of the observed alarm rate. Raising the threshold to $\alpha\cdot\mathrm{Quantile}_\delta$ with
$\alpha>1$ can only lower the alarm probability, so the bound continues to hold, whereas $\alpha<1$ is
not covered by it. The bound is one-sided by design; it says nothing about detection power, which is the
content of Property~3 and of Sec.~6.

\medskip\noindent\textbf{Property 3 (detectable margin of a corruption).}
Under the description of Eq.~(\ref{eq:app-shift}),
\begin{equation}
\mathbb{E}\bigl[V(e_t)\bigr] \;=\; \bar{d} + \lVert b_t\rVert^2_{\tilde{\Sigma}_t^{-1}},
\end{equation}
so a benign event has $\mathbb{E}[V]=\bar{d}$ and the expected margin of a corrupted event over a benign
one is exactly $\lVert b_t\rVert^2_{\tilde{\Sigma}_t^{-1}}$. At $\beta=\gamma=1$ this reads
$\mathbb{E}[V]=d_z+\lVert b_t\rVert^2_{\Sigma_t^{-1}}$, and $V$ is non-central $\chi^2_{d_z}$ with
$\mathrm{Var}[V]=2d_z+4\lVert b_t\rVert^2_{\Sigma_t^{-1}}$.

\noindent\emph{Support.} With $x = b_t+\varepsilon_t$ and $\varepsilon_t\sim\mathcal{N}(0,\Sigma_t)$,
$\mathbb{E}[x^\top\tilde{\Sigma}_t^{-1}x] = \mathrm{tr}(\tilde{\Sigma}_t^{-1}\Sigma_t) + b_t^\top\tilde{\Sigma}_t^{-1}b_t$.
At $\beta=\gamma=1$, $\Sigma_t^{-1/2}x\sim\mathcal{N}(\Sigma_t^{-1/2}b_t,I)$, and the variance is that of a
non-central $\chi^2$ with $d_z$ degrees of freedom and non-centrality $\lVert b_t\rVert^2_{\Sigma_t^{-1}}$.
\hfill$\square$

Property~3 turns a design question into a measurable one. Whether a corruption is visible depends not
on its textual size but on the Mahalanobis norm of the deviation it induces \emph{relative to the
benign spread of that context}, the same quantity that Mahalanobis-based out-of-distribution detectors
threshold \citep{lee2018simple}. This has three consequences for our design. First, conditioning the
contract on role and relation type (Sec.~\ref{app:encoder}) shrinks $\Sigma_t$ for a given edit and
therefore increases $\lVert b_t\rVert^2_{\tilde{\Sigma}_t^{-1}}$: the same injected message is more visible
against a role-specific expectation than against a global one. Second, modeling transitions rather than
absolute embeddings (Sec.~\ref{app:features}) removes benign content variation that would otherwise
inflate $\Sigma_t$. Third, it explains Obs.~1 without assuming anything about the attacker: a
coordination-level (L3) edit can leave the state channel almost unchanged, so monitors restricted to
agent activations see a small $b_t$ in the only channel they observe, while the same edit produces a
large shift in the message and completion channels the contract covers. We therefore do not assume a
separation margin between attacked and benign transitions; Property~3 identifies what such a
margin would consist of, and Table~\ref{tab:detection} measures whether it materializes.

\subsection{Why Magnitude Ranking Does Not Localize}
\label{app:theory-propagation}

Detection asks whether some event deviates; attribution asks which event started it. These are
different questions because the deviation propagates.

Consider a corruption at $c^{\mathrm{gt}}$ and a descendant $v$ reached along
$E_{\mathrm{msg}}\cup E_{\mathrm{dep}}$. The context $c_v$ that the contract conditions on is computed
from the (now corrupted) fields of the events upstream of $v$, so both the observed residual at $v$ and
the prediction $\mu_v$ change. Writing the induced shift at $v$ to first order in the shift at
the source,
\begin{equation}
b_v \;\approx\; J_{v\leftarrow c^{\mathrm{gt}}}\, b_{c^{\mathrm{gt}}},
\qquad
J_{v\leftarrow c^{\mathrm{gt}}} \;=\; \frac{\partial\bigl(\widetilde{\Delta e}_v - \mu_v\bigr)}{\partial b_{c^{\mathrm{gt}}}},
\end{equation}
Property~3 gives the resulting margins, and their ratio is
\begin{equation}
\frac{\mathbb{E}[V(v)]-\bar{d}}{\mathbb{E}[V(c^{\mathrm{gt}})]-\bar{d}}
\;\approx\;
\frac{\lVert J_{v\leftarrow c^{\mathrm{gt}}} b_{c^{\mathrm{gt}}}\rVert^2_{\tilde{\Sigma}_v^{-1}}}{\lVert b_{c^{\mathrm{gt}}}\rVert^2_{\tilde{\Sigma}_{c^{\mathrm{gt}}}^{-1}}} ,
\end{equation}
where the approximation is only the first-order one above. Nothing forces this ratio to be smaller than
one. It exceeds one whenever the propagation operator amplifies in the $\tilde{\Sigma}$-metric, which happens in exactly the situations our benchmarks contain: a
downstream role with tighter benign spread than the injecting role (small $\Sigma_v$), or a corrupted
payload that a downstream agent expands into a longer, more confidently expected step. This is the
formal content of Obs.~2 and of the mean-hop column of Table~\ref{tab:localization}: the
largest-violation event is frequently a symptom rather than the origin, so
$\arg\max_t V(e_t)$ is not guaranteed to return $c^{\mathrm{gt}}$, the same gap that motivates counterfactual root-cause analysis of outliers in causal models \citep{budhathoki2022causal}. Throughout Secs.~\ref{app:theory-propagation}--\ref{app:theory-recovery}, $c^{\mathrm{gt}}$ is the true initiating event, and the question is whether $c^\star=\arg\max_c R(c)$ recovers it.

We stress that we do \emph{not} posit an attenuating-propagation condition of the form
$V(v)\le r^{d(c^{\mathrm{gt}},v)}V(c^{\mathrm{gt}})$ with $r<1$. Such a condition would make magnitude ranking
correct by fiat and would contradict our own measurements. The attribution rule of Sec.~5.3 is designed
for the regime in which the ratio above can be large, and the next subsection shows why it survives
there.

\subsection{Why Counterfactual Responsibility Separates Source from Symptom}
\label{app:theory-recovery}

Two facts about $\mathrm{Restore}$ carry the argument; only the second involves the behavior of the
learned contract.

\medskip\noindent\textbf{Property 4 (locality of restoration).}
$\mathrm{Restore}(\tilde{G},c,\cdot)$ modifies the fields of $c$ and the contexts of
$\mathrm{Desc}(c)$ only. Hence $\Delta_c(e)=0$ for every $e\notin\{c\}\cup\mathrm{Desc}(c)$, and
$R_{\mathrm{cf}}(c)$ is a sum over $\{c\}\cup\mathrm{Desc}(c)$ only.

\noindent\emph{Support.} Immediate from the definition of $\mathrm{Restore}$
(Sec.~\ref{app:restore-rollout}) and from the encoder being causal: the context of an event depends
only on that event and its ancestors. \hfill$\square$

Property~4 already removes a large class of candidates from contention. If $c$ is unrelated to the
corrupted branch, then no corrupted event lies in $\{c\}\cup\mathrm{Desc}(c)$, so $R_{\mathrm{cf}}(c)$
collects only its own term $\Delta_c(c)=V(c)$ and benign-level changes on its descendants, all at the
scale of benign variability; such a candidate cannot outrank one that explains real violation mass
except through the validity penalty. The remaining competition is therefore between the true source
and the symptoms on its own branch, and that is where the hop weights and the source's own term act.

\medskip\noindent\textbf{Condition 1 (restoration at the origin is at least as effective).}
Let $v\in\mathrm{Desc}(c^{\mathrm{gt}})$ be a symptom such that every path from $c^{\mathrm{gt}}$ to an event of
$\mathrm{Desc}(v)$ passes through $v$. We use the condition
\begin{equation}
\Delta_{c^{\mathrm{gt}}}(e) \;\ge\; \Delta_{v}(e)
\qquad \text{for all } e\in\mathrm{Desc}(v).
\end{equation}

This is the single behavioral condition of this appendix, and it has a direct reading: on the part of
the branch that lies strictly below $v$, replacing the corrupted transition at its origin removes at least as
much violation as replacing it at $v$, because restoring $v$ leaves the corrupted context that feeds the
path from $c^{\mathrm{gt}}$ to $v$ in place. The condition is not imposed at $v$ itself, where $\Delta_v(v)=V(v)$
holds by construction and restoring the origin need not clear $v$ completely. It is also checkable
rather than hypothetical: on labeled trajectories in which both $c^{\mathrm{gt}}$ and $v$ are among the restored
candidates, both sides are quantities computed during attribution.

\medskip\noindent\textbf{Property 5 (responsibility margin).}
Write $S=\mathrm{Desc}(v)$ for the shared sub-branch,
$X=(\{c^{\mathrm{gt}}\}\cup\mathrm{Desc}(c^{\mathrm{gt}}))\setminus(\{v\}\cup S)$ for the exclusive segment (which
contains $c^{\mathrm{gt}}$ itself with weight $\psi^0=1$ and own term $\Delta_{c^{\mathrm{gt}}}(c^{\mathrm{gt}})=V(c^{\mathrm{gt}})$),
and $\zeta=\psi^{\,d(c^{\mathrm{gt}},v)}$. Under Condition~1 and Property~4,
\begin{equation}
R_{\mathrm{cf}}(c^{\mathrm{gt}})-R_{\mathrm{cf}}(v)
\;\ge\;
\underbrace{\sum_{e\in X}\psi^{\,d(c^{\mathrm{gt}},e)}\Delta_{c^{\mathrm{gt}}}(e)}_{\text{explained only by the origin}}
\;-\;\underbrace{\bigl[V(v)-\zeta\,\Delta_{c^{\mathrm{gt}}}(v)\bigr]}_{\text{symptom's own term}}
\;-\;(1-\zeta)\underbrace{\sum_{e\in S}\psi^{\,d(v,e)}\Delta_{c^{\mathrm{gt}}}(e)}_{\text{shared sub-branch}} .
\label{eq:app-margin}
\end{equation}
Consequently $R(c^{\mathrm{gt}})>R(v)$ whenever the right-hand side exceeds the penalty difference
$\lambda_{\mathrm{pen}}\bigl(\mathcal{P}(c^{\mathrm{gt}})-\mathcal{P}(v)\bigr)/\omega$.

\noindent\emph{Support.} Split $R_{\mathrm{cf}}(c^{\mathrm{gt}})$ over $X$, $\{v\}$, and $S$. The term at $v$ is
$\zeta\Delta_{c^{\mathrm{gt}}}(v)$, and $R_{\mathrm{cf}}(v)$ contributes $\Delta_v(v)=V(v)$ at weight $\psi^0=1$.
For $e\in S$ every path from $c^{\mathrm{gt}}$ passes through $v$, so $d(c^{\mathrm{gt}},e)=d(c^{\mathrm{gt}},v)+d(v,e)$ and
$\psi^{\,d(c^{\mathrm{gt}},e)}=\zeta\,\psi^{\,d(v,e)}$; Condition~1 then gives
$\zeta\Delta_{c^{\mathrm{gt}}}(e)-\Delta_v(e)\ge -(1-\zeta)\Delta_{c^{\mathrm{gt}}}(e)$ on $S$. Subtracting
$R_{\mathrm{cf}}(v)$ and collecting the three parts gives Eq.~(\ref{eq:app-margin}). \hfill$\square$

Property~5 is the analytical counterpart of the design decisions in Sec.~\ref{app:attribution-resp},
and it makes explicit both what the rule gains over magnitude ranking and where it can still fail.
Summing over the candidate itself as well as its descendants is what puts $V(c^{\mathrm{gt}})$ into the
exclusive segment at full weight. The symptom's own violation does enter the margin: since
$V^{\mathrm{res}}_{c^{\mathrm{gt}}}(v)\ge 0$ we have $\Delta_{c^{\mathrm{gt}}}(v)\le V(v)$, so the middle term is at
least $(1-\zeta)V(v)$. What changes relative to magnitude ranking is its weight. Magnitude ranking
selects $c^{\mathrm{gt}}$ only if $V(c^{\mathrm{gt}})>V(v)$; when restoring the origin clears the symptom
($\Delta_{c^{\mathrm{gt}}}(v)\approx V(v)$), responsibility charges $V(v)$ only at weight $1-\zeta$ ($0.5$ for a
one-hop symptom at the default $\psi{=}0.5$) and credits the origin with the rest of the exclusive
segment. The rule therefore tolerates a bounded amount of amplification, and this covers part of the
regime of Sec.~\ref{app:theory-propagation} in which magnitude ranking already fails. It does not tolerate
arbitrary amplification: if $V(v)$ or the mass on the shared sub-branch is large relative to the
exclusive segment, or if restoring the origin leaves much of $v$'s violation in place, the right-hand
side can be negative and the symptom can outrank the source. This is one source of the non-zero mean
hop distances in Table~\ref{tab:localization}. The Hit@1 and mean-hop results in
Table~\ref{tab:localization} and the comparison with \textsc{Trace-Mas} (w/o CVA) in Fig.~2 are the
empirical test of this mechanism.

\subsection{Stability of Repair Selection and Size of the Rollout Error}
\label{app:theory-repair}

Repairs are ranked by the rollout estimate of Sec.~5.4 rather than by executing every candidate, so two
questions arise: how selection behaves under rollout error, and how large that error is for the rollout
we actually use. Write $\hat{\mathcal{J}}(u)$ for the rollout estimate of the repair objective and
$\mathcal{J}(u)$ for its realized value after committing $u$ and re-executing $\mathrm{Desc}(c^\star)$,
both defined in Appendix~\ref{app:rollout}, and $u^\star=\arg\min_{u\in\mathcal{U}(c^\star)}\hat{\mathcal{J}}(u)$,
$u^{\mathrm{opt}}=\arg\min_{u\in\mathcal{U}(c^\star)}\mathcal{J}(u)$.

\medskip\noindent\textbf{Property 6 (selection is stable up to twice the offset-free error).}
Suppose there is a constant $\bar{b}$, which may depend on the trajectory but not on $u$, such that
\begin{equation}
\sup_{u\in\mathcal{U}(c^\star)}\bigl|\hat{\mathcal{J}}(u)-\mathcal{J}(u)-\bar{b}\bigr|\;\le\;\epsilon' .
\label{eq:app-offset-error}
\end{equation}
Then $\mathcal{J}(u^\star) \le \mathcal{J}(u^{\mathrm{opt}}) + 2\epsilon'$.

\noindent\emph{Support.} Selection is unchanged if $\bar{b}$ is subtracted from every $\hat{\mathcal{J}}(u)$, so
$\mathcal{J}(u^\star)\le\hat{\mathcal{J}}(u^\star)-\bar{b}+\epsilon'\le\hat{\mathcal{J}}(u^{\mathrm{opt}})-\bar{b}+\epsilon'\le \mathcal{J}(u^{\mathrm{opt}})+2\epsilon'$.
\hfill$\square$

\medskip\noindent\textbf{Corollary 1 (exact selection under a rollout margin).}
Under Eq.~(\ref{eq:app-offset-error}), if $\hat{\mathcal{J}}(u)-\hat{\mathcal{J}}(u^\star)>2\epsilon'$ for every
$u\neq u^\star$, then $u^\star=u^{\mathrm{opt}}$.

\noindent\emph{Support.} For $u\neq u^\star$,
$\mathcal{J}(u)\ge\hat{\mathcal{J}}(u)-\bar{b}-\epsilon'>\hat{\mathcal{J}}(u^\star)-\bar{b}+\epsilon'\ge\mathcal{J}(u^\star)$.
\hfill$\square$

Property~6 needs no assumption about which operator is correct, and the offset $\bar{b}$ makes it
insensitive to any error shared by all operators on a trajectory; $\bar{b}=0$ recovers the uniform
form. Corollary~1 turns the premise into a certificate that can be checked per trajectory once
$\epsilon'$ is known. What remains is to show that $\epsilon'$ is small for the rollout of
Appendix~\ref{app:rollout}, which the next property does by tracing the error to the two places where
rollout and re-execution differ.

For each $u$, let $\mathcal{A}(u)$ be the (event, channel) pairs that the rollout sets to the contract
mean (the edited channel of $c^\star$ and all channels of the regenerated window events), and
$\mathcal{B}(u)$ the remaining pairs in the window, i.e.\ the two unedited channels of $c^\star$ and all
channels of the carry-over events. For a pair $(i,\mathrm{ch})$, let $\hat{V}_{i,\mathrm{ch}}$ and
$V^{\mathrm{re}}_{i,\mathrm{ch}}$ be its weighted channel term ($V_h$, $\beta V_m$, or $\gamma V_s$) in the
rollout and in the re-execution, and define
\begin{equation}
\nu(u)=\sum_{(i,\mathrm{ch})\in\mathcal{B}(u)}\bigl|\hat{V}_{i,\mathrm{ch}}-V^{\mathrm{re}}_{i,\mathrm{ch}}\bigr|,
\qquad
\theta(u)=\bigl|\hat{\mathcal{L}}_{\mathrm{task}}(u)-\mathcal{L}_{\mathrm{task}}(u)\bigr|\in[0,1].
\end{equation}

\medskip\noindent\textbf{Property 7 (rollout error of \textsc{Trace-Mas}).}
Suppose re-execution under $u$ produces the same $K$ window events as the rollout (matched event by
event), which holds for any operator that does not add or remove events among the first $K$
descendants of $c^\star$.
(a)~For every such $u$,
\begin{equation}
\bigl|\hat{\mathcal{J}}(u)-\mathcal{J}(u)\bigr|
\;\le\;
\sum_{(i,\mathrm{ch})\in\mathcal{A}(u)} V^{\mathrm{re}}_{i,\mathrm{ch}} \;+\; \nu(u) \;+\; \lambda_1\,\theta(u).
\label{eq:app-rollout-decomp}
\end{equation}
(b)~Suppose that each re-executed regenerated window event is a benign event of its context, and that
the re-executed edited channel of $c^\star$ is distributed as that channel of a benign event of its
context, in both cases exchangeable with the benign calibration events of Eq.~(\ref{eq:app-tau}). Let
$\tau$ be the threshold on the scale of $V$ (Sec.~\ref{app:theory-bridge}) with $\alpha\ge 1$. Then, with
probability at least $1-|\mathcal{U}(c^\star)|\,(K{+}1)\bigl(1-\delta+\tfrac{1}{n+1}\bigr)$,
\begin{equation}
\sup_{u\in\mathcal{U}(c^\star)}\bigl|\hat{\mathcal{J}}(u)-\mathcal{J}(u)\bigr|
\;\le\;
(K{+}1)\,\tau \;+\; \max_u \nu(u) \;+\; \lambda_1\max_u \theta(u),
\label{eq:app-rollout-bound}
\end{equation}
so Eq.~(\ref{eq:app-offset-error}) holds with $\bar{b}=0$ and $\epsilon'$ equal to the right-hand side.

\noindent\emph{Support.} (a)~The cost $\kappa(u)$ is the same deterministic quantity in both, so it
cancels. The rollout scores every pair in $\mathcal{A}(u)$ as zero (Eq.~(\ref{eq:app-rollout})), whereas
re-execution scores it as $V^{\mathrm{re}}_{i,\mathrm{ch}}\ge0$; pairs in $\mathcal{B}(u)$ contribute at
most $\nu(u)$; and the task terms differ by $\lambda_1\theta(u)$. The triangle inequality gives
Eq.~(\ref{eq:app-rollout-decomp}). (b)~For each $u$, the first term of
Eq.~(\ref{eq:app-rollout-decomp}) is a sum of at most $K{+}1$ quantities: the full score of each of the at
most $K$ regenerated window events, and the edited channel term of $c^\star$, which is at most the full
score of a benign event with that channel. By Property~2 and $\alpha\ge1$, each exceeds $\tau$ with
probability at most $1-\delta+1/(n+1)$; a union bound over the at most $|\mathcal{U}(c^\star)|(K{+}1)$
quantities bounds the first term by $(K{+}1)\tau$ for every $u$ simultaneously. \hfill$\square$

Property~7 ties the error to the design choices of Appendix~\ref{app:rollout} and to the operating point
of Sec.~\ref{app:method-hparams}. First, the only term that grows with the depth is $(K{+}1)\tau$, and the
failure probability grows with $K$ as well, so deeper rollouts can only loosen the guarantee; at the
deployed $K{=}1$ the first term is $2\tau$, and with $\delta{=}0.99$ and $\alpha{=}1$ the bound holds with
probability at least $1-2|\mathcal{U}(c^\star)|\,(0.01+1/(n+1))$, e.g.\ $1-6\,(0.01+1/(n+1))$ with one
candidate per operator type. Deeper rollouts thus buy no tighter guarantee, which is consistent with TCR
being unchanged from $K{=}1$ to $K{=}6$ in Fig.~\ref{fig:efficiency}(c). Second, the first term is on the
benign scale: $\tau$ is the level below which a benign event stays with probability at least
$\delta-1/(n+1)$, whereas an operator that leaves a corrupted field in place keeps that field's carry-over
violation in $\hat{\mathcal{J}}$, whose expected size is, by the channel-wise form of Property~3, the benign
mean of that channel plus $\lVert b\rVert^2_{\tilde{\Sigma}^{-1}}$, with $b$ the shift the field induces
relative to the repaired context. By Corollary~1, rollout selection is therefore exact whenever the
rollout objective of every other operator, which contains the carry-over violation that operator leaves in
place, exceeds that of $u^\star$ by more than $2\epsilon'$. Third, $\theta(u)$ is zero whenever re-execution
invokes the template steps the rollout predicts, and $\nu(u)$ is zero whenever the carry-over events
reproduce their logged transitions under the same contexts as in the rollout; neither is controlled analytically, which is why Appendix~\ref{app:rollout-fidelity} measures
both on the paired pool $\mathcal{P}$. There, the matched-window precondition holds on every trajectory and
rollout selection coincides with exhaustive re-execution, so whatever value gap $\nu$ and $\theta$ leave is
smaller than the separation between operators, which is what Corollary~1 needs; the experiment bounds the
gap through its effect on the ordering rather than measuring it directly. The assumption in (b) is the one the rollout itself makes, namely that an event whose
inputs are repaired behaves benignly; it can fail for an operator that does not remove the corruption, in
which case $\mathcal{J}(u)$ exceeds the bound for that $u$. The cost term $\lambda_2\kappa(u)$ biases
selection in a definite direction: between operators with similar violation and task terms, the rule prefers
the smaller intervention, which is consistent with the low operator cost reported in
Table~\ref{tab:recovery-cost}.

\subsection{Cost Accounting}
\label{app:theory-cost}

The pipeline of Sec.~\ref{app:pipeline} has a cost that follows from the definitions above rather than
from measurement. Detection evaluates the contract once per event, at $O(|\mathcal{V}|\,d_z^2)$ time after context
encoding. Attribution restores the top $C$ candidates and rescores their descendants, at
$O\bigl(C\cdot\max_c|\mathrm{Desc}(c)|\cdot d_z^2\bigr)$; without gating on $\Omega$ and on $C$, scoring
every ancestor of every violated event would scale with $|\mathcal{V}|\cdot\max_c|\mathrm{Desc}(c)|$, which is the
comparison the $C$ sweep in Fig.~\ref{fig:efficiency} makes concrete. Repair selection adds
$O\bigl(|\mathcal{U}(c^\star)|\cdot (K{+}1)\cdot d_z^2\bigr)$ for the candidate operators at $c^\star$ (the
edited event and $K$ window events per operator, Appendix~\ref{app:rollout}). The re-executions of every
candidate in Appendix~\ref{app:rollout-fidelity} are an offline validation and are not part of this cost.

Two structural points follow. Counterfactual scoring and rollout are executed in contract space, so the
number of LLM calls added by \textsc{Trace-Mas} is the number of \emph{committed} repairs, not the number
of candidates considered; and because $C$, $K$, and $|\mathcal{U}(c^\star)|$ are small constants, the added non-LLM work is
linear in trajectory length. This is why the wall-clock profile in Fig.~\ref{fig:efficiency} is reported
excluding bulk generation: the design moves the search off the generation path by construction.

\subsection{Scope}
\label{app:theory-scope}

The statements above support the components of Sec.~5 and are deliberately narrow. Properties~1--3
characterize what the detection score measures and what its benign calibration guarantees; they do not
claim that every corruption produces a large margin, and Property~3 makes the dependence on the induced
shift explicit instead of assuming separability. Sec.~\ref{app:theory-propagation} derives why magnitude
ranking is not localization but does not predict the amount of amplification for a given workflow.
Properties~4--5 compare the true source against symptoms on its own branch under one checkable
condition, and they apply only when $c^{\mathrm{gt}}$ enters the candidate set, i.e.\ lies in $\Omega$ or within
$H$ hops above it and survives the top-$C$ pre-ranking; they do not establish identifiability against an adversary that corrupts several events, nor
optimality over the whole candidate set when Condition~1 fails, and Eq.~(\ref{eq:app-margin}) can be negative under strong amplification at a symptom. Property~6 and Corollary~1 bound the suboptimality of
repair selection given rollout error, and Property~7 bounds that error for our rollout only under a matched
window and the benign-regeneration assumption of Property~7(b), with the carry-over and template terms
left to measurement (Appendix~\ref{app:rollout-fidelity}); none of them covers sequences of several
committed edits or the completeness of the operator space $\mathcal{U}(c^\star)$. The quantitative behavior
of all these mechanisms is established empirically in Sec.~6 and Appendix~\ref{app:results}.

\end{document}